\PassOptionsToPackage{unicode}{hyperref}
\PassOptionsToPackage{hyphens}{url}
\PassOptionsToPackage{dvipsnames,svgnames,x11names}{xcolor}
\documentclass[
  12pt]{article}

\usepackage{amsmath,amssymb}
\usepackage{iftex}
\ifPDFTeX
  \usepackage[T1]{fontenc}
  \usepackage[utf8]{inputenc}
  \usepackage{textcomp}
\else
  \usepackage{unicode-math}
  \defaultfontfeatures{Scale=MatchLowercase}
  \defaultfontfeatures[\rmfamily]{Ligatures=TeX,Scale=1}
\fi
\usepackage{lmodern}
\ifPDFTeX\else  
\fi
\IfFileExists{upquote.sty}{\usepackage{upquote}}{}
\IfFileExists{microtype.sty}{
  \usepackage[]{microtype}
  \UseMicrotypeSet[protrusion]{basicmath}
}{}
\makeatletter
\@ifundefined{KOMAClassName}{
  \IfFileExists{parskip.sty}{
    \usepackage{parskip}
  }{
    \setlength{\parindent}{0pt}
    \setlength{\parskip}{6pt plus 2pt minus 1pt}}
}{
  \KOMAoptions{parskip=half}}
\makeatother
\usepackage{xcolor}
\makeatletter
\ifx\paragraph\undefined\else
  \let\oldparagraph\paragraph
  \renewcommand{\paragraph}{
    \@ifstar
      \xxxParagraphStar
      \xxxParagraphNoStar
  }
  \newcommand{\xxxParagraphStar}[1]{\oldparagraph*{#1}\mbox{}}
  \newcommand{\xxxParagraphNoStar}[1]{\oldparagraph{#1}\mbox{}}
\fi
\ifx\subparagraph\undefined\else
  \let\oldsubparagraph\subparagraph
  \renewcommand{\subparagraph}{
    \@ifstar
      \xxxSubParagraphStar
      \xxxSubParagraphNoStar
  }
  \newcommand{\xxxSubParagraphStar}[1]{\oldsubparagraph*{#1}\mbox{}}
  \newcommand{\xxxSubParagraphNoStar}[1]{\oldsubparagraph{#1}\mbox{}}
\fi
\makeatother

\usepackage{longtable,booktabs,array}
\usepackage{calc}
\usepackage{etoolbox}
\makeatletter
\patchcmd\longtable{\par}{\if@noskipsec\mbox{}\fi\par}{}{}
\makeatother
\IfFileExists{footnotehyper.sty}{\usepackage{footnotehyper}}{\usepackage{footnote}}
\makesavenoteenv{longtable}
\usepackage{graphicx}
\makeatletter
\def\maxwidth{\ifdim\Gin@nat@width>\linewidth\linewidth\else\Gin@nat@width\fi}
\def\maxheight{\ifdim\Gin@nat@height>\textheight\textheight\else\Gin@nat@height\fi}
\makeatother
\setkeys{Gin}{width=\maxwidth,height=\maxheight,keepaspectratio}
\makeatletter
\def\fps@figure{htbp}
\makeatother

\makeatletter
\@ifpackageloaded{caption}{}{\usepackage{caption}}
\AtBeginDocument{
\ifdefined\contentsname
  \renewcommand*\contentsname{Table of contents}
\else
  \newcommand\contentsname{Table of contents}
\fi
\ifdefined\listfigurename
  \renewcommand*\listfigurename{List of Figures}
\else
  \newcommand\listfigurename{List of Figures}
\fi
\ifdefined\listtablename
  \renewcommand*\listtablename{List of Tables}
\else
  \newcommand\listtablename{List of Tables}
\fi
\ifdefined\figurename
  \renewcommand*\figurename{Figure}
\else
  \newcommand\figurename{Figure}
\fi
\ifdefined\tablename
  \renewcommand*\tablename{Table}
\else
  \newcommand\tablename{Table}
\fi
}
\@ifpackageloaded{float}{}{\usepackage{float}}
\floatstyle{ruled}
\@ifundefined{c@chapter}{\newfloat{codelisting}{h}{lop}}{\newfloat{codelisting}{h}{lop}[chapter]}
\floatname{codelisting}{Listing}

\makeatother
\makeatletter
\@ifpackageloaded{caption}{}{\usepackage{caption}}
\@ifpackageloaded{subcaption}{}{\usepackage{subcaption}}
\makeatother

\ifLuaTeX
  \usepackage{selnolig}
\fi
\usepackage[]{natbib}
\usepackage{bookmark}

\IfFileExists{xurl.sty}{\usepackage{xurl}}{}
\hypersetup{
  pdftitle={Neural Variational Cut Posteriors without Upstream Data},
  pdfauthor={Jiafang Song; Sandipan Pramanik; Abhirup Datta},
  pdfkeywords={Modular Bayes, Cutting feedback, Normalizing Flows, Variational Inference},
  colorlinks=true,
  linkcolor={black},
  filecolor={black},
  citecolor={black},
  urlcolor={black},
  pdfcreator={LaTeX via pandoc}}

\newcommand{\anon}{1}

\usepackage[page wise]{lineno}
\usepackage{enumitem}
\setlist{topsep=2pt, partopsep=0pt, itemsep=2pt, parsep=0pt}
\usepackage{pdflscape}
\usepackage{amsmath,color,amsfonts,rotating,amsthm,mathtools,setspace,mathabx,bbm,multirow,threeparttable,lipsum,multicol,tabularx,comment}
\usepackage{algorithm}
\usepackage{algpseudocode}
\usepackage{xspace}
\usepackage{appendix}
\usepackage{makecell}
\usepackage{adjustbox}
\usepackage{threeparttable}
\usepackage{multirow}

\newcommand{\given}{\,|\,}
\newcommand{\iid}{\overset{\text{iid}}\sim}
\newcommand{\law}{\overset{\text{d}}=}
\newcommand{\ind}{\overset{\text{ind}}\sim}
\newcommand{\probarg}{(\theta,\eta \given D_1,D_2)}
\DeclareMathOperator{\Law}{Law}

\newcommand{\clip}{\mathrm{clip}}
\newcommand{\diam}{\mathrm{diam}}

\newtheorem{assumption}{Assumption}
\newtheorem{theorem}{Theorem}
\newtheorem{lemma}{Lemma}

\newtheorem{proposition}{Proposition}
\newtheorem{corollary}{Corollary}
\newcommand{\classindex}{\left(L_S\left(\cdot\right),L_{CP}\left(\cdot\right),A_d,B_d,c,\alpha\right)}

\newcommand{\calB}{\mathcal B}

\newcommand{\calD}{\mathcal D}

\newcommand{\calF}{\mathcal F}

\newcommand{\calI}{\mathcal I}

\newcommand{\calL}{\mathcal L}

\newcommand{\calN}{\mathcal N}

\newcommand{\calP}{\mathcal P}
\newcommand{\calQ}{\mathcal Q}

\newcommand{\bc}{ {\boldsymbol c} }

\newcommand{\bq}{ {\boldsymbol q} }

\newcommand{\bv}{ {\boldsymbol v} }

\newcommand{\bbP}{\mathbb{P}}
\newcommand{\bbR}{\mathbb{R}}
\newcommand{\con}{{\,|\,}}

\newcommand{\abs}[1]{\left\vert#1\right\vert}

\newcommand{\pc}{p_{\text{cut}}}

\usepackage{hyperref}

\begin{document}
\def\spacingset#1{\renewcommand{\baselinestretch}
{#1}\small\normalsize} \spacingset{1}

\if1\anon
{
  \title{\bf Neural Variational Cut Posteriors without Upstream Data}
  \author{Jiafang Song$^\dag$, Sandipan Pramanik$^\dag$, Abhirup Datta\footnote{Email address for correspondence: abhidatta@jhu.edu $\quad ^\dag$ Equal contributions.} \\
Department of Biostatistics, Johns Hopkins University, USA}
  \date{}
  \maketitle
} \fi

\if0\anon
{
  \title{\bf Neural Variational Cut Posteriors without Upstream Data}
  \date{}
  \maketitle
} \fi

\begin{abstract}
    In many scientific applications, it is essential to propagate parameter uncertainty from an earlier (upstream) analysis, available as samples, to subsequent (downstream) analyses without feedback (downstream data updating upstream posterior). The problem is called \emph{cutting feedback} or \emph{cut-Bayes}, and \emph{the cut-posterior}, the optimal posterior that preserves the information flow constraints, is well characterized.
    However, sampling from the cut-posterior (e.g., via nested MCMC) can be computationally intensive, while existing variational inference (VI) methods for cut-Bayes require
    access to upstream data and model, which are often not available. We propose a modular and provably accurate cut-Bayes approach
    that does not require access to upstream data or model. We leverage the characterization of the cut-posterior as the minimizer of the expected downstream conditional Kullback-Leibler divergence, where the expectation is over the upstream posterior distribution. This allows us to simply replace the population expectation with the sample average over the upstream parameter draws.
    For accurate posterior approximation, our method, \emph{NeVI-Cut (neural variational inference for cut-Bayes)}, employs conditional normalizing flows (neural network-based expressive distribution class) as the variational family for the downstream parameters.
    We provide fixed-data convergence rates of NeVI-Cut in terms of the richness of neural architecture and complexity of the target cut-posterior. In the process, we establish, to our knowledge, first results
    on uniform Kullback-Leibler approximation rates of conditional distributions by common flow classes (e.g., neural spline flows), which, in turn, yields widely applicable results on fixed-data error rates of variational flows. A stochastic algorithm implements NeVI-Cut efficiently. We demonstrate the speed and accuracy of NeVI-Cut across multiple real-world applications.
\end{abstract}


\spacingset{1.8}

\section{Introduction}\label{sec:Introduction}

In many scientific applications, uncertainty about parameters or predictive variables from an upstream analysis, available as precomputed distributions, must be propagated into a downstream analysis. Examples include climate impacts studies \citep{butler2022increasing,
rao2024future} using ensemble forecasts from climate models \citep{thrasher2022nasa}, health impact studies of air pollution \citep{murray2020global} using posterior samples of exposure risk curves published by \cite{gbd2019_pm_risk_curves}, incorporating estimated exposure uncertainty in health outcome analyses \citep{
comess2024bayesian},
utilizing published posterior samples of confusion matrices of cause-of-death classifiers \citep{Pramanik2026bmjgh} for calibrating disease burden attribution, and employing probabilistic word-embeddings for large language models (LLMs) such as {\em word2Gauss} and {\em Bayesian skip-gram} for classifying text \citep{vilnis2014word,bravzinskas2018embedding}.

The upstream quantities are often causally precedent (e.g., PM$_{2.5}$ concentrations) to downstream ones (e.g., health outcomes) and often come from large-scale, broadly used analyses (climate-model ensembles or pretrained LLM embeddings), serving as inputs to many downstream applications. It is thus both conceptually inappropriate and operationally unrealistic for such globally shared upstream distributions to be revised in response to the data or modeling choices of any single (and potentially misspecified) downstream analysis.

Approaches that prevent feedback from downstream to upstream are known as \emph{cutting feedback} or \emph{cut-Bayes} \citep{Bayarri2009, plummer2015cuts}. It estimates the \emph{cut-posterior}, a valid joint distribution of upstream and downstream parameters that preserves the upstream distribution. The literature on cut-Bayes is now large and rapidly growing, see e.g., \cite{smith2025cutting,liu2025general} for recent theoretical and methodological advancements, and  \cite{frazier2026posterior,jacob2017better} for data-driven justification of cutting. A comprehensive review is beyond the scope of this paper. We recommend \cite{nott2023bayesian} for a review of earlier work.
A major thread of developments in cut-Bayes has been computational.
\cite{plummer2015cuts} showed that the `cut' algorithm in OpenBUGS does not converge to the cut-posterior.
The gold-standard \emph{nested Markov Chain Monte Carlo}  \citep[nested MCMC][]{plummer2015cuts} runs a separate downstream MCMC analysis for each upstream posterior sample and pools the results to represent the cut-posterior \citep{plummer2015cuts}. This greatly increases computational cost \citep{carmona2020semi}. Other sampling-based methods like the \emph{tempered} cut algorithm \citep{plummer2015cuts}, an unbiased MCMC estimator with coupled chains \citep{jacob2020unbiased}, and semi-modular inference \citep[SMI;][]{carmona2020semi} also require nested sampling.

Variational inference (VI) has also been explored to address computational challenges of cut-Bayes. 
\cite{yu2023variational} demonstrated that the cut-posterior is the solution of a constrained variational optimization --- the constraint being preservation of the upstream posterior. Many proposals, including theirs, rely on parametric variational families (e.g., Gaussian), which can be problematic if misspecified. To mitigate this, conditional normalizing flows \citep{kruse2021hint} have been employed for semi-modular inference,
a broader framework for managing feedback across modules, with cut-Bayes as a special case \citep{carmona2020semi}.  \citet{carmona2022scalable}  used conditional normalizing flows for a flexible variational approximation of the SMI \emph{variational meta-posterior} (a continuum of posteriors indexed by and amortized over the feedback magnitude parameter). \citet{battaglia2025amortising} extended it to additionally amortize over the hyperparameters. 

The current strand of literature of variational cut-Bayes has two major gaps:

\begin{itemize}
    \item Methodologically, all of the existing variational cut-Bayes methods need the upstream data and model (or, at least, a variational fit of it) and thus cannot be used without upstream data (i.e., when only the posterior samples are available for the upstream parameters, as is the case for many of the aforementioned applications).
\item On the theoretical side, while normalizing flows are expressive distribution classes, 
there has been no theoretical study of how well the normalizing flow-based variational cut posterior, optimizing the constrained Kullback-Leibler Divergence (KLD), approximates the true cut-posterior, and what dictates their accuracy. In fact, we did not find any fixed-target KL-convergence rate results for conditional normalizing flows.
\end{itemize}

This manuscript addresses these two issues. 
Leveraging neural networks and VI, we present the \emph{Neural Variational Inference for Cut-Bayes (NeVI-Cut)}, a modular and flexible cut-Bayes method. NeVI-Cut requires no knowledge of the upstream data and model, and directly uses parameter samples from the upstream analysis and is agnostic to the method that generates them, provided they sufficiently approximate the upstream posterior. NeVI-Cut aims to be interoperable, since it couples modules through upstream posterior draws alone rather than shared data or model code \citep{nicholson2022interoperability}. This compartmentalization is achieved by exploiting the characterization of the joint KLD for the cut posterior as
the expected conditional KLD for the downstream posterior, with the expectation being over the upstream posterior. We approximate the loss by simply replacing the upstream population expectation with the sample average based on the upstream draws. While methodologically simple, this change makes variational cut-Bayes far more broadly applicable by eliminating any need to access upstream data or models, which are often unavailable because of privacy concerns, as with human-subjects data, or because the upstream analysis is a massive, complex process, such as climate-model systems or proprietary LLM-training pipelines.
For accurate cut-posterior estimation, we use conditional normalizing flows as the conditional variational class. 
This allows NeVI-Cut to leverage the expressive power of neural networks and better approximate complex (e.g., multimodal, skewed) cut-posteriors.

Our second contribution is to develop a comprehensive theoretical framework for NeVI-Cut, variational cut-Bayes, and normalizing flows.
Our theory operates in a `fixed-data' or `fixed-target' regime, analyzing the quality of variational approximation for the exact cut-posterior, conditional on the given data, without assuming any sample size growth. This setting is more challenging for VI. With large samples, Bernstein-von Mises theorem suggests that cut posteriors are asymptotically Gaussian \citep{pompe2021asymptotics,smith2025cutting}. So, even simple Gaussian VI can be accurate. This safeguard is absent in the fixed-data regime, where posteriors may exhibit complex geometries. We prove that the NeVI-Cut variational solution converges to the conditional cut-posterior given the data. The theory accounts for dependencies among upstream samples as they can be MCMC posterior draws.

For the theory of NeVI-Cut, we establish universal and uniform (over the conditioning variable) KLD approximation rates for conditional normalizing flows. This differs from previous findings on the representational capacities of normalizing flows \citep{huang2018neural, papamakarios2021normalizing} and the universal approximation results for flow-based variational meta-posteriors \citep{battaglia2025amortising}. These results focus on convergence in distribution (weak metric); they do not ensure control over the behavior of variational solutions, specifically the variational cut- or meta-posterior estimates, which are minimizers of a more stringent KLD criterion. Our theory fills this important gap. We
also explicitly quantify how the KLD approximation rates of conditional normalizing flows relate to the complexity of the target conditional cut-posterior and the expressiveness of the neural architectures defining the flows, a contribution we believe is novel. The theory applies to common flow classes, including \textit{Rational Quadratic Neural Spline Flows} \citep[RQ-NSF;][]{durkan2019neural} and \textit{Unconstrained Monotonic Neural Network Flows} \citep[UMNN;][]{wehenkel2019unconstrained}. 
Beyond helping establish theoretical guarantees for the NeVI-Cut estimate, the theory is of independent importance and facilitates the study of other applications of conditional normalizing flows, such as variational meta-posteriors in SMI.

The remainder of this paper is organized as follows: In Section \ref{sec:rev}, we
review cut-Bayes, VI, and normalizing flows. In Section \ref{sec:method} we present our NeVI-Cut algorithm.
Theoretical results are presented in Section \ref{sec:theory}.
In Section \ref{sec:realdata}, we benchmark NeVI-Cut on four real-world datasets spanning global health, disease mapping, and linguistics.
Section \ref{sec:discussion} concludes the paper with a discussion. The NeVI-Cut Python package
is made available on {GitHub}.

\section{Background}
\label{sec:rev}
\subsection{Cutting Feedback Approaches}\label{sec:cutrev}

We consider a setup where the upstream module involves data $D_1$ and quantities of interest (parameters, predictive variables) $\eta$. The downstream data is $D_2$ which is modeled using both $\eta$ and some additional parameters $\theta$.
If jointly analyzed,
both $D_1$ and $D_2$ inform the Bayesian posterior $p(\eta \given D_1,D_2)$,
which therefore differs from its upstream posterior $p(\eta \given D_1)$. As mentioned in the Section~\ref{sec:Introduction}, this is often undesirable.
Cutting feedback (or cut-Bayes)
targets the
cut-posterior 
\begin{equation}\label{eq:cutjoint}
    p_{\text{cut}}(\theta,\eta \given D_1,D_2) :=\, p (\theta \given \eta,D_1, D_2) \, p_{\text{cut}}(\eta \given D_1, D_2) = p (\theta \given \eta,D_2) \, p(\eta \given D_1).
\end{equation}
Here, $p (\theta \given \eta,D_1, D_2) = p (\theta \given \eta,D_2)$
since the likelihood of $D_1$ does not depend on $\theta$.
From \eqref{eq:cutjoint}, we have 
$p_{\text{cut}}(\eta \given D_1,D_2) := p(\eta \given D_1)$. Hence, the cutting feedback preserves the posterior of $\eta$ from upstream analysis.

\subsection{Variational Inference for Cut-Bayes}\label{sec:virev}
Given data $ D =(D_1,D_2)$ and model parameters $\phi=(\theta,\eta)$, variational inference (VI)
seeks a distribution $q(\phi)$ within a variational family $\mathcal{Q}$ to approximate the posterior distribution $p(\phi \given  D)$. Formally, the variational solution minimizes the KLD to $p(\phi \given  D)$ and is defined as
\begin{equation}\label{eq:vi}
    {q}^{*}(\phi)=\underset{q(\phi) \in \mathcal{Q}}{\arg \min}\,\,\mathrm{KL} \left({q}(\phi)\parallel p(\phi\given D)\right).
\end{equation}
If $\calQ$ is unrestricted, then $q^*(\phi) = p(\phi\given D)$, the target posterior distribution. However, in practice, parametric assumptions are often imposed on $\calQ$ to enable feasible optimization, yielding an approximate solution.

In a fundamental contribution, Lemma 4.1 of \cite{yu2023variational} facilitates the use of VI for cut-Bayes. Under the constraint $\calQ = \left\{q(\theta,\eta) : q(\eta) = p(\eta \given D_1) \right\}$ to preserve the upstream posterior, it shows that the variational solution is the joint cut-posterior (\ref{eq:cutjoint}):
\begin{equation}\label{eq:cutvi}
    p_\text{cut} (\theta, \eta \given D_1,D_2) =
    \underset{q(\theta,\eta) \in \mathcal{Q} : q(\eta) = p(\eta \given D_1)}{\arg \min}\,\,\mathrm{KL}\,\!({q}(\theta,\eta)\parallel p(\theta,\eta \given D_1,D_2)).
\end{equation}
Thus, under the constraint of no feedback from $D_2$ to $\eta$, the cut-posterior best approximates the Bayes posterior in KLD. For implementation, they decompose $q(\theta,\eta) = q(\theta \given \eta) \, q(\eta)$ and use two variational approximations using parametric families. While these can be learnt simultaneously using the stop-gradient operator \citep{carmona2022scalable}, requiring access to both upstream data and model is unavoidable.
\cite{smith2025cutting} used similar ideas for copula models. The cut-posterior can also be obtained via VI for semi-modular inference \citep[SMI,][]{carmona2020semi,carmona2022scalable,battaglia2025amortising}, but they also rely on the availability of the upstream data and model (or at least a variational fit for the upstream posterior). 

\subsection{Normalizing Flows}

Normalizing flows provide a flexible variational family by constructing complex densities through a sequence of invertible and differentiable transformations of a simple base distribution, and have been widely explored in VI \citep{rezende2015variational}. 
For any cumulative distribution function (cdf) $F$ of a scalar $\theta$, there exists a function $T$ such that $\theta \overset{d}{=} T(Z)$, where $Z$ is some known base distribution. The function $T$ can be explicitly obtained as the inverse cdf transformation, e.g., if $Z \sim N(0,1)$, then $T(Z) = F^{-1}(\Phi(Z))$ where $\Phi$ is the standard normal cdf. Thus, approximating the distribution of $\theta$ is equivalent to approximating the function $T$. Since neural networks are universal approximators of various function classes, it is sufficient to model $T$ using an appropriate class of neural networks $T_\vartheta$ and estimate its parameters $\vartheta$ (weights and biases) with the KLD loss. Normalizing flows utilize this core idea, though specific choices to model $T$ vary widely. \cite{papamakarios2021normalizing} provides a thorough review, while \cite{kruse2021hint} extends it to modeling conditional distributions, which is used to model the variational meta-posteriors  \citep{carmona2022scalable,battaglia2025amortising} and will also be used in our method.

\section{Neural Variational Inference for Cut-Bayes}\label{sec:method}

All of the existing variational cut or SMI methods described above require the upstream data and model \citep{carmona2022scalable,yu2023variational,smith2025cutting,battaglia2025amortising}. We propose a VI  cut method that can directly use draws of $\eta$ from the upstream posterior $p(\eta \given D_1)$ without requiring
the upstream data or any component of the upstream model.
Recall from (\ref{eq:cutvi}) that
the cut-posterior is the solution to the variational optimization with the variational family $q(\theta, \eta)$ constrained to have $q(\eta)=p(\eta \given D_1)$ to prevent feedback. As $q(\eta)$ is fixed,
the optimization in (\ref{eq:cutvi}) is only
over the class of conditional distributions $q(\theta \given \eta)$. Under the constraint $q(\eta)=p(\eta \given D_1)$, the KLD in (\ref{eq:cutvi}) can be written as
\begin{equation}\label{eq:KLnested}
\mathrm{KL} ({q}(\theta,\eta)\parallel p(\theta,\eta \given D_1,D_2)) = \mathbb{E}_{\eta \sim p(\eta \given D_1)} \big[ \mathrm{KL} (q(\theta \given \eta)  \parallel p(\theta \given \eta, D_2)) \big]. 
\end{equation}

Consequently, the joint cut-posterior can be expressed in terms of the variational solution as
\begin{equation}\label{eq:cutvicond}
\begin{aligned}
\pc\probarg &= q^{*}_{\text{cut}}(\theta \given \eta) \, p(\eta \given D_1), \mbox{ where }\\ 
q^{*}_{\text{cut}}(\theta \given \eta) &= 
\underset{q(\theta \given \eta)}{\arg \min}\; \mathbb{E}_{\eta \sim p(\eta \given D_1)} \left[ \mathrm{KL}\,\!(q(\theta \given \eta)  \parallel p(\theta \given \eta, D_2)) \right]. 
\end{aligned}
\end{equation}
Thus, finding the cut-posterior reduces to estimating the collection of conditional distributions $\{q(\theta \given \eta)\}_{\eta} $. 

The representation \eqref{eq:KLnested} is crucial to our method. The joint KLD loss under the cut-constraint becomes the expectation of the conditional KLD, with the expectation being over the upstream posterior. This allows approximating the outer expectation with samples $\eta_1,\ldots,\eta_N$ drawn from $p(\eta \given D_1)$. We will show later that as long as we have enough samples that are iid or from an ergodic MCMC, we can approximate the population expectation (\ref{eq:KLnested}) with the sample average. This leads to the joint and conditional cut-posterior estimates
\begin{equation}\label{eq:cutvicondsample}
\begin{aligned}
\hat{p}_{\text{cut}} \probarg &= \hat{q}_{\text{cut}}(\theta \given \eta) \, p(\eta \given D_1), \mbox{ where } \\ 
\hat{q}_{\text{cut}}(\theta \given \eta) &= 
\underset{q(\theta \given \eta)}{\arg \min}\; \frac 1N \sum_{i=1}^N \mathrm{KL}\,\!(q(\theta \given \eta_i)  \parallel p(\theta \given \eta_i, D_2)) \\
 &= \underset{q(\theta \given \eta)}{\arg \min}\; \frac 1N \sum_{i=1}^N E_{q(\theta \given \eta_i)} \big[ \log q(\theta \given \eta_i) - \log p(D_2 \given \theta,\eta_i) - \log p(\theta \given \eta_i) \big].
\end{aligned}
\end{equation}
From \eqref{eq:cutvicondsample}, it is evident that we do not need access to the upstream module or a variational approximation of $p(\eta \given D_1)$, unlike existing methods. It only needs the samples $\{\eta_i\}$ which are often the only accessible quantity, as in the examples mentioned at the onset. Even if the upstream data and model are available, one can simply run an MCMC 
to obtain the samples $\{\eta_i\}$ from $p(\eta \given D_1)$ in (\ref{eq:cutvicondsample}), and then run NeVI-Cut, although in this scenario several of the existing VI-based cut methods can also be used.

\subsection{Normalizing Flows for Conditional Distribution}\label{sec:norm flow conditional var family}

Once the likelihood $p(D_2 \given \theta, \eta)$ and prior $p(\theta \given \eta)$ for downstream analysis are chosen, the only unknowns in (\ref{eq:cutvicondsample}) are the variational parameters specifying the class of conditional distributions $q(\theta \given \eta)$.
We model $q(\theta \given \eta)$ using conditional normalizing flows \citep{kruse2021hint}, which represent distributions as transformations of simple base distributions. For $\theta \in \mathbb{R}^d$ and given $\eta \in \mathbb R^{d_\eta}$, we define the family of conditional distributions $q_\vartheta(\theta \given \eta)$ corresponding to the law $\theta = T_\vartheta(\eta, Z)$. Here $T_\vartheta(\eta, Z)$ is a neural network-based transformation with arguments $(\eta,Z)$, $Z\in\mathbb{R}^d$ is the random variable endowed with a known base distribution with density $p(Z)$ that is both analytically tractable and easy to sample from, and $\vartheta$ are the parameters specifying the neural network $T_\vartheta$.
NeVI-Cut can be implemented with any choice of conditional normalizing flow. We detail some popular specifications in Section \ref{sec:NeVI-Cut Specification} of the Supplementary Materials. 
A common choice for the base is $Z \sim N(0, I_d)$.
However, Cauchy or Student's $t$ may be more appropriate base distributions for true posteriors with heavy tails (see Section \ref{sec:theory} for theoretical guidance).

Using the classical {\em reparameterization trick} of VI,
the Kullback-Leibler objective (\ref{eq:cutvicond})
can be expressed in terms of $p(Z)$ as  
\begin{equation}\label{eq:cutvicondflow}
\mathbb E_{\eta\sim p(\eta \given D_1)}\,\mathbb E_{Z\sim p(Z)}\big[\log q_\vartheta\!\left(T_\vartheta(\eta,Z)\given\eta\right)-\log p\!\left(T_\vartheta(\eta,Z)\given\eta,D_2\right)\big]. 
\end{equation}
Writing \(J_z T_\vartheta(\eta,z)\) for the Jacobian of \(T_\vartheta\) with respect to \(z\), and using change of variables (see Lemma \ref{lem:quantile} for a formal statement and proof), we have the identity
\[\log q_\vartheta\!\left(T_\vartheta(\eta,z)\given\eta\right)=\log p(z)-\log\!\left|\det J_z T_\vartheta(\eta,z)\right|.\]
Plugging this in (\ref{eq:cutvicondflow}), we can write the loss as 
\[\mathbb E_{\eta\sim p(\eta \given D_1)}\,\mathbb E_{Z\sim p(Z)}\big[\log p(Z)-\log|\det J_z T_\vartheta(\eta,Z)|-\log p\!\left(T_\vartheta(\eta,Z)\given\eta, D_2\right)\big]. \]
Since $\log p(Z)$ does not involve $\vartheta$ and $\log p \left(T_\vartheta(\eta,Z)\given\eta,D_2\right) = \log p \left(D_2 \given T_\vartheta(\eta,Z),\eta\right) + \log p \left( T_\vartheta(\eta,Z) \given \eta\right) + \text{ terms free of } \vartheta,$
minimizing \eqref{eq:cutvicondflow} is equivalent to maximizing the average conditional {\em evidence lower bound (ELBO)}
$$ \mathcal L(\vartheta)= \mathbb E_{\eta\sim p(\eta \given D_1)}\,\mathbb E_{Z\sim p(Z)}\big[\log|\det J_z T_\vartheta(\eta,Z)|+\log p\!\left(D_2 \given T_\vartheta(\eta,Z),\eta\right) + \log p\!\left( T_\vartheta(\eta,Z) \given \eta\right)\big].$$
In practice, as we only work with the upstream samples $\{\eta_i\}_i$, we use
\begin{equation}\label{eq:ELBO}
\begin{aligned}
\hat{\calL} (\vartheta) =\frac 1N \sum_{i=1}^N \,\mathbb E_{Z\sim p(Z)}\big[\log|\det J_z T_\vartheta(\eta_i,Z)| + \log p\!\left(D_2 \given T_\vartheta(\eta_i,Z),\eta_i\right) + \log p\!\left( T_\vartheta(\eta_i,Z) \given \eta_i\right)\big].
\end{aligned}
\end{equation} 

The ELBO $\hat{\calL} (\vartheta)$ can be calculated for any value of $\vartheta$ as every other quantity is known. Hence, we obtain $\hat{\vartheta} \in {\arg\max}_\vartheta \; \hat{\calL} (\vartheta)$.
and estimate the cut-posterior as 
\begin{equation}\label{eq:nevicut}
    \hat{p}_{\text{cut}}\probarg = q_{\hat{\vartheta}}(\theta \given \eta) \, p(\eta \given D_1).
\end{equation}
By drawing $Z_i \overset{\text{iid}}{\sim} p(Z)$, we obtain samples $\{\eta_i,T_{\hat{\vartheta}}(\eta_i,Z_i)\}$ from the cut-posterior. We refer to our method as \emph{NeVI-Cut} (neural variational inference for cut-Bayes).

\begin{algorithm}[!t]
\caption{Neural Variational Inference for Cut-Bayes (NeVI-Cut)}
\label{alg}
\begin{algorithmic}
\State \textbf{Input:} \textit{Upstream:} Posterior samples $\{\eta_i\}_{i=1}^N$

\hspace{6.6mm}\textit{Downstream:} Data $D_2$, likelihood $p(D_2 \given \theta, \eta)$, prior $p(\theta \given \eta)$

\hspace{6.6mm}\textit{Normalizing flow:} Choice of flow (e.g., Neural Spline Flows, or Unconstrained

\hspace{40.5mm}Monotonic Neural Network Flows)

\hspace{6.6mm}\textit{Optimization:} Learning rate $\alpha$, maximum number of iterations ($S$), 

\hspace{33.5mm}patience ($S_{\text{patience}}$; number of steps before early stopping)
\State \textbf{Initialize:} Set some initial values of $\vartheta$

\For{$s = 1$ to $S$}
    \State Sample base variables $\{Z_i\}_{i=1}^N \iid N(0,I)$
    \State Transform $\theta_i = T_\vartheta(\eta_i, Z_i)$ for $i=1,\ldots,N$
    \State Estimate ELBO:
    \begin{equation}\label{eq:ELBOprac}
        \hat{\calL}_{Z}(\vartheta)  = \frac 1{N} \sum_{i =1}^N  \big[\log|\det J_z T_\vartheta(\eta_i,Z_i)|+ \log p\!\left(D_2 \given \theta_i,\eta_i\right) + \log p\!\left( \theta_i \given \eta_i\right)\big]
    \end{equation}
    \State Update parameters $\vartheta \leftarrow \vartheta + \alpha \, \mathrm{Adam}\bigl(\nabla_\vartheta \hat{\calL}_Z(\vartheta)\bigr)$\;(\mbox{gradient ascent})
    \State \textbf{Early stopping:} End if $\hat{\calL}_{Z}(\vartheta)$ fails to improve for $S_{\text{patience}}$ steps
\EndFor
\State \textbf{Final estimate:} $\hat{\vartheta} \leftarrow \vartheta$
\State \textbf{Samples:} $\widetilde Z_i \iid N(0,I);\; \widetilde \theta_i = T_{\hat{\vartheta}}(\eta_i, \widetilde Z_i)$ for $i=1,\ldots,N$
\State \textbf{Output:} Trained conditional flow $\hat{q} (\theta \given \eta) \law T_{\hat{\vartheta}}(\eta,Z), Z \sim N(0,I)$, 
cut-posterior estimate 
\begin{equation}\label{eq:cutsample}
[\theta \given \eta \sim \hat{q} (\theta \given \eta)] \times [\eta \sim \text{Unif}\{\eta_1,\ldots,\eta_N\}],
\end{equation}
samples from the cut-posterior $\{\eta_i,\widetilde \theta_i\}_{i=1}^N$
\end{algorithmic}
\end{algorithm}

\subsection{Triply-Stochastic Algorithm}
Computing the objective in (\ref{eq:ELBO}) poses three main computational challenges. First, a large number of samples $\left\{\eta_i\right\}$ may be available from the upstream analysis. This makes it costly to evaluate the ELBO averaged over all upstream samples at every iteration. Second, the expectation with respect to $Z$ is typically intractable. Third, evaluating the likelihood $p(D_2 \given \theta,\eta)$ can be expensive, particularly for large datasets. We address these issues by exploiting three corresponding sources of stochasticity.

We deploy mini-batching and stochastic gradients, which are widely used to speed up large-scale optimization by replacing averages over all terms with averages over small, randomly chosen subsets (mini-batches).
At each iteration, we can replace the average over all $N$ samples of $\eta$ with a small mini-batch $\calB_\eta$ of size $N_\eta$. Similarly, the expectation with respect to $Z$ is replaced by an average of $N_Z$ iid observations $\{Z_j\}$ sampled from $p(Z)$. The doubly stochastic approximation of (\ref{eq:ELBO}) is thus given by
\begin{equation*}
\hat{\calL}_{DS}(\vartheta) = \frac 1{N_\eta N_Z} \sum_{i \in \calB_\eta}  \sum_{j=1}^{N_Z} \Big[ \log|\det J_z T_\vartheta(\eta_i,Z_j)| + \log p\!\left(D_2 \given T_\vartheta(\eta_i,Z_j),\eta_i\right) + \log p\!\left( T_\vartheta(\eta_i,Z_j) \given \eta_i\right)\big].
\end{equation*}

Additionally, if the downstream data $D_2$ consists of iid units $Y_1,\ldots,Y_n$, then we have $p(D_2 \given \theta, \eta) = \prod_{i=1}^n p(Y_i \given \theta, \eta)$. And  $\log p(D_2 \given \theta, \eta)$ can be approximated using a mini-batch $\calB_D$ of size $N_D$ as $\frac n {N_D} \sum_{k \in \calB_D} \log p(Y_k \given \theta, \eta)$, giving a triply stochastic approximation. 

In practice, the three approximations need not be used simultaneously, and the stochasticity level, determined by the mini-batch sizes $N_\eta, N_Z, N_D$, can be adjusted according to computational trade-offs. To ensure comparability with the nested MCMC in our experiments, we restrict stochasticity to Monte Carlo integration over $Z$. The optimization is performed using the Adam algorithm \citep{kingma2015adam}.
Algorithm \ref{alg} summarizes the method, and we made the
NeVI-Cut Python package publicly available on
{GitHub}.

\section{Theory}\label{sec:theory}
In this section, we provide theoretical rates of universal approximation of the true cut-posterior $\pc(\theta,\eta \given D_1,D_2)$ by the NeVI-Cut estimator $\hat{p}_{\text{cut}}\probarg$.

\subsection{Uniform Kullback-Leibler Divergence Approximation Rates of Conditional Normalizing Flows}\label{sec:klrate}

We first provide a new and general result on universal approximation rates (in Kullback-Leibler divergence) of conditional normalizing flows in terms of the richness of the neural network architecture and complexity of the class of conditional densities it is approximating. These results are then used to prove the convergence of NeVI-Cut, but are also of independent importance, as discussed in the Introduction.
We consider the setting in which the conditioning variable $\eta$ is multivariate and the observation variable $\theta$ is univariate.
This setting helps to illustrate the main proof ideas and insights.
The extension to multivariate $\theta=(\theta_1,\ldots,\theta_d)^\top$ is conceptually straightforward by iterating on a sequence of univariate conditional distributions $(\theta_1 \given \eta),  (\theta_2 \given \eta,\theta_1), \ldots$ but will involve more technical derivations.

We first specify the true class of target conditional distributions. 
\begin{assumption}[True distribution class]\label{asm:mono}Let $K_\eta \in \mathbb R^{d_\eta}$ be a compact set and let $K_M := [-M,M]\times K_\eta$.
The target conditional posterior $p(\theta \given \eta)$ is supported on $\mathbb R\times K_\eta$ with $\sup_{\theta,\eta} p(\theta \given \eta) < \infty$, and belongs to a class 
$\calP\classindex$ of positive, continuous conditional probability densities
characterized by the following properties:
\begin{enumerate}[label=(\roman*)]
    \item {\em Spikiness rate (Local log-Lipschitz in $\theta$):} There is a positive increasing function $L_S:(0,\infty)\to(0,\infty)$ such that, 
\[
\big|\log p(\theta \given \eta)-\log p(\theta' \given \eta)\big|
\;\le\; L_S(M)\,|\theta-\theta'|,
\; \forall (\theta,\eta),(\theta',\eta)\in K_M ,\;  M > 0.
\]
\item {\em Conditionals perturbation rate (Local log-Lipschitz in $\eta$):} Let $F_p(\theta \given \eta)$ denote the conditional CDF corresponding to $p(\theta \given \eta)$. There is a positive increasing function  $L_{CP}:(0,\infty)\to(0,\infty)$ such that, 
\[
\begin{aligned}
\big|\log p(\theta \given \eta)-\log p(\theta \given \eta')\big|
\;&\le\; L_{CP}(M) \, \|\eta-\eta'\|_1 ,
\; \forall (\theta,\eta),(\theta,\eta')\in K_M,\;  M > 0,\\
\big|F_p(\theta \given \eta)-F_p(\theta \given \eta')\big|
\;&\le\; L_{CP}(M) \, \|\eta-\eta'\|_1 ,
\; \forall (\theta,\eta),(\theta,\eta')\in K_M,\;  M > 0.
\end{aligned}
\]
\item {\em Quantile growth rate:}
Let $\Phi$ being the $N(0,1)$ CDF and $F^{-1}_p(u \given \eta)$ for $u \in [0,1]$ is the conditional quantile map for $p(\cdot \mid \eta)$ (with respect to the default {\em Unif}$[0,1]$ base). Then the conditional quantile map with respect to $N(0,1)$ base, defined as 
\begin{equation}\label{eq:qtldefn}
T_p(\eta,z)=F_p^{-1}\!\big(\Phi(z)\given \eta\big),\quad z\in\mathbb R,
\end{equation}
is continuous in $\eta$, strictly increasing and $C^1$ (continuously differentiable) in $z$, and there exist constants $A_d,B_d \geq 0$
such that
\[
\sup_{\eta \in K_\eta } |\log \partial_z T(\eta,z)| \le H(z) := A_d + B_d|z| \;\forall z \in \mathbb R.
\]
\item   {\em Tail thinness bound:} {There exist constants $\alpha > 0$, $c > 0$ such that $p(\theta\given\eta)\ge c\,e^{-\alpha \theta^2}$ for all $(\theta,\eta) \in \mathbb R \times K_\eta$.}
\end{enumerate}
\end{assumption}

Instead of studying the properties of variational conditional flows for a specific density $p(\theta \given \eta)$, we assume $p(\theta \given \eta)$ to satisfy Assumption \ref{asm:mono}, enabling studying the entire class of conditional families $\calP\classindex$ characterized by the spikiness, conditional perturbation rate, quantile growth, and thinness bound. This is akin to how Hölder or
Sobolev classes provide functional envelopes for smoothness and are used to derive the theory of functional estimation for these classes rather than individual functions.

Assumptions \ref{asm:mono}$(i)$--$(ii)$ specify that the log conditional density is local Lipschitz on compacts, with respect to both $\theta$ and $\eta$.
The log-Lipschitz modulus \(L_S(M)\) controls how spiky the conditional density is with respect to the observation variable $\theta$.
 The modulus \(L_{CP}(M)\)
controls how smoothly the conditional law varies with the conditioner \(\eta\).
Assumption $(iii)$ is a derivative bound, with the parameters $A_d$ and $B_d$
governing the global spread of the distribution via controlling the conditional quantile map. Continuity of $T_p$ in $\eta$, along with compactness of $K_\eta$ imply a uniform bound on the conditional median, i.e., 
\begin{equation}\label{eq:median}
    M_0 := \sup_{\eta \in K_\eta} \abs{T_p(\eta,0)} < \infty.
\end{equation} 

From the median and derivative bounds, we immediately have the following bound on the conditional quantile map: 
\begin{equation}\label{eq:qtl}
|T_p(\eta,z)| \;\leq\; G(z) := M_0 + |z|e^{A_d + B_d|z|} \quad \forall (\eta,z).
\end{equation}
Thus, we allow the quantile map to grow up to exponentially faster than the base (Gaussian) quantile, enabling it to model densities with thicker tails than the Gaussian.

The triplet \((L_S,L_{CP},G)\) thus describes different aspects of the conditional distributions --- respectively, local
sharpness in \(\theta\), sensitivity across \(\eta\), and quantile growth relative to the
Gaussian base. Table \ref{tab:envelopes} provides the expressions of $L_S(M), L_{CP}(M)$ and $G(M)$ as $M$ grows for common families of conditional distributions
whose parameters (e.g., location, scale, shape) depend smoothly ($C^1$) on the conditioning variable 
\(\eta\). These rates are derived in Section \ref{sec:proofrates} of the Supplementary Materials. We see that the conditions of Assumption \ref{asm:mono}(i) -(iii) are satisfied for a wide variety of distributions, Gaussian, Laplace, Generalized Gaussian, and their finite mixtures. Thicker-tailed families like Cauchy or Student's-t have a sharper quantile growth with respect to the Gaussian base and do not satisfy Assumption \ref{asm:mono}$(iii)$. If the goal is to approximate such very heavy-tailed densities with normalizing flows, one can simply use a thicker-tailed base distribution than $N(0,1)$, and similar theoretical results as we derive here can be obtained.
\begin{table}[!t]
\centering
\footnotesize
	\resizebox{\columnwidth}{!}{
	\begin{threeparttable}
\begin{tabular}{l l l l}
\noalign{\hrule height 1.5pt}
\textbf{Distribution} &
\makecell[l]{\textbf{Spikiness}\\$L_S(M)$} &
\makecell[l]{\textbf{Conditional Perturbation Rate}\\$L_{CP}(M)$} &
\makecell[l]{\textbf{Quantile growth}\\$G(M)$} \\
\noalign{\hrule height 1.5pt}
Normal:
$N(\mu(\eta), \sigma^2(\eta))$ &
$O(M)$ &
$O(M^2)$ &
$O(M)$ \\
\midrule
Laplace:
$\mathrm{Lap}(\mu(\eta), b(\eta))$ &
$O(1)$ &
$O(M)$ &
$O(M^2)$ \\
\midrule
Generalized Gaussian:\\
$\mathrm{GGD}(\mu(\eta), \alpha(\eta), 1 \leq \beta(\eta) \leq 2)$ &
$O(M^{\bar\beta - 1})$,  where  $\bar \beta = \sup_\eta \beta(\eta)$ &
$O(M^{\bar\beta} \log M)$ &
$O(M^{2/\underline\beta})$, where $\underline \beta = \inf_\eta \beta(\eta)$\\
\midrule
Cauchy: $C(\mu(\eta), \gamma(\eta))$ &
$O(1)$ &
$O(1)$ &
$O(M e^{M^2 / 2})$ \\
\midrule
Student's $t$: 
$t_{\mu(\eta), s(\eta), \nu(\eta)}$ &
$O(1)$ &
\begin{tabular}[l]{@{}l@{}}$O(\log M)$ if $\nu$ depends on $\eta$,\\ $O(1)$ otherwise\end{tabular} &
\begin{tabular}[l]{@{}l@{}}$O\!\left((M e^{M^2/2})^{1/\nu_{\min}}\right)$,\\ where $\nu_{\min} := \inf_{\eta \in K_\eta} \nu(\eta) > 0$\end{tabular} \\
\midrule
Finite mixtures of the above &
\multicolumn{1}{l}{$\max$ of component rates} &
\multicolumn{1}{l}{$\max$ of component rates} &
\multicolumn{1}{l}{$\max$ of component growths} \\
\noalign{\hrule height 1.5pt}
\end{tabular}
\caption{Tail orders of spikiness $L_S(M)$, Conditional Perturbation Rate $L_{CP}(M)$, and Quantile Growth $G(M)$ for common distribution families. Distribution parameters are $C^1$ in $\eta$. Quantile growths are with respect to $N(0,1)$ base.}
\label{tab:envelopes}
\end{threeparttable}
}
\end{table}

The following lemma shows that
Assumption \ref{asm:mono}$(i)$--$(iii)$ leads to (local) Lipschitz bounds on the conditional median and
the log-derivative of the conditional quantile. These bounds are key quantities determining the approximation rates. All proofs are in the Supplement. 
\begin{lemma}\label{lem:lips} Let $p \in \calP\classindex$ and $T_p$ denote the corresponding conditional quantile function with a Gaussian base as defined in (\ref{eq:qtldefn}). Then on $K_\eta$, the conditional median $T_p(\eta,0)$ is $\mathsf L_a$-Lipschitz and on $(z,\eta) \in K_M$, the log-derivative $\log \partial_z T_p$ is $\mathsf L_b(M)$-Lipschitz,
where

\begin{equation}\label{eq:lips}
\begin{aligned}
\mathsf L_a &: = c^{-1}e^{\alpha M_0^2}\,L_{CP}(M_0) \, \\
\mathsf L_b(M)
\, & : =\max\Bigl\{M+e^{H(M)}L_S\!\big(G(M)\big),\ L_{CP}\!\big(G(M)\big)\bigl(1+c^{-1}e^{\alpha G(M)^2}L_S\!\big(G(M)\big)\bigr)\Bigr\}.
\end{aligned}
\end{equation}

\end{lemma}

Finally, Assumption \ref{asm:mono}$(iv)$ imposes the Gaussian-type lower tail bound $p(\theta\given\eta)\ \ge\ c\,e^{-\alpha\theta^2}$ for all $(\theta,\eta)\in\mathbb R\times K_\eta$
to rule out pathologically thin tails (as the Gaussian tail itself is quite thin). In VI, the candidate $q(\theta\given\eta)$ is generally required to have mass wherever the target density has mass. If the target density is too thin at the tails, it may require a highly irregular variational 
approximant.
The lower bound assumption prevents this. In Supplemental Section \ref{sec:asmver} we briefly discuss how the assumptions can be verified when applying NeVI-Cut.

Our second assumption is on the variational class of conditional normalizing flows.

\begin{assumption}[Conditional Normalizing Flow Classes]\label{asm:pm}
For integers $m\ge1$, let
\[
\mathcal F_{m}\;:=\;\{\,T_\vartheta:K_\eta\times\mathbb R\to\mathbb R\;:\;\vartheta\in\Theta_{m}\,\}
\]
where $\Theta_{m}\subset \mathbb R^m$ is compact, and $m$ represents the number of free parameters defining the map $T_\vartheta$, reflecting the flow complexity.
Assume the following:

\medskip
\noindent
\emph{$(i)$ Regularity and monotonicity.}
For every $m$ and $\vartheta\in\Theta_{m}$, $T_\vartheta(\eta,z)$ is continuous in $(\eta,z,\vartheta)$, is strictly increasing and $C^1$ in $z$, with $\partial_z T_\vartheta(\eta,z)$ being continuous in $\vartheta$.

\noindent

\noindent
\emph{$(ii)$ Global
linear envelopes.}
There exist positive constants $M^*, A^*, B^*$
such that
\[
\sup_{m\ge1}\ \sup_{\vartheta\in\Theta_{m}}\ \sup_{\eta\in K_\eta}\ |T_\vartheta(\eta,0)|\ \le\ M^*,
\quad
\sup_{m\ge1}\ \sup_{\vartheta\in\Theta_{m}}\ \sup_{(\eta,z)\in K_\eta\times\mathbb R}\ |\log\partial_z T_\vartheta(\eta,z)|
\ \le\ A^*+B^*|z|.
\]

\noindent
\emph{$(iii)$ Universal approximation.}
For any conditional density $p\in\mathcal P\classindex$
with $T_p$ as in (\ref{eq:qtldefn}), and for every $\varepsilon>0$,
there exists a complexity budget $C(M,\varepsilon)$, not depending on $p$, only on $M,\varepsilon$ and the class $\calP\classindex$, such that  for any $m \geq C(M,\varepsilon)$ there exists a $\vartheta\in\Theta_{m}$ such that
\[
\sup_{\eta\in K_\eta}\big|T_\vartheta(\eta,0)-T_p(\eta,0)\big|\le\varepsilon,
\quad
\sup_{(z,\eta)\in K_M}\big|\log\partial_z T_\vartheta(\eta,z)-\log\partial_z T_p(\eta,z)\big|\le\varepsilon.
\]

\end{assumption}

We will show in subsequent results
that popular conditional normalizing flow specifications satisfy Assumption \ref{asm:pm}. First, we provide a more general result on KLD approximation rates for any class of conditional normalizing flows satisfying Assumption \ref{asm:pm}. 

\begin{theorem}[Uniform KLD rates of conditional flows]\label{thm:ckl-explicit} Let $p(\theta \given \eta)$ belong to class of conditional densities $\calP\classindex$. 
Let $q_T(\theta \given \eta)$ be a collection of conditional densities corresponding to $\theta \given \eta = T(\eta,Z)$, $Z\sim N(0,1)$, and $T \in \calF_m$ for some class of conditional flows $\calF_m$ satisfying Assumption \ref{asm:pm} and with $M^* > M_0, A^* > A_d, B^* > B_d$.
Define $\widetilde G(M) := M^* + Me^{A^* + B^*M}$ for $M>0$. Then for any $\varepsilon \leq$ some $\varepsilon_0 < 1$ and $$m \geq C^*(\varepsilon):=C\left(R\sqrt{-\log \varepsilon},\frac{\varepsilon}{6\left(1+\widetilde G(R\sqrt{-\log \varepsilon})L_S(\widetilde G(R\sqrt{-\log \varepsilon}))\right)}\right),$$ there exists $T\in\mathcal F_m$
such that $\sup_{\eta \in K_\eta} \mathrm{KL}\,\!\big(q_T(\cdot\given\eta)\,\|\,p(\cdot\given\eta)\big)
\ \le\ \varepsilon$,
for a constant $R>0$ depending only on
$M^*,A^*,B^*,\alpha,c$. Here $C(\cdot,\cdot)$ is the complexity budget as defined in Assumption \ref{asm:pm} (iii). 
\end{theorem}

Theorem \ref{thm:ckl-explicit} is a result of independent interest. To our knowledge, it offers the first KLD approximation rates for conditional normalizing flows to a fixed target (the density $p$ or the class $\calP\classindex$). We do not assume any sample size growth and the result is uniform over the conditioning variable $\eta$. The result can be used to obtain the KLD error between any target conditional posterior and the variational solution based on conditional normalizing flows. Existing theoretical results on this topic primarily
show that any conditional distribution can be approximated by a sequence of conditional flows in the weak convergence metric \citep{huang2018neural,papamakarios2021normalizing,battaglia2025amortising}. \citet{ishikawa2023universal} gives universality in the total-variation (TV) metric, without rates. However, KLD is a stronger notion of convergence (it implies weak and TV convergence). The existing results do not provide guarantees of KLD convergence, which are needed to study the variational solution that minimizes the KLD loss. \citet{draxler2022whitening} rate-bound the KLD of a fixed distribution to a Gaussian, not to an arbitrary target.  \citet{draxler2024universality} prove distributional universality
in a surrogate loss-improvement functional. Closer in flavor to our result is the deterministic KL
approximation theory for transport maps of \citet{baptista2025approximation}. However, they consider generic transport maps, not tied to the common flow classes used in practice. Also, all of the above concerns unconditional densities, not conditional ones where the notion of convergence needs to be uniform over the conditioner. Our
Theorem~\ref{thm:ckl-explicit} instead gives these uniform rates, and can
be applied for popular conditional flow architectures (RQ-NSF, UMNN), as we present next.

\subsection{KLD approximation rates of Neural Spline Flows}\label{sec:thnsf}
We first consider Rational Quadratic Neural Spline Flows \citep[RQ-NSF,][]{durkan2019neural}. RQ-NSF specify the conditional pushforward map $\theta = T(\eta, z)$ flexibly as a monotone piecewise rational-quadratic function of $z$  whose knot heights, widths, and interior derivatives are modeled by neural networks of the conditioner $\eta$.
Fix $M>0$ and an integer $K\ge 1$. Set $w:=M/K$ and define knots
\(
z_j:=-M+jw,\, j=0,1,\dots,2K,
\)
so $z_K=0$ and $[-M,M]=[z_0,z_{2K}]$ is partitioned into $2K$ bins.
Let $H_c(t):=A_c+B_c t$ for fixed clipping constants $A_c > 0$ and $B_c > 0$. Let the conditioner $\psi_\vartheta:K_\eta\to\mathbb R^{2K+2}$
be a ReLU neural network that outputs $(\tilde u(\eta),v(\eta))$ where $\tilde u(\eta)\in\mathbb R$ and $v(\eta)=(v_0(\eta),\dots,v_{2K}(\eta))\in\mathbb R^{2K+1}$.
Let
\(
y_K(\eta):=\clip(\tilde u(\eta),-M^*,M^*).
\)
Define clipped knot log slopes and knot slopes by
\[
\beta_j(\eta):=\clip\bigl(v_j(\eta),-H_c(|z_j|),H_c(|z_j|)\bigr),\; s_j(\eta):=\exp(\beta_j(\eta)),\; j=0,\dots,2K.
\]
For each bin $[z_i,z_{i+1}]$ define $\bar z_i:=\tfrac12(z_i+z_{i+1})$ and
\[
\Delta_i(\eta):=\exp\,\!\left(\tfrac12\bigl(\beta_i(\eta)+\beta_{i+1}(\eta)\bigr)\right),
\;
h_i(\eta):=w\,\Delta_i(\eta),
\; i=0,\dots,2K-1.
\]
Define all knot values by integrating heights away from the mid-point on both sides:
\[
y_{i+1}(\eta):=y_i(\eta)+h_i(\eta)\;\text{for }i=K,\dots,2K-1,
\;\;
y_{i-1}(\eta):=y_i(\eta)-h_{i-1}(\eta)\;\text{for }i=K,\dots,1.
\]
On a bin $[z_i,z_{i+1}]$ with $\Delta_i:=h_i/w$ and $t:=(z-z_i)/w\in[0,1]$, the monotone rational quadratic formula \citep{durkan2019neural} gives
\begin{equation}\label{eq:rqnsf}
\begin{aligned}
T(\eta,z) &=y_i+h_i\,
\frac{\Delta_i t^2+s_i t(1-t)}{\Delta_i+(s_{i+1}+s_i-2\Delta_i)t(1-t)}.
\end{aligned}
\end{equation}

Outside of $[-M,M]$, the function is extended with linear tails $z\mapsto y_0+s_0(z-z_0)$ for $z\le z_0$ and $z\mapsto y_{2K}+s_{2K}(z-z_{2K})$ for $z\ge z_{2K}$. This construction is a simplified RQ-NSF for theoretical tractability. The splines in \cite{durkan2019neural} also model the widths and heights $h_i(\eta)$ by neural networks. 
The next theorem gives uniform KLD rates of RQ-NSF.

\begin{theorem}[Uniform KLD rate, conditional RQ-NSF]\label{th:rqs}
Let $p\in\mathcal P(L_S,L_{CP},A_d,B_d,c,\alpha)$ and let $q_T$ be the law of $T(\eta,Z)$,
$Z\sim N(0,1)$, with $T$ as in \eqref{eq:rqnsf} using clip $H_c(t)=A_c+B_ct$, thresholds
$M^{*}>M_0,\ A_c>A_d,\ B_c>B_d$, and bin-width upper bound $w_0>0$. The class then satisfies
Assumption~\ref{asm:pm}.
Let $R$ be the constant of Theorem~\ref{thm:ckl-explicit}. For $\varepsilon\in(0,1)$ put $M:=R\sqrt{-\log\varepsilon}$,
$\widetilde L_S(M):=ML_S(M)$, $\widetilde G(M):=M^{*}+Me^{A^{*}+B^{*}M}$, and
\[
\begin{aligned}
A:&=\frac{\exp(2H_c(M))\,\widetilde L_S(\widetilde G(M))\,(\mathsf L_a\vee\mathsf L_b(M))\operatorname{diam}_\eta}{\varepsilon},\\
K_*:&=\Bigl\lceil\tfrac{C_r\mathsf L_b(M)M\exp(2H_c(M))\widetilde L_S(\widetilde G(M))}{\varepsilon}\Bigr\rceil\vee\lceil M/w_0\rceil .
\end{aligned}
\]
Then some RQ spline flow with $2K_*$ bins has $m\le C_r(2K_*+2)A^{d_\eta}\log A$ parameters and
$\sup_{\eta\in K_\eta}\mathrm{KL}(q_T(\cdot\mid\eta)\,\|\,p(\cdot\mid\eta))\le\varepsilon$, where
$C_r$ depends only on $d_\eta$.
\end{theorem}

The Theorem quantifies the upper bound on the number of parameters needed to achieve a certain uniform KLD rate $\varepsilon$. Both $K_*$ and $A$ are increasing in the two Lipschitz constants $L_S$ and $L_{CP}$  and the quantile growth rate $G$ (via $\mathsf L_a$ and $\mathsf L_b(M)$, see Lemma \ref{lem:lips}). This is expected as growth of each of these adds complexity to the target posterior. The upper bound for
a given error rate grows exponentially with the dimension $d_\eta$. This is not surprising and aligns with the complexity theory of ReLU neural networks \citep{yarotsky2017error}, which grows exponentially with the dimension of the function domain to achieve a specified approximation threshold. In Section \ref{sec:umnn}, we provide a similar result (Theorem \ref{th:umnn}) for {\em Unconstrained Monotonic Neural Networks} \citep[UMNN,][]{wehenkel2019unconstrained}, another popular class of conditional normalizing flows. The rates provide similar insights.

\subsection{Theory of NeVI-Cut}\label{th:nvct}

We now establish error rates of the NeVI-Cut estimate approximating a target cut-posterior using the general theory of Section \ref{sec:klrate}. We first provide guarantees on existence and KLD approximation rates of the NeVI-Cut estimate obtained as the variational solution based on the loss (\ref{eq:cutvicond}), which integrates the conditional KLD over the exact cut distribution $p(\eta \given D_1)$ of the upstream parameters. The practical scenario where we only observe samples from $p(\eta \given D_1)$ is presented subsequently. 
We hide the conditioning datasets as they remain fixed for all the theory and simply use $\pc(\theta,\eta)$, $\pc(\theta \given \eta)$, and $\pc(\eta)$ to denote the joint, conditional, and marginal cut-posteriors. For any conditional distribution $q(\theta \given \eta)$, define the expected conditional KLD  to the conditional cut-posterior as
\begin{equation}\label{eq:kldavg}
R(q)\;=\;\;\int \mathrm{KL}\,\!\big(q(\cdot\given\eta)\,\|\,p(\cdot\given\eta)\big)\,\pc(\eta)\,d\eta.
\end{equation}
Note that $R(q)$ is also the KLD  of the joint density $q(\theta \given \eta) \pc(\eta)$ to the joint cut-posterior $\pc(\theta,\eta)$, i.e., $R(q) = \mathrm{KL}\, \Big(q(\cdot \given \cdot)\pc(\cdot) \,\|\, \pc(\cdot,\cdot)\Big)$.

\begin{corollary}[Integral-based NeVI-Cut]\label{cor:condvarflow}
Let $\pc(\theta, \eta) = \pc(\eta) \pc(\theta \given \eta)$ be the target cut-posterior, where $\pc(\eta)$ is a distribution with support within a compact set $K_\eta \in \mathbb R^{d_\eta}$ and $\pc(\theta \given \eta)$ is in $\calP\classindex$. Let $\calF_m$ be a class of the conditional flows that satisfies Assumption~\ref{asm:pm} for the class $\calP\classindex$ with $M^* > M_0, A^* > A_d, B^* > B_d$, and complexity budget $C(M,\varepsilon)$.
Let $\calQ_m$ be the class of conditional distributions corresponding to $\calF_m$, i.e., 
\[
\mathcal Q_m:=\big\{\,q_T(\cdot\given\eta):=\Law\big(T(\eta,Z)\big),\ Z\sim\mathcal N(0,1)\ :\ T\in\mathcal F_m\,\big\}.
\]
Then, for any small $\varepsilon > 0$, we have the following conclusions:
\begin{enumerate}[label=(\alph*)]
\item For every $m\ge1$, a minimizer \(\hat q_m\in\arg\min_{q\in\mathcal Q_m}R(q)\) exists.
\item Fix any small accuracy $\varepsilon >0$ and let $m \geq C^*(\varepsilon)$
be the budget of Theorem~\ref{thm:ckl-explicit}.
Then
\(
R(\hat q_m)
\;=\;\min_{q\in\mathcal Q_{m}}R(q)
\;\le \varepsilon
\).
\end{enumerate}
\end{corollary}

The result shows that the integral-based NeVI-Cut solution, using the integrated KLD loss (\ref{eq:cutvicond}), approximates the true cut-posterior up to any error rate $\varepsilon$ for a suitably large neural network architecture. For specific choices of flows like RQ-NSF or UMNN, the respective complexities given by Theorem \ref{th:rqs} and Theorem \ref{th:umnn} hold. The result is in the KLD, which is stronger than the total-variation or the weak distributional metric. 

In practice, optimization cannot be done using the exact integration as in (\ref{eq:cutvicond}). The actual NeVI-Cut algorithm uses $N$ draws from $\pc(\eta)$ and optimizes over the sample-averaged loss in (\ref{eq:cutvicondsample}). We now provide the analogous bounds for this actual NeVI-Cut solution. 

\begin{theorem}[Sample-based NeVI-Cut]\label{th:mcmc}
Consider the setting of Corollary \ref{cor:condvarflow} for a class of conditional flows $\calF_m$ where $m \geq C^*(\varepsilon)$.
Let $\{\eta_i\}_{i=1}^N$ be iid samples from $\pc(\eta)$ or be samples from a stationary ergodic Markov chain with
invariant distribution $\pc(\eta)$.
Let
\[
R_{N}(q)=\frac1N\sum_{i=1}^N \mathrm{KL}\,\!\big(q(\cdot\given\eta_i)\,\|\,p(\cdot\given\eta_i)\big), \quad \text{for} \quad q\in\mathcal Q_m
\]
be the loss used in NeVI-Cut. 
Then for each $N$, the minimizer
\(
\hat q_{N,m}\in\arg\min_{q\in\mathcal \calQ_m} R_{N}(q)
\)
exists. Moreover, for any small $\varepsilon > 0$ and $m \geq C^*(\varepsilon)$, we have
    $\limsup_{N\to \infty} R(\hat q_{N,m}) \le \varepsilon$. 
\end{theorem}

Theorem \ref{th:mcmc} provides an existence guarantee and an error bound in the sample-averaged KLD  for the actual NeVI-Cut solution using the sample-based loss. The key technical step in proving the theorem is establishing a uniform version of Birkhoff's ergodic law of large numbers showing that $R_N$ converges to $R$ uniformly in $q$.
The result assumes only that the samples are from a stationary ergodic Markov chain, which covers the iid case (Monte Carlo samples) but also the more realistic case where the samples are MCMC posterior samples from an upstream Bayesian analysis. 

\section{Applications to Real Datasets}\label{sec:realdata}

We present analyses of four real datasets from four scientific applications
that vary widely in terms of dimensionality of both upstream and downstream parameters, availability of upstream data and model, and scientific application. For each dataset, at least one existing cut algorithm can be run to provide a reference cut posterior, so real-data analyses alone suffice to benchmark both the accuracy and the speed of NeVI-Cut. We also conducted a set of simulation studies, which are presented in Section \ref{sec:sim} of the Supplementary Materials. 

The set of existing cutting feedback methods used for comparison varied by dataset, depending on their compatibility and scalability to that dataset's features. For presenting the results here, we report NeVI-Cut, one sampling-based algorithm (the nested MCMC, serving as the gold standard against which all the other algorithms are compared), one other neural variational algorithm \citep[SMI-Cut, the semi-modular inference, or SMI method of][which runs cut-Bayes as a limiting case and also uses normalizing flows]{carmona2022scalable}, and one variational algorithm with parametric variational family \citep[Parametric-Cut, which uses the parametric VI cut of][when upstream data and model are available, or a parametric version of NeVI-Cut when only upstream samples are available]{yu2023variational}. Table \ref{tab:realdata_summary} summarizes the main benchmarking results between these four algorithms across the four applications. In Section~\ref{app:real_data} of the Supplementary Materials, we provide additional comparisons and results for each application, along with details on data availability, preprocessing, model specifications, and computing platform.

\begin{table}[!t]
\centering
\caption{Comparison of dataset features and performance in real-data analyses.  `h' and `m' in the Time columns respectively denote hours and minutes.}
\label{tab:realdata_summary}
\resizebox{\textwidth}{!}{
\begin{tabular}{lcccccccccc}
\noalign{\hrule height 1.5pt}
\multirow{3}{*}{\textbf{Application}} 
& \multicolumn{3}{c}{\textbf{Dataset Features}} 
& \multicolumn{7}{c}{\textbf{Performance}} \\
\cmidrule[1.5pt](r){2-4} \cmidrule[1.5pt](l){5-11}
& \multirow{2}{*}{dim(${\eta}$)} 
& \multirow{2}{*}{dim(${\theta}$)} 
& \multirow{2}{*}{\makecell{\textbf{Upstream Data}\\\textbf{Availability}}} 
& \textbf{Nested MCMC} 
& \multicolumn{2}{c}{\textbf{NeVI-Cut}} 
& \multicolumn{2}{c}{\textbf{SMI-Cut}} 
& \multicolumn{2}{c}{\textbf{Parametric-Cut}}\\
\cmidrule[1.5pt](lr){5-5} \cmidrule[1.5pt](lr){6-7} \cmidrule[1.5pt](l){8-9} \cmidrule[1.5pt](l){10-11}
& 
& 
& 
& \textbf{Time} 
& \textbf{WD} 
& \textbf{Time} 
& \textbf{WD} 
& \textbf{Time} 
& \textbf{WD} 
& \textbf{Time}\\
\noalign{\hrule height 1.5pt}
1. HPV 
& 13 
& 2 
& Yes
& 1.47m 
& \textbf{0.12}
& 1.22m 
& 0.17 
& 12.85m 
& 0.78
& 1.10m\\
\noalign{\hrule}

2. LP 
& 9503 
& 574 
& Yes
& NA 
& 0.28$^*$
& 5.17h
& \textbf{0.27$^*$}
& 8.08h
& NA
& NA\\
\noalign{\hrule}

\multicolumn{9}{l}{3a. COMSA-Moz (Neonates)} \\

\quad EAVA
& & &  & 5.4m & \textbf{0.075} & 4.0m & NA & NA & 0.31 & 0.07m \\

\quad InSilicoVA
& 20 & 5 & No & 5.7m & \textbf{0.051} & 4.0m & NA & NA & 0.43 & 0.07m \\

\quad InterVA
& & & & 5.2m & \textbf{0.075} & 3.8m & NA & NA & 0.43 & 0.06m\\

\noalign{\hrule}

\multicolumn{9}{l}{3b. COMSA-Moz (Children)} \\ 

\quad EAVA & & & & 9.6m & \textbf{0.082} & 6.2m & NA & NA & 0.34 & 0.09m \\ 

\quad InSilicoVA & 56 & 8 & No & 10.6m & \textbf{0.072} & 5.9m & NA & NA & 0.21 & 0.08m \\ 

\quad InterVA & & & & 11.3m & \textbf{0.076} & 5.9m & NA & NA & 0.15 & 0.06m \\ 

\noalign{\hrule}

4. GBD 
& 42
& 2737 
& No
& 330h 
& \textbf{0.005}
& 35.0h 
& NA 
& NA 
& 0.025
& 1.82h\\
\noalign{\hrule height 1.1pt}
\end{tabular}
}

\vspace{0.5em}
\begin{minipage}{.99\textwidth}
{\footnotesize
\textit{Note 1:} WD represents the 2-Wasserstein distance using the \href{https://cran.r-project.org/package=transport}{\texttt{transport}} R package. In HPV, COMSA-Moz, and GBD, WD evaluates the NeVI-Cut, SMI-Cut, and Parametric-Cut posteriors relative to the Nested MCMC posterior.
In the LP application, the $*$ next to the WD numbers indicates reporting of the average predictive WD relative to the hold-out data.}
\end{minipage}

\vspace{0.5em}
\begin{minipage}{.99\textwidth}
{\footnotesize
\textit{Note 2:} The SMI-Cut jointly models upstream and downstream. NeVI-Cut is only a downstream analysis using upstream posterior samples. The times are thus not directly comparable.}
\end{minipage}

\vspace{0.5em}
\begin{minipage}{.99\textwidth}
{\footnotesize
\textit{Note 3:} In Table \ref{tab:realdata_summary}, for COMSA-Moz, the Parametric-cut is based on the logistic-normal distribution. Similar results for another version based on the Dirichlet distribution are provided in Table~\ref{tab:wasserstein_time}.}
\end{minipage}

\vspace{0.5em}
\begin{minipage}{.99\textwidth}
{\footnotesize
\textit{Note 4:} NeVI-Cut uses a fully autoregressive RQ-NSF for the HPV, COMSA-Moz, and GBD analyses, and a coupling RQ-NSF for the LP analysis. See Supplement for details.}
\end{minipage}
\end{table}

\paragraph*{Prevalence of Human Papillomavirus in Cervical Cancer.} 
First, we revisit the epidemiological analysis exploring the association between human papillomavirus (HPV) prevalence and cervical cancer incidence \citep{maucort2008international}. This is a common benchmarking dataset in the cut-Bayes literature \citep{plummer2015cuts, yu2023variational, carmona2022scalable} due to the availability of the upstream data, and modest parameter dimensionality (dim$(\eta)=13$, dim$(\theta)=2$), thereby making it feasible to run most of the cut-Bayes methods. The upstream data ($D_1$) consists of counts  $z = (z_1, \ldots, z_{13})^\top$, where $z_i$ is the number of individuals infected with high-risk HPV among $n_i$ individuals in country $i$. The upstream model is $z_i \con \gamma_i \sim \text{Binomial}(n_i, \gamma_i)$ with upstream parameters $\eta=\{\gamma_i\}$, where $\gamma_i$ is the true HPV prevalence in country $i$. The downstream data ($D_2$) is $w = (w_1, \ldots, w_{13})^\top$, the observed cervical cancer cases, modeled as $w_i \con \theta, \gamma_i \sim \text{Poisson} \left(T_i \exp(\theta_1 + \theta_2 \gamma_i)\right)$ with downstream parameters $\theta = (\theta_1, \theta_2)^\top$, where $T_i$ denotes the follow-up duration (in years) for country $i$, and $\theta_1$ and $\theta_2$ are the regression parameters of interest.

\begin{figure}[!t]
  \centering
  \begin{subfigure}[b]{0.35\textwidth}
    \centering
    \includegraphics[width=\linewidth]{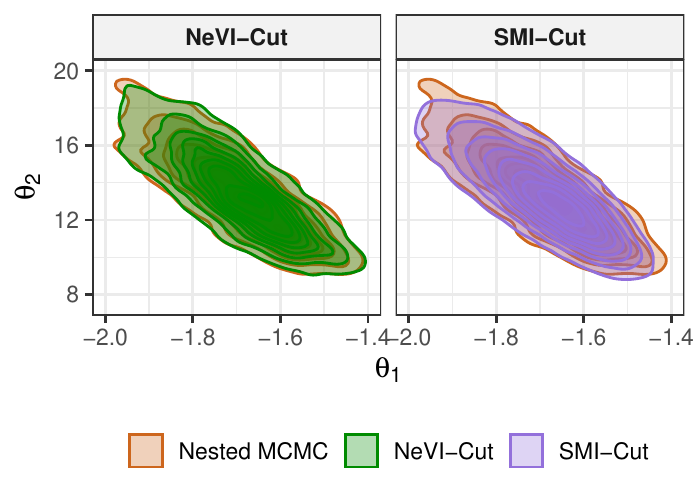}
    \caption{HPV Prevalence}
    \label{fig:main_hpv}
  \end{subfigure}
  \hfill
  \begin{subfigure}[b]{0.45\textwidth}
    \centering
    \includegraphics[width=\linewidth]{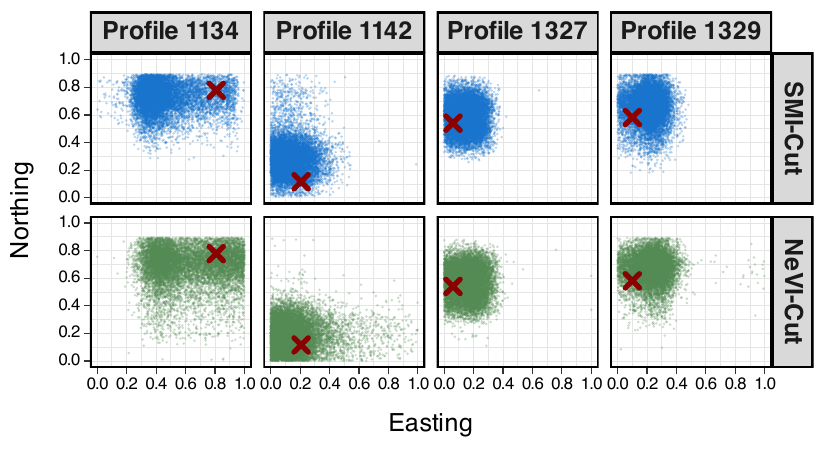}
    \caption{Linguistic Profile}
    \label{fig:main_lp}
  \end{subfigure}
\hfill
\vspace{0.25cm}
  \begin{subfigure}[b]{0.9\textwidth}
    \centering
    \includegraphics[width=\linewidth]{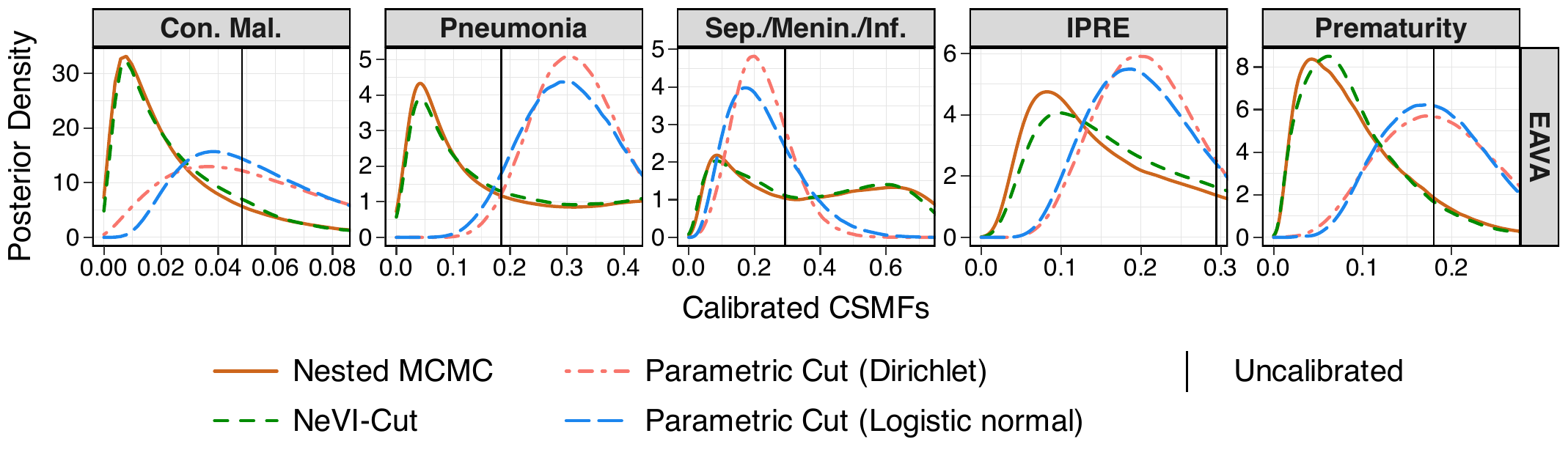}
    \caption{Child Mortality Surveillance in Mozambique}
    \label{fig:main_comsa}
  \end{subfigure}
  \hfill
  \vspace{0.25cm}
  \begin{subfigure}[b]{\textwidth}
    \centering
    \includegraphics[width=\linewidth]{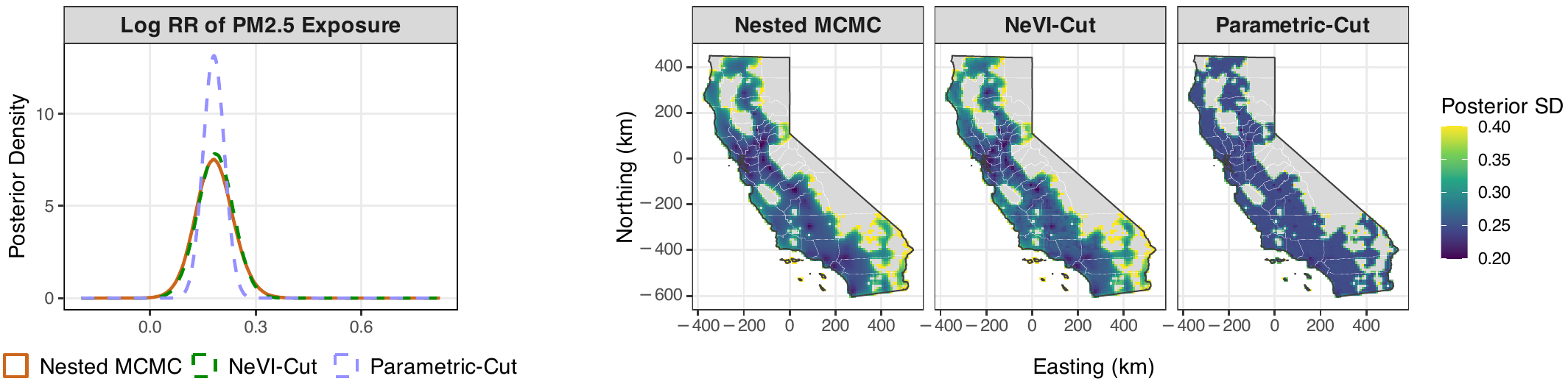}
    \caption{Global Burden of Disease in California}
    \label{fig:main_gbd}
  \end{subfigure}
  \caption{A comparison of the cut posterior distribution of the downstream parameter $\theta$ obtained using different methods across four real data applications: $(a)$ exploring the association between human papillomavirus (HPV) prevalence and cervical cancer incidence, $(b)$ inferring locations of medieval linguistic profiles, $(c)$ estimating cause-specific mortality fractions in Mozambique using verbal autopsy and computer-coded algorithms, and $(d)$ analyzing association of air pollution with cardiovascular disease mortality in California.}
  \label{fig:main}
\end{figure}

Figure \ref{fig:main_hpv} compares the bivariate cut posteriors of $\theta$ from the gold-standard nested MCMC with NeVI-Cut and SMI-Cut. The SMI-Cut posterior is largely accurate but misses a notch in the top-left. The NeVI-Cut posterior captures this notch better and is overall nearly identical to the nested MCMC posterior, demonstrating its ability to capture both large- and fine-scale features of the posterior surface. Table \ref{tab:realdata_summary}
reveals that NeVI-Cut achieves considerably lower Wasserstein distance to the nested MCMC than the SMI cut-posterior, while also being faster. The parametric-Cut, while being faster, has much worse Wasserstein distance to the nested MCMC. The larger comparison
with various other cut-Bayes algorithms like the {\em OpenBUGS-Cut}, and the {\em tempered-Cut algorithm} of \cite{plummer2015cuts} can be found in Appendix \ref{sec:hpv}. NeVI-Cut achieves the lowest Wasserstein distance amongst all.

\paragraph*{Analyzing Locations of Medieval Linguistic Profiles.} Next, we revisit the linguistic-profile (LP) analysis of \citet[Section 6.2]{battaglia2025amortising} where the upstream data and model are also available but which involves a high-dimensional parameter set. The goal is to jointly model medieval charter LPs with observed locations (``anchor'' profiles) and without observed locations (``floating'' profiles) to estimate the locations of floating profiles. Each profile is summarized by a collection of binary records $y_{p,i,f}\in\{0,1\}$ indicating whether a form $f \in \calF$ of item $i \in \calI$ appears in document $p$ at location $x_p$. The locations $x_p$ are known on anchors $A$ and unknown on floats $M$. Presence 
is a thresholded zero-inflated Poisson, with form-usage $\phi_{i,f}$ being a softmax of $B$
shared Gaussian-process basis fields:
\begin{gather*}
\Pr \left( y_{p,i,f}=1 \mid x_p, \zeta_i, \mu_i, \phi_{i,f} \right) = (1-\zeta_i)\big(1-e^{-\mu_i\phi_{i,f}(x_p)}\big),\\
\phi_{i,f} (x) \,\propto \,\exp \Big(a_{i,f}+\textstyle\sum_{b=1}^{B}w_{i,f,b}\,\gamma_b(x)\Big),
\qquad \gamma_b(\cdot)\sim\mathcal{GP}\big(0,k(\cdot,\cdot)\big).
\end{gather*}
The upstream block models anchored profiles where the unknowns are the parameters $\eta=(\mu=\{\mu_i\},\zeta=\{\zeta_i\},a=\{a_i\},W=\{w_{if,b}\},\Gamma=\{\gamma_b\})$ which are shared by
both modules. The downstream block consists of the likelihood for the floating locations which are a function of both $\eta$ and the unknown locations $\theta=X_M=\{x_p\}_{p \,\in \,M}$. 
The upstream module learns $\eta$ from the anchor profiles with observed locations, and the downstream module
infers locations $\theta$ of floating profiles given the upstream posterior uncertainty
in $\eta$.

The analysis involves over $10,000$ total parameters with dim$(\eta)=9503$ and dim$(\theta)=574$, and the nested MCMC is computationally prohibitive. Instead, we hold out the locations of $40$ of the $120$ anchored profiles for out-of-sample evaluation of NeVI-Cut and SMI-Cut. 
The SMI-Cut results are reproduced utilizing the code available on \href{https://github.com/llaurabatt/amortised-variational-flows}{GitHub}.
NeVI-Cut can utilize upstream posterior samples of $\eta$ available from any algorithm. Here we simply reuse the ones provided by SMI-Cut. 

The hold-out performance using predictive Wasserstein distance
is summarized in Table~\ref{tab:realdata_summary}, and shows that
both SMI-Cut and NeVI-Cut are comparable. Other metrics like root mean squared error and energy scores (presented in Table \ref{tab:LP diagnostic table} of the Supplement) are also nearly identical between the two methods, as are most of the profile location cut posteriors (see Figure \ref{fig:main_lp} for a sample of these).
Detailed analyses are in Supplementary Section \ref{app:LP}. 

This illustration, in a setting where the upstream data and model are available, allowing implementation of SMI-Cut, shows that NeVI-Cut performs at par with this state-of-the-art existing neural variational method for cut-Bayes even in high-dimensional scenarios.

\paragraph*{Cause-Specific Mortality Fractions in Mozambique.} Our next two applications concern datasets where only posterior samples are available but not the upstream data and model. This is the main focus of NeVI-Cut, since most existing Cut-VI approaches are incompatible with this setting. The first application is a global health initiative aimed at estimating cause-specific mortality fractions (CSMFs) among neonatal (0-27 days) and children (1-59 months) deaths in Mozambique. The Countrywide Mortality Surveillance for Action (COMSA) program in Mozambique \citep[\href{https://comsamozambique.org/}{COMSA-Moz};][]{Macicame2023} employs verbal autopsy (VA), a WHO-standardized caregiver interview process to surveil individual-level mortality in low- and middle-income countries (LMICs). Following this, VA records are processed through computer-coded verbal autopsy (CCVA) algorithms to ascertain individual-level COD (VA-COD).
However, these CCVA algorithms misclassify the cause for a large fraction of deaths, thereby requiring adjustments of their algorithmic bias to estimate CSMFs from VA-COD. This is accomplished through the method {\em VA-calibration}, which calibrates the CSMF estimates by using estimated CCVA confusion (misclassification) matrices \citep{datta2020,fiksel2022,pramanik2025modeling}. 

For $C$ causes, the upstream analysis uses a secondary labeled dataset ($D_1$) from the \href{https://champshealth.org/}{CHAMPS} project \citep[see][for the exact upstream model specification]{pramanik2025modeling} and provides posterior estimates of the confusion matrix of the algorithm $\eta=\Phi=(\phi_{ij})$, where $\phi_{ij}=\bbP(\text{CCVA predicts cause } j \given \text{True cause is } i)$.
While the CHAMPS dataset is not publicly available, posterior samples of the confusion matrices (segregated by age-group and algorithm) are publicly available for Mozambique \citep{Pramanik2026bmjgh}. Let $\bv=(v_1,\ldots,v_C)^\top$ denote the VA-COD data ($D_2$) from \href{https://comsamozambique.org/}{COMSA-Moz}, where $v_j$ is the number of deaths classified as cause $j$ by a CCVA algorithm.
The downstream VA-Calibration model is
$
\bv \con \theta, \Phi \sim \text{Multinomial}(n, \Phi^\top \theta)
$
with downstream parameters $\theta=(p_1,\ldots,p_C)^\top$, where $p_i=\bbP(\text{True cause is } i)$.
The goal is to estimate the calibrated CSMFs $\theta$.

We use cut-Bayes to propagate the upstream uncertainty in estimates of $\eta$ when estimating $\theta$ downstream. This prevents feedback from the unlabeled \href{https://comsamozambique.org/}{COMSA-Moz} downstream data to
the confusion matrix $\eta$, which was estimated based on the gold-standard labeled CHAMPS data.
We compare nested MCMC, NeVI-Cut, two parametric versions of our NeVI-Cut where we replace normalizing flows with parametric distributions (logistic normal and Dirichlet), and an approximate full Bayes (using the samples to construct a parametric prior for $\eta$, see Section \ref{app:comsa} for details). Figure \ref{fig:main_comsa} presents the posteriors of calibrated CSMFs for the five major causes of death in neonates (Congenital malformation, Pneumonia, Sepsis-Meningitis, Intra-partum related events, Prematurity) for one of the CCVA algorithms (EAVA). We see that the true cut posteriors, as estimated by the nested MCMC, have widely varying geometry, often having skewness, multimodality, or both. Each of these posteriors is very well estimated by NeVI-Cut,  whereas the parametric cut-VI methods produce largely unimodal and symmetric cut-posterior estimates, far away from the true cut-posterior. From Table \ref{tab:realdata_summary}, we see that a single analysis using NeVI-Cut takes around 3.8-4 minutes, whereas for nested MCMC it is around 5.5 minutes. This gain may seem small, however, these runs are repeated across three CCVA algorithms, two age groups, and multiple values of hyperparameters. So the 30-40\% time reduction per run for NeVI-Cut leads to considerable total time gained across all the \href{https://comsamozambique.org/}{COMSA-Moz} analysis. More detailed results, including posterior density plots, Wasserstein distances, and timings for all the cut methods, CCVA algorithms, and age groups are presented in Supplementary Section \ref{app:comsa}.

\paragraph*{Association of Air Pollution with Cardiovascular Disease Mortality in California.} 
Our final application is in a high-dimensional setting where the upstream data and model are again unavailable.  We study the population-level association between long-term PM$_{2.5}$ exposure and ischemic heart disease (IHD), accounting for spatial dependence. Posterior draws $\left\{\text{RR}(\delta_i)\right\}_i$ of PM$_{2.5}$ exposure relative-risk curves produced by the upstream {\em Global Burden of Disease 2019 study} are available \href{https://ghdx.healthdata.org/record/ihme-data/global-burden-disease-study-2019-gbd-2019-particulate-matter-risk-curves}{publicly} \citep{gbd2019_pm_risk_curves}.
The downstream data $D_2$ are county-level IHD death counts in California $y_j$, $j=1,\ldots,J$. With log relative risk $z_j = \log\text{RR}(\delta_j)$, county-level covariates $x_j$, and grid-cell-level spatial effects $w_i$, we use a hierarchical Poisson model:
\begin{gather*}
y_j \mid \theta, z_j \ind \operatorname{Poisson}\left(\mu_j\right), \quad
\mu_j = \sum_{i \in A_j} a_{ij}\,
\exp\left\{ \alpha + z_j \beta_{\mathrm{RR}} + x_j^\top \beta + w_i \right\},\\
\theta = \left( \alpha, \beta_{\mathrm{RR}}, \beta, w = \left\{ w_i \right\}, \sigma_w \right), \quad\alpha,\, \beta_{\mathrm{RR}} \sim \mathcal{N}(0,\,5^2), \quad
\beta \sim \mathcal{N}(0,\,5^2 I_p),\\
p(w \mid \sigma_w) \propto
\exp\left\{ -\frac{1}{2\sigma_w^2} \bigg[
\sum_{i \sim i'} (w_i - w_{i'})^2 + \lambda \sum_{i=1}^I w_i^2 \bigg] \right\}, \quad
\sum_{i=1}^I w_i = 0, \quad \sigma_w \sim \text{Half-Normal}(0,1).
\end{gather*}
Here $A_j$ is the set of 10-km grid cells in county $j$, $a_{ij}$ the population offset for cell $i$, and $i \sim i'$ denotes neighboring cells; the $\{w_i\}$ carry an ICAR prior with weak $L_2$ regularization (Supplementary Section \ref{app:gbd}). The upstream parameter is $\eta = z = \{z_j\}$ and the downstream parameter is $\theta$.

We evaluate nested MCMC, NeVI-Cut, and a parametric-Cut (a parametric variant of NeVI-Cut using Gaussian or log-Gaussian variational families for the parameter, as detailed in Supplementary Section \ref{app:gbd}, along with additional results).
The results are presented in Figure \ref{fig:main_gbd} and Table~\ref{tab:realdata_summary}. We first see from Figure \ref{fig:main_gbd} (left) that, for the Poisson regression coefficients, all methods estimate a positive association between IHD and PM$_{2.5}$ risk, as indicated by the slope ($\log$ RR). However, the posteriors from nested MCMC and NeVI-Cut are nearly identical, whereas the parametric cut is overconfident and underestimates the tails. So, unsurprisingly, NeVI-Cut has a considerably lower Wasserstein distance to the nested MCMC than the parametric cut. For the random effects, although the posterior means are largely similar between the three methods (Figure \ref{fig:w_mean_compare}), we see from Figure \ref{fig:main_gbd} (right) that the parametric cut produces a flat posterior standard deviation (sd) map, failing to capture the spatial variation in the random effect sd. The maps from nested MCMC and NeVI-Cut look very similar and capture this variation, particularly the high SD near the boundaries which is expected because of less data in those areas.  

While defining spatial random effects at the grid-cell level enhances spatial resolution, it results in a high-dimensional latent Gaussian field with dim$(\theta)=2737$, making posterior inference from the nested MCMC extremely slow ($330$ hours). The NeVI-Cut significantly mitigates this challenge, achieving a nearly ten-fold reduction in computation time ($35$ hours), without sacrificing the accuracy of estimating this high-dimensional cut-posterior.

\paragraph*{Summary of the real data benchmarking for NeVI-Cut.}
The analyses show NeVI-Cut is accurate, fast, and robust even in high dimensions. In the first two examples (HPV and linguistics), the upstream data and model are available, letting us benchmark NeVI-Cut against state-of-the-art VI cut methods; it matches their accuracy. The upstream data are not publicly available in the other applications (\href{https://comsamozambique.org/}{COMSA-Moz} and GBD), so existing VI cut methods cannot be applied. Using only upstream posterior draws, NeVI-Cut recovers cut-posterior inference nearly indistinguishable from nested MCMC while running several times faster. Where competing parametric VI cut methods struggle with complex-shaped posteriors, NeVI-Cut's normalizing flows mitigate this issue even in settings involving thousands of parameters.

\section{Discussion}
\label{sec:discussion}

NeVI-Cut contributes to the literature on cutting feedback by eliminating the need for upstream data or models, relying solely on upstream posterior samples. NeVI-Cut utilizes neural-network-based normalizing flows to enhance expressiveness over parametric variational families. We showcase multiple real-life applications where NeVI-Cut captures complex shapes of conditional and marginal cut-posteriors, and achieves accuracy comparable to nested MCMC while significantly reducing runtime. We offer publicly available software that requires only the upstream posterior sample along with downstream data and model.

Theoretically, we provide, to our knowledge, the first uniform KLD approximation rates for conditional normalizing flows, which are at the heart of NeVI-Cut. Compared to existing results on universal expressiveness of flows in the weak or TV metric, these results provide stronger conclusions and directly lead to convergence guarantees of variational solutions (e.g., NeVI-Cut) for modeling conditional distributions. Furthermore, the results are in the fixed-data regime, aiming at actual cut-posteriors rather than their large-sample limits. Importantly, they link approximation error to the complexities of neural network architecture and target posterior. These findings hold independent significance, as conditional flows are applied in numerous other contexts, such as semi-modular inference and hyperparameter-amortized Bayesian inference \citep{carmona2020semi,battaglia2025amortising}. Convergence guarantees of these methods can be studied in the future using our general results.

\section*{Acknowledgements}
We acknowledge the use of generative AI to help with the literature review, proof development and writing, algorithmic implementation, and improving writing in some parts. Everything was independently verified by the authors, who take full responsibility for all content.

\clearpage
\pdfpagewidth=597.50787pt
\pdfpageheight=845.04684pt
\paperwidth=597.50787pt
\paperheight=845.04684pt
\textwidth=474.64801pt
\textheight=722.18698pt
\oddsidemargin=-10.84006pt
\evensidemargin=-10.84006pt
\topmargin=-47.84006pt
\headheight=12.0pt
\headsep=25.0pt
\footskip=30.0pt
\marginparwidth=35.0pt
\setlength{\parskip}{0pt}
\setstretch{1.52}
\setlength{\topskip}{\baselineskip}
\setlength\abovedisplayskip{6pt}
\setlength\belowdisplayskip{6pt}
\setlength\abovedisplayshortskip{4pt}
\setlength\belowdisplayshortskip{4pt}
\setlength\jot{2pt}
\makeatletter
\gdef\@title{Supplement to `Neural Variational Cut Posteriors without Upstream Data'}
\gdef\@author{}
\gdef\@date{}
\@maketitle
\makeatother
\renewcommand\thesection{S\arabic{section}}
\renewcommand\theequation{S\arabic{equation}}
\renewcommand\thelemma{S\arabic{lemma}}

\renewcommand\thefigure{S\arabic{figure}}
\renewcommand\thetable{S\arabic{table}}
\renewcommand\thetheorem{S\arabic{theorem}}
\renewcommand\thetheorem{S\arabic{proposition}}
\renewcommand\thelemma{S\arabic{lemma}}
\setcounter{figure}{0}
\setcounter{section}{0}
\setcounter{equation}{0}
\setcounter{table}{0}
\setcounter{lemma}{0}
\setcounter{theorem}{0}

\vskip 1.4cm

\section{Conditional Normalizing Flow Choices}\label{sec:NeVI-Cut Specification}

We explore two specific formulations of the transformation $T_\vartheta$ used in NeVI-Cut, which are detailed below. Additional flow-based transformations are discussed in \cite{papamakarios2021normalizing}.
\subsection{Neural Spline Flows}
\cite{durkan2019neural} introduced Neural Spline Flows (NSF) for modeling $q(\theta \given \eta)$, which implies a normalizing-flow-based transformation. Following the terminology of \citet{papamakarios2021normalizing}, an NSF is characterized by three key components: the choice of flow, the transformer, and the conditioner that parameterizes the transformer. Together, they imply the transformation $\theta = T_\vartheta (\eta, Z)$. \cite{durkan2019neural} proposes the monotonic rational-quadratic (RQ) spline functions as the transformer, and a neural network as the conditioner. The NSF framework corresponding to these choices is denoted by RQ-NSF and has been shown to enhance expressiveness while preserving analytic invertibility.

The choice of flow determines dependencies among $\theta=(\theta_1, \dots, \theta_d)\in \mathbb{R}^d$ while preserving efficient implementation and computation of the Jacobian log-determinant. \cite{durkan2019neural} discussed two commonly used flow choices for RQ-NSF which, as \cite{papamakarios2021normalizing} describes, represent two extremes on a spectrum of possible implementations: autoregressive flow (RQ-NSF(AR)) and coupling flow (RQ-NSF(C)).

Let $Z=(Z_{1},\dots,Z_{d})\in \mathbb{R}^d$ be a base random vector with a tractable density and simple sampling, for example, $Z\sim N(0,I_d)$. Neural Spline Flows
model the conditional distribution $q(\theta\given\eta)$ according to the law
\begin{equation*}
\theta = T_{\vartheta} (\eta,Z) \quad \text{where} \quad T_{\vartheta} = T_{\vartheta_J} \circ \dots \circ T_{\vartheta_1}.
\end{equation*}
We assume $J=1$ for simplicity to explain the construction and consider $\theta = T_{\vartheta_1}(\eta,Z) \equiv T_{\vartheta}(\eta,Z)$. Below, we discuss how components of NSF define $T_{\vartheta}$.

\paragraph{Autoregressive flow.} Define $Z_{<i} = (Z_1,\dots,Z_{i-1})$ with $Z_{<1} = \emptyset$ being the null set. Moreover, let $\tau \left( \cdot ; h \right)$ denote a transformer with parameter $h$, and $c_{\rho} \left( \cdot \right)$ denotes a conditioner with parameter $\rho$. Given a conditioning variable $\eta$ and following notations from \cite{papamakarios2021normalizing}, the NSF with autoregressive flow (NSF (AR)) transforms $Z$ to $\theta$ as
\begin{equation}\label{eq: RQNSF AR}
\theta_i = \tau \left( Z_{i} ; h_i \right), \quad h_i = c_{\rho_i} \left(Z_{<i},\eta \right) \quad i=1,\dots,d.
\end{equation}

Specifically, RQ-NSF (AR) variant sets the transformer $\tau(\cdot;h_i)$ to be the monotonically increasing rational-quadratic spline (RQS) function with $K$ bins, and the conditioner $c_{\rho_i}$ as the neural network.
Together, the flow choice \eqref{eq: RQNSF AR} and $\left\{\tau,c_{\rho_1},\dots,c_{\rho_d}\right\}$ implicitly defines $T_\vartheta$ for the RQ-NSF (AR) with parameter $\vartheta = \left(\rho_1,\dots,\rho_d\right)^\top$.

The RQS map $Z_i \mapsto \tau(Z_i;h_i)$ is monotonically increasing, continuously differentiable, piecewise rational-quadratic functions in each of $K$ bins, and it interpolates between knots given the derivatives at internal points. $\tau$ is parameterized by $h_i=\{h_i^w,h_i^h,h_i^s\}$, and they correspond to $K-1$ bin widths, $K-1$ bin heights, and derivatives at $K-1$ internal points, respectively. Each conditioner $(Z_{<i},\eta) \mapsto c_{\rho_i}(Z_{<i},\eta)$ is a neural network with parameter $\rho_i$ that outputs the spline parameters $h_i$ based on previous latent variables $Z_{<i}$ and conditioning variable $\eta$. The autoregressive structure yields a lower triangular Jacobian, so the Jacobian determinant reduces to $\prod_{i=1}^d \partial_{Z_i} \tau(Z_i;h_i)$, the product of diagonal elements of the Jacobian.
When $J > 1$, permutation layers (e.g., reversing or rotating the variable order) are inserted between autoregressive transformations.

\paragraph{Coupling flow.} The above formulation can be adapted to include the coupling flow in the NSF framework. Compared to the autoregressive flow, the coupling flow splits $Z$ into two parts with roughly equal dimension: $Z_{<p+1}=(Z_1,\dots,Z_{p})^\top$ and $Z_{\geq p+1} =(Z_{p+1},\dots,Z_d)^\top$ with $p=\lfloor d/2 \rfloor$. The NSF with this flow choice (NSF (C)) then keeps $Z_{<p+1}$ fixed, and transforms $Z_{\geq p+1}$ element-wise given $Z_{<p+1}$ as
\begin{equation}\label{eq: RQNSF C}
\theta_i =
\begin{cases}
Z_i &i=1,\dots,p,\\
\tau \left( Z_{i} ; h_i \right) \quad \text{with} \quad h_i = c_{\rho_i} \left(Z_{<p+1},\eta \right) &i=p+1,\dots,d.\\
\end{cases}
\end{equation}
NSF (C) corresponding to the choice of RQS as transformer $\tau$, and neural network as the conditioner $c_{\rho_i}$ is denoted by RQ-NSF (C). As in RQ-NSF (AR), the flow choice \eqref{eq: RQNSF C} and $\left\{\tau,c_{\rho_{p+1}},\dots,c_{\rho_d}\right\}$ implicitly defines $T_\vartheta$ in RQ-NSF (C) with parameter $\vartheta = \left(\rho_{p+1},\dots,\rho_d\right)^\top$. The transformer $\tau$ and conditioners $c_{\rho_i}$ are defined as in RQ-NSF(AR), except that the input to $c_{\rho_i}$ is $\left(Z_{<p+1},\eta \right)$ which remains identical for all $i=p+1,\dots,d$.
The coupling structure is also a block lower triangular Jacobian, and the determinant of the Jacobian equals $\prod_{i>p} \partial_{Z_i} \tau(Z_i;h_i)$. To ensure every coordinate is updated and to propagate dependencies across all dimensions, multiple coupling layers are stacked with permutations that swap the roles of the fixed and transformed subsets between layers.\\

In both flow choices, the spline parameters (bin widths, bin heights, and derivatives at internal points) are parameterized by neural networks. To construct conditional normalizing flows in NeVI-Cut, we employ a ReLU-based neural network as the conditioner and incorporate the conditioning variable $\eta$ as an additional input.

\subsection{Unconstrained Monotonic Neural Networks}

\cite{wehenkel2019unconstrained} proposed the Unconstrained Monotonic Neural Networks (UMNN), which is an integration-based transformation. We extend this framework to the conditional setting of NeVI-Cut by defining
\begin{equation}\label{eq:umnn}
\theta = T_\vartheta(\eta, Z) = a_{\vartheta_a}(\eta) + \int_0^Z \exp \left(g_{\vartheta_g}(\eta, t)\right)dt .
\end{equation}
Here, $a_{\vartheta_a}(\cdot)$ is a neural network parameterized by $\vartheta_a$ that models the conditional median, and $g_{\vartheta_g}(\cdot)$ is a neural network with parameters $\vartheta_g$ that models the log-derivative $g(\eta,z)$ of the conditional quantile $T_\vartheta(\eta,z)$ with respect to $z$. $\vartheta = (\vartheta_a, \vartheta_g)^\top$ is the overall parameter of the transformation. The construction ensures that the derivative $\partial_Z T_\vartheta >0$, thereby guaranteeing monotonicity and invertibility of the transformation.

\section{Proofs}\label{sec:proofs}

In this Section, we provide proofs of the general results on conditional normalizing flows and NeVI-Cut solutions using them. Proofs of the results for specific flow models (Theorems \ref{th:rqs} and \ref{th:umnn}) are in the next Section.

We first state a simple identity relating conditional density to the conditional quantile map which is used at multiple places in the theory. 
\begin{lemma}\label{lem:quantile}
Let a positive conditional density $p(\theta \given \eta)$ satisfy Assumption~\ref{asm:mono}. Define the conditional quantile map $T_p$ as in (\ref{eq:qtldefn}).
Then \(T_p\) is continuous on \( K_\eta \times \mathbb R\), strictly increasing and continuously differentiable in \(z\), and
\begin{equation}\label{eq:identityderiv}
\log p \left(T_p(\eta,z)\given \eta\right) \;=\; \log \phi(z) \;-\;\log \partial_z T_p(\eta,z) .
\end{equation}
\end{lemma}

\begin{proof}[Proof of Lemma \ref{lem:quantile}]
By definition \eqref{eq:qtldefn}, $T_p(\eta,z)$ is a composition of $ u \mapsto F_p^{-1} (u \mid \eta)$ and $z \mapsto \Phi (z)$. Under Assumption~\ref{asm:mono},
\(T_p\) is continuous on \( K_\eta \times \mathbb R\), strictly increasing and continuously differentiable in \(z\).

Following \eqref{eq:qtldefn}, we have
\(
F_p \big(T_p(\eta,z) \given \eta\big)=\Phi(z).
\)
Differentiating both sides with respect to \(z\) and using the chain rule, we have:
\[
\partial_\theta F_p \big(T_p(\eta,z) \given \eta \big)\cdot \partial_z T_p(\eta,z)
= \phi(z).
\]
Because \(\partial_\theta F_p(\theta \given \eta)=p(\theta\mid \eta)\), substituting \(\theta=T_p(\eta,z)\) yields
\[
p \big(T_p(\eta,z)\mid \eta\big)\,\partial_z T_p(\eta,z)=\phi(z).
\]
Both factors are positive, hence taking logarithms gives the result. 
\end{proof}

As the KLD features the log-density, Lemma \ref{lem:quantile} helps write the KLD in terms of the derivative of the log-quantile function and the log base density $\phi(z)$, enabling us to apply respective bounds for each term to bound the KLD. 

\begin{proof}[Proof of Lemma \ref{lem:lips}]
\emph{Lipschitz modulus for $\log \partial_z T_p$.}
For $(z_i,\eta_i)\in K_M$ set $\theta_i=T_p(\eta_i,z_i)$.  
Applying Lemma \ref{lem:quantile}, by the two moduli condition on $K_\eta \times [-G(M),G(M)]$ and since $\log\phi$ is $M$-Lipschitz on $[-M,M]$,
$$
\begin{aligned}
\big|\log \partial_z T_p(\eta_1,z_1)-\log \partial_z T_p(\eta_2,z_2)\big|
&\le \big|\log\phi(z_1)-\log\phi(z_2)\big|
   +\big|\log p(\theta_1\mid\eta_1)-\log p(\theta_2\mid\eta_2)\big| \\
   &\le \big|\log\phi(z_1)-\log\phi(z_2)\big|
   +\big|\log p(\theta_1\mid\eta_1)-\log p(\theta_2\mid\eta_1)\big|  \\
      &\quad +
   \big|\log p(\theta_2\mid\eta_1)-\log p(\theta_2\mid\eta_2)\big|  \\
&\le M\,|z_1-z_2|
   + L_S\!\big(G(M)\big)\,|\theta_1-\theta_2|
   + L_{CP}\!\big(G(M)\big)\,\|\eta_1-\eta_2\|_1.
\end{aligned}
$$
It remains to bound $|\theta_1-\theta_2|$. Decompose
\[
\begin{aligned}
|\theta_1-\theta_2|
&=\big|T_p(\eta_1,z_1)-T_p(\eta_2,z_2)\big| \\
&\le \big|T_p(\eta_1,z_1)-T_p(\eta_1,z_2)\big|
   +\big|T_p(\eta_1,z_2)-T_p(\eta_2,z_2)\big|.
\end{aligned}
\]
For the first term, by the mean value theorem and Assumption~\ref{asm:mono}$(iii)$,
\[
\big|T_p(\eta_1,z_1)-T_p(\eta_1,z_2)\big|
\le \sup_{\eta\in K_\eta}\sup_{|u|\le M}\partial_u T_p(\eta,u)\,|z_1-z_2|
\le e^{H(M)}\,|z_1-z_2|.
\]
For the second term, set $z=z_2$ and write $\tilde\theta_1=T_p(\eta_1,z)$ and $\tilde\theta_2=T_p(\eta_2,z)$.
Since $F_p(\tilde\theta_i\mid\eta_i)=\Phi(z)$,
\[
\big|F_p(\tilde\theta_1\mid\eta_1)-F_p(\tilde\theta_1\mid\eta_2)\big|
=\big|F_p(\tilde\theta_2\mid\eta_2)-F_p(\tilde\theta_1\mid\eta_2)\big|.
\]
By the cdf modulus on $[-G(M),G(M)]\times K_\eta$,
\[
\big|F_p(\tilde\theta_1\mid\eta_1)-F_p(\tilde\theta_1\mid\eta_2)\big|
\le L_{CP}\!\big(G(M)\big)\,\|\eta_1-\eta_2\|_1.
\]
By the mean value theorem applied to $\theta\mapsto F_p(\theta\mid\eta_2)$ between $\tilde\theta_1$ and $\tilde\theta_2$,
there exists $\bar\theta$ between $\tilde\theta_1$ and $\tilde\theta_2$ such that
\[
\big|F_p(\tilde\theta_2\mid\eta_2)-F_p(\tilde\theta_1\mid\eta_2)\big|
= p(\bar\theta\mid\eta_2)\,|\tilde\theta_2-\tilde\theta_1|.
\]
Since $|\bar\theta|\le G(M)$ and Assumption~\ref{asm:mono}$(iv)$ gives $p(\bar\theta\mid\eta_2)\ge c e^{-\alpha G(M)^2}$,
\[
\big|T_p(\eta_1,z_2)-T_p(\eta_2,z_2)\big|
=|\tilde\theta_1-\tilde\theta_2|
\le c^{-1}e^{\alpha G(M)^2}\,L_{CP}\!\big(G(M)\big)\,\|\eta_1-\eta_2\|_1.
\]
Combining the two pieces,
\[
|\theta_1-\theta_2|
\le e^{H(M)}\,|z_1-z_2|
+ c^{-1}e^{\alpha G(M)^2}\,L_{CP}\!\big(G(M)\big)\,\|\eta_1-\eta_2\|_1.
\]
Substituting into the previous display yields
\[
\begin{aligned}
\big|\log \partial_z T_p(\eta_1,z_1)-\log \partial_z T_p(\eta_2,z_2)\big|
&\le \Bigl(M+e^{H(M)}L_S\!\big(G(M)\big)\Bigr)\,|z_1-z_2| \\
&\quad + \Bigl( L_{CP}\!\big(G(M)\big)
+ c^{-1}e^{\alpha G(M)^2}L_S\!\big(G(M)\big)L_{CP}\!\big(G(M)\big)\Bigr)\,\|\eta_1-\eta_2\|_1.
\end{aligned}
\]
Therefore $\log \partial_z T_p$ is $\mathsf L_b(M)$-Lipschitz on $K_M$ with respect to the $\ell_1$ distance
$\|(\eta,z)-(\eta',z')\|_1:=\|\eta-\eta'\|_1+|z-z'|$, where one may take
\[
\mathsf L_b(M)
:=\max\Bigl\{M+e^{H(M)}L_S\!\big(G(M)\big),\ L_{CP}\!\big(G(M)\big)\bigl(1+c^{-1}e^{\alpha G(M)^2}L_S\!\big(G(M)\big)\bigr)\Bigr\}.
\]

\noindent \emph{Lipschitz modulus for the median $T_p(\eta,0)$ in $\eta$.}
Fix $\eta_1,\eta_2\in K_\eta$ and write $\theta_i=T_p(\eta_i,0)$, so that
\[
F_{\eta_i}(\theta_i)=\Phi(0)=\tfrac12,\qquad i=1,2,
\]
where $F_\eta(\cdot)=F_p(\cdot\mid\eta)$.
Then
\[
|F_{\eta_2}(\theta_1)-F_{\eta_2}(\theta_2)|
=|F_{\eta_2}(\theta_1)-\tfrac12|
=|F_{\eta_2}(\theta_1)-F_{\eta_1}(\theta_1)|
\le L_{CP}(M_0)\,\|\eta_1-\eta_2\|_1,
\]
where the last inequality uses the cdf perturbation bound from Assumption~\ref{asm:mono}$(ii)$ and the fact that
$|\theta_1|\le M_0$ by Assumption~\ref{asm:mono}$(iii)$.

By the mean value theorem applied to the function $\theta\mapsto F_{\eta_2}(\theta)$ between $\theta_1$ and $\theta_2$,
there exists $\bar\theta$ between $\theta_1$ and $\theta_2$ such that
\[
|F_{\eta_2}(\theta_1)-F_{\eta_2}(\theta_2)|
= p(\bar\theta\mid\eta_2)\,|\theta_1-\theta_2|.
\]
Since $|\bar\theta|\le M_0$ and Assumption~\ref{asm:mono}$(iv)$ yields
$p(\bar\theta\mid\eta_2)\ge c e^{-\alpha M_0^2}$, we obtain
\[
|\theta_1-\theta_2|
\le c^{-1}e^{\alpha M_0^2}\,L_{CP}(M_0)\,\|\eta_1-\eta_2\|_1,
\]
which proves the desired Lipschitz continuity of $\eta\mapsto T_p(\eta,0)$ on $K_\eta$.

\end{proof}

\begin{proof}[Proof of Theorem \ref{thm:ckl-explicit}]
We use the metric $d_M$ on conditional laws induced by their Gaussian quantile maps:
\[
d_M(p,q):=\sup_{(z,\eta)\in K_M}\max\Big(\,|T_p(\eta,0)-T(\eta,0)|,\;
|\log\partial_zT_p(\eta,z)-\log\partial_zT(\eta,z)|\,\Big).
\]

\emph{Step 1: Core uniform approximation on $K_M$.}
By Assumption~\ref{asm:pm}$(iii)$, for 
budget 
$m=C(M,\varepsilon)$ there exists $T\in\mathcal F_m$ such that
\begin{equation}\label{eq:ua-core}
d_M\big(q_T,p\big)\ 
\ \le\ \varepsilon. 
\end{equation}
Write $\Delta_0(\eta):= T(\eta,0)-T_p(\eta,0)$, and $\Delta_\ell(\eta,z):=\log\partial_zT(\eta,z)-\log\partial_z T_p(\eta,z)$.

Then $|\Delta_0|\le\varepsilon$ and $|\Delta_\ell|\le\varepsilon$ on $K_M$.\\

\emph{Step 2: Bounding the KL on the core $|z|\le M$.}
For each $\eta$,
\[
\mathrm{KL} \big(q_T(\cdot\mid\eta)\,\|\,p(\cdot\mid\eta)\big)
=\mathbb E\Big[\log\partial_z T_p(\eta,Z)-\log\partial_zT(\eta,Z)
+\log p\,\!\big(T_p(\eta,Z)\mid\eta\big)-\log p\,\!\big(T(\eta,Z)\mid\eta\big)\Big].
\]
On $|Z|\le M$, the first difference is $\le \varepsilon$ by \eqref{eq:ua-core}.
For the second, use the spikiness modulus on the bounded $\theta$–range:
on $|z|\le M$,
\[
|T(\eta,z)-T_p(\eta,z)|
\le |\Delta_0(\eta)| + \int_0^{|z|}\!\big|e^{\Delta_\ell(\eta,u)}-1\big|\,\partial_u T_p(\eta,u)\,du
\le \varepsilon + 2\varepsilon\,|z|\,e^{H(M)}
\le \varepsilon\big(1+2M\,e^{H(M)}\big),
\]
because $|\log\partial_u T_p|\le H(M)$ and $|e^{\Delta_\ell}-1|\le e^{\varepsilon}-1\le (e-1)\varepsilon \leq 2\varepsilon$ for $\varepsilon \leq 1/2$.
Note that on $K_M$,  $|T_p(\eta,Z)| \leq G(M) := M_0 + Me^{A_d + B_dM}$ by (\ref{eq:qtl}). Similarly, $|T(\eta,Z)| \leq \widetilde G(M) := M^* + |M|e^{A^* + B^*|M|}$. As $M^* > M_0$, $A^* > A_d$, and $B^* > B_d$, we have $\widetilde G(M) \geq G(M)$ and both $T_p(\eta,Z)$ and $T(\eta,Z)$ lie in $[-\widetilde G(M),\widetilde G(M)]$. 
Hence, 
\begin{align*}
\big|\log p \big(T(\eta,z)\mid\eta\big)-\log p \big(T_p(\eta,z)\mid\eta\big)\big|
&\le L_S \big(\widetilde G(M)\big)\,|T(\eta,z)-T_p(\eta,z)|\\
&\le \varepsilon\,L_S \big(\widetilde G(M)\big)\,(1+2M e^{H(M)}).
\end{align*}
Integrating $\phi(z)$ over $|z|\le M$ (mass $\le1$) gives the core contribution is 
\[
\begin{aligned}
& \le \varepsilon\Big\{1+L_S \big(\widetilde G(M)\big) \left(1+2M e^{H(M)}\right)\Big\}\\
& \le 3\varepsilon\Big\{1+L_S \big(\widetilde G(M)\big)\widetilde G(M)\Big\}.
\end{aligned}
\]
For this to be less than $\varepsilon'/2$,
\begin{equation*}
\varepsilon \le \frac{\varepsilon'}{6\Big\{1+\widetilde G(M) L_S \big(\widetilde G(M)\big)\Big\}} .
\end{equation*}

\emph{Step 3: Gaussian tails $|z|>M$.}
Using the identity above and Assumption~\ref{asm:pm}$(ii)$,
\[
\log\partial_z T(\eta,z)\le A^*+B^*|z|,\quad
|T(\eta,z)|\le |T(\eta,0)|+\!\int_0^{|z|}\!\partial_uT(\eta,u)\,du
\le M^*+e^{A^*}|z|e^{B^*|z|}.
\]
Using $(a+b)^2 \le 2(a^2 + b^2)$ for any real $a$ and $b$ leads to 
\[
\begin{aligned}
T(\eta,z)^2 \;\le\; \Big(M^{*}+e^{A^{*}}|z|e^{B^{*}|z|}\Big)^2 \;\le\; 2M^{*2}+
2e^{2A^{*}}
z^2e^{2B^{*}|z|}.
\end{aligned}
\]
By the tail-thinness lower bound in Assumption~\ref{asm:mono}$(iv)$,
\begin{equation*}
-\log p \big(T(\eta,z)\mid\eta\big)\le \alpha\,T(\eta,z)^2 - \log c .
\end{equation*}
With $\log\phi(z)=-\tfrac{z^2}{2}-\tfrac12\log(2\pi)$, Assumption~\ref{asm:mono}$(iv)$ and \ref{asm:pm}$(ii)$ give,
for $|z|>M$ and all $\eta$,
\[
\begin{aligned}
&\log\phi(z)-\log p \big(T(\eta,z)\mid\eta\big)-\log\partial_z T(\eta,z)\\
\le &-\frac{z^2}{2}
-\frac12\log(2\pi)-\log c
+\alpha\,\left( 2M^{*2}+ 
2e^{2A^{*}} z^2e^{2B^{*}|z|} \right)
+A^{*}+B^{*}|z| \\
\le &-\frac12\log(2\pi)-\log c
+\alpha\,\left( 2M^{*2}+ 
2e^{2A^{*}} z^2e^{2B^{*}|z|} \right)
+A^{*}+B^{*}|z| \\
\le 
 &C_{1}
\;+\; C_{2}\,z^2\,e^{2B^{*}|z|}
\;+\; B^{*}|z|,
\end{aligned}
\]
where the explicit constants are
\[
C_{1}=A^{*}-\tfrac12\log(2\pi)-\log c+2\alpha\,M^{*2},
\qquad
C_{2}=
2\alpha e^{2A^{*}}.
\;
\]
Then the tail bound is \[
\begin{aligned}
\int_{|z|>M}\!\phi(z)\,[\cdots]\,dz
&\le \frac{1}{\sqrt{2\pi}}\Big[
C_1\,I_0(M)\;+\;B^{*}\,I_1(M)\;+\;C_2\,I_2(M)\Big],
\end{aligned}
\]
where
\begin{align*}
I_0(M) :=\!\int_{|z|>M}\!e^{-z^2/2}dz,\quad
I_1(M):=\!\int_{|z|>M}\!|z|e^{-z^2/2}dz,\quad
I_2(M) :=\!\int_{|z|>M}\!z^2 e^{-z^2/2+2B^{*}|z|}dz .
\end{align*}

By Mills ratio,
\[
I_0(M)\le \frac{2\sqrt{2\pi}\,\phi(M)}{M},\qquad
I_1(M)=2\!\int_M^\infty\!ze^{-z^2/2}dz=2\,e^{-M^2/2}.
\]

Rewriting $I_2$ and using $z^2-4B^{*}z = (z-2B^{*})^2 - 4{B^{*}}^2$, 
\begin{equation*}
I_2(M) = 2 \int_M^\infty z^2 e^{-z^2/2+2B^{*}z}dz = 2 e^{2{B^{*}}^2} \int_M^\infty z^2 e^{-\frac{1}{2}(z-2B^{*})^2}dz .
\end{equation*}
Using change of variable $t=z-\gamma$ where $\gamma=2B^*$, and \((t+\gamma)^2 \le 2(t^2+\gamma^2)\), we have

\[
I_2(M)=2e^{\gamma^2/2}\int_{M-\gamma}^{\infty} (t+\gamma)^2 e^{-t^2/2}\,dt
\;\le\;2e^{\gamma^2/2}M_\gamma\!\int_{M-\gamma}^{\infty}\!(1+t^2)e^{-t^2/2}\,dt,
\]
where \(M_\gamma:=\max\{2\gamma^2,\,2\}\).

For \(x>0\),
\[
\int_x^\infty (1+t^2)e^{-t^2/2}\,dt
=\int_x^\infty e^{-t^2/2}\,dt+\int_x^\infty t^2 e^{-t^2/2}\,dt .
\]
Mills ratio gives
\[
\int_x^\infty e^{-t^2/2}\,dt \le \frac{1}{x}e^{-x^2/2}\qquad(x>0).
\]
Also,
\[
\int_x^\infty t^2 e^{-t^2/2}\,dt
=\Big[-t\,e^{-t^2/2}\Big]_x^\infty+\int_x^\infty e^{-t^2/2}\,dt
= x\,e^{-x^2/2}+\int_x^\infty e^{-t^2/2}\,dt
\le \Big(x+\frac{1}{x}\Big)e^{-x^2/2}.
\]
Hence
\[
\int_x^\infty (1+t^2)e^{-t^2/2}\,dt
\le \Big(x+\frac{2}{x}\Big)e^{-x^2/2}.
\]
For \(x\ge 1\), \(x+2/x\le 2(1+x)\), so
\[
\int_x^\infty (1+t^2)e^{-t^2/2}\,dt
\le 2(1+x)e^{-x^2/2}.
\]
Taking \(x=M-\gamma\) with $M \geq \gamma +1$
gives
\[
I_2(M)\;\le\;4M_\gamma\,e^{\gamma^2/2}\,(1+M-\gamma)\,e^{-\frac12(M-\gamma)^2}
\;\le\;C_\gamma\,(1+M)\,e^{-\frac12(M-\gamma)^2},
\]
with \(C_\gamma:=4M_\gamma e^{\gamma^2/2}\). Setting \(\gamma=2B^{*}\) and $M_\gamma \leq 2\gamma^2 + 2 \leq 2 (1 + 4B^{*2})$, yields
\[
I_2(M)\;\le\;\widetilde C_{B^*}\,(1+M)\,e^{-\frac12\,(M-2B^{*})^2}, \mbox{ where } \widetilde C_{B^*} = 8(1+4B^{*2})e^{2B^{*2}}. 
\]

Putting the pieces together,
\[
\begin{aligned}
\int_{|z|>M}\!\phi(z)\,[\cdots]\,dz
\ &\le\ \frac{C_1}{\sqrt{2\pi}}\frac{2\,e^{-M^2/2}}{M}
\;+\;\frac{B^{*}}{\sqrt{2\pi}}\,2e^{-M^2/2}
\;+\;\frac{C_2}{\sqrt{2\pi}}\,\widetilde C_{B^*}\,(1+M)\,e^{-\frac12 (M-2B^{*})^2}
\, \\
&\le \, R(1+M)\exp\!\Big(-\tfrac12(M-2B^*)^2\Big)
\end{aligned}
\]
after absorbing all constants in $R$ and using the largest term.
For this to be less than $\varepsilon'/2$, we 
write
\[
t := M-2B^*, \qquad 1+M=1+2B^*+t\le (1+2B^*)e^{t}.
\]
we have
\[
R(1+M)e^{-\tfrac12 t^2}
\;\le\; R(1+2B^*)\,e^{t-\tfrac12 t^2}
\;=\; R(1+2B^*)\,e^{-\tfrac12 (t-1)^2+\tfrac12}.
\]
For this to be less than $\varepsilon'/2$
\[
e^{-\tfrac12 (t-1)^2}\ \le\ \frac{\varepsilon'}{2\,R(1+2B^*)\,e^{1/2}}
\quad\Longleftrightarrow\quad
(t-1)^2\ \ge\ 2\log\!\frac{2 e^{1/2} R(1+2B^*)}{\varepsilon'}.
\]
\[
M\ \ge\ 2B^* + 1 + \sqrt{\,2\log\!\frac{2 e^{1/2} R(1+2B^*)}{\varepsilon'}\,}.
\]
Absorbing $2 e^{1/2} (1+2B^*)$ in $R$ we can choose $M = 2B^* + 1 + \sqrt{\,2\log\!\frac{R}{\epsilon'}}$ for a suitable constant $R$. So, for any small $\varepsilon'$, $M$ can be expressed as $O(\sqrt{-\log \varepsilon'})$ where the constant depends only on $M^*,A^*,B^*,\alpha$ and $c$.

Combining the findings of steps 2 and 3 completes the proof.

\end{proof}

\begin{proof}[Proof of Corollary \ref{cor:condvarflow}]
\emph{(a) Existence.}
For notational simplicity we denote $p_{\text{cut}}$ by $p$. Recall that the parameter set $\Theta_{\,m}\subset\mathbb R^{m}$ is compact by Assumption~\ref{asm:pm}.  
For each fixed $(\eta,z)$, the maps $\vartheta\mapsto T_\vartheta(\eta,z)$ and
$\vartheta\mapsto\partial_z T_\vartheta(\eta,z)$ are continuous by (Assumption \ref{asm:pm}$(i)$),
hence the integrand
\[
I_\vartheta(\eta,z)
:=\log \phi(z) - \log\partial_z T_\vartheta(\eta,z)-\log p \big(T_\vartheta(\eta,z)\mid\eta\big)
\]
is continuous in $\vartheta$.  By assumptions \ref{asm:pm}$(ii)$ and \ref{asm:mono}$(iv)$,
\[
\begin{aligned}
|I_\vartheta(\eta,z)|
&\;\le\; \frac 12 \log 2 \pi+ \frac{z^2}2  + A^{*}+B^{*}|z|+|\log c|+\alpha\Big(M^{*}+e^{A^{*}}|z|e^{B^{*}|z|}\Big)^{\!2}\\
&\;\le\; C_1 + C_2z^2 + C_3(1+ z^{2})e^{2B^{*}|z|},
\end{aligned}
\]
with $C_1,C_2$ independent of $\vartheta$.  Since
$\int_{\mathbb R}(1+z^{2})e^{2B^{*}|z|}\phi(z)\,dz<\infty$
(complete the square), the envelope is integrable against $\phi(z)p(\eta)$.
By dominated convergence, $\vartheta\mapsto R(q_{T_\vartheta})$ is continuous on the compact set $\Theta_{\,m}$, hence by Weierstrass Extreme Value Theorem it attains its minimum; existence of $\hat q_m$ follows.

\emph{(b) KLD bound.}
By Theorem~\ref{thm:ckl-explicit}, given a small $\varepsilon$, for the budget
$m:=C^*(\varepsilon)$ there exists $T\in\mathcal F_m$ such that
\(
R\big(q_{T}\big)
\;\le\;\varepsilon.
\)
Because $\hat q_m$ minimizes $R(q)$ over $q \in \mathcal Q_m$,
\[
R(\hat q_m)
\;\le\;R(q_T) \leq \varepsilon. 
\]
This proves the claim.
\end{proof}

\begin{proof}[Proof of Theorem \ref{th:mcmc}]

Since $\mathcal Q_m=\{q_{T_\vartheta}: \vartheta\in\Theta_m\}$, with a slight abuse of notation, we can write $R(q)$ as $R(\vartheta)$ and $R_N(q)$ as $R_N(\vartheta)$ for the corresponding $\vartheta$. As $\Theta_m$ is compact and
$\vartheta\mapsto R_{N}(q_\vartheta)$ is continuous, once again by Weierstrass Extreme Value Theorem, we have
$\hat q_{N,m}\in\arg\min_{\mathcal Q_m}R_{N}(q)$ exists.

Before proceeding further we state and prove a proposition on uniform laws of large numbers for stationary ergodic processes. 

\begin{proposition}[Uniform ergodic LLN for the empirical KL]\label{prop:ULLN}
Let $p(\theta \given \eta)$ satisfy Assumption \ref{asm:mono} for some compact $K_\eta$ and $p(\eta)$ is a distribution supported on $K_\eta$. 

Let $\{\eta_r\}_{r\ge1}$ be a stationary ergodic Markov chain with
invariant distribution $p(\eta)$. Let $\calF_m=\{T_\vartheta |\; \vartheta \in\Theta_m\}$ denote a class satisfying Assumption \ref{asm:pm}. Let $q_\vartheta$ be the conditional distribution $\theta \given \eta = T_\vartheta(\eta,Z)$, $Z \sim N(0,1)$, and 
\[
f_\vartheta(\eta)=\mathrm{KL} \big(q_\vartheta(\theta\mid\eta)\,\|\,p(\theta\mid\eta)\big),\qquad
R(\vartheta)=\mathbb E[f_\vartheta(\eta)],\quad
R_{N}(\vartheta)=\frac1N\sum_{r=1}^N f_\vartheta(\eta_r).
\]
 Then
\(
\sup_{\vartheta\in\Theta_m}\big|R_{N}(\vartheta)-R(\vartheta)\big|\xrightarrow[N\to\infty]{a.s.}0 .
\)
\end{proposition}

\begin{proof}[Proof of Proposition \ref{prop:ULLN}]
\textit{Envelope.} Let $H(\vartheta,\eta,z)=\log\phi(z)-\log p \big(T_\vartheta(\eta,z)\mid\eta\big)-\log\partial_z T_\vartheta(\eta,z)$. Then $f_\vartheta(\eta)=\mathbb E_{\theta \given \eta = T_\vartheta(\eta,Z)}[H(\vartheta,\eta,Z)]$. Following the proof of Corollary \ref{cor:condvarflow}, 
$f_\vartheta(\eta) \leq \tau$ for some positive $\tau$ free of $\eta,\vartheta$ and for each fixed $\eta$, the map $\vartheta \mapsto f_\vartheta(\eta)$ is continuous on compact $\Theta_m$, implying that it is uniformly continuous. 

Define for $\varepsilon>0$
\[
L_\varepsilon(\eta)
=\sup\{|f_\vartheta(\eta)-f_{\vartheta'}(\eta)|:\ \vartheta,\vartheta'\in\Theta_m,\ \|\vartheta-\vartheta'\|\le\varepsilon\}.
\]
Then by uniform continuity $L_\varepsilon(\eta)\to 0$ pointwise as $\varepsilon\downarrow0$, and
\(0\le L_\varepsilon(\eta)\le 2\tau\).

For each fixed \(\vartheta\in\Theta_m\), the map \(\eta\mapsto f_\vartheta(\eta)\) is measurable
(it is an integral of a continuous function dominated by an integrable envelope).
Let \(D\subset\Theta_m\) be a countable dense subset. By continuity of
\(\vartheta\mapsto f_\vartheta(\eta)\) on compact \(\Theta_m\), for every \(\eta\)
\[
L_\varepsilon(\eta)
=\sup\{\,|f_\vartheta(\eta)-f_{\vartheta'}(\eta)|:\ \vartheta,\vartheta'\in D,\ \|\vartheta-\vartheta'\|\le\varepsilon\,\}.
\]
The rightmost expression is the supremum of a countable family of measurable
functions, hence \(L_\varepsilon\) is measurable. 

So \(L_\varepsilon\in L^1(p)\) and by dominated convergence,
\begin{equation}\label{eq:ELto0}
\mathbb E\,L_\varepsilon(\eta)\longrightarrow 0 \qquad (\varepsilon\downarrow0).
\end{equation}.

Since \(\{\eta_r\}_{r\ge1}\) is stationary ergodic and
\(L_\varepsilon\in L^1(p)\), Birkhoff’s ergodic theorem yields
\[
\frac1N\sum_{r=1}^N L_\varepsilon(\eta_r)
\;\xrightarrow[N\to\infty]{a.s.}\;
\mathbb E_{p(\eta)}[L_\varepsilon(\eta)] .
\]

Let $\{\vartheta_j\}_{j=1}^{n(\varepsilon)}$ be an $\varepsilon$-net of $\Theta_m$.
For any $\vartheta \in \Theta_m$, choose $j$ with $\|\vartheta-\vartheta_j\|\le\varepsilon$ and write
\[
\begin{aligned}
\big|R_{N}(\vartheta)-R(\vartheta)\big|
&\le\big|R_{N}(\vartheta)-R_N(\vartheta_j)\big|
+\big|R_{N}(\vartheta_j)-R(\vartheta_j)\big|
+\big|R(\vartheta)-R(\vartheta_j)\big|.\\
&\le\frac1N\sum_{r=1}^N L_\varepsilon(\eta_r)
+\big|R_{N}(\vartheta_j)-R(\vartheta_j)\big|
+\mathbb E\,L_\varepsilon(\eta).\\
\end{aligned}
\]

\noindent
Taking $\sup_{\vartheta\in\Theta_m}$ in the inequality, 
we get
\begin{align*}
\sup_{\vartheta\in\Theta_m}\big|R_{N}(\vartheta)-R(\vartheta)\big|
&\le \frac1N\sum_{r=1}^N L_\varepsilon(\eta_r)+\max_{1\le j\le n(\varepsilon)}\big|R_{N}(\vartheta_j)-R(\vartheta_j)\big|
      +\mathbb E\,L_\varepsilon(\eta).
\end{align*}

Taking $\limsup_{N\to\infty}$ on both sides and applying the ergodic theorem to the
empirical average of $L_\varepsilon(\eta_r)$ yields
\begin{align}
\limsup_{N\to\infty}\sup_{\vartheta\in\Theta_m}\big|R_{N}(\vartheta)-R(\vartheta)\big|
&\le \max_{1\le j\le n(\varepsilon)}\limsup_{N\to\infty}\big|R_{N}(\vartheta_j)-R(\vartheta_j)\big|
   +2\,\mathbb E\,L_\varepsilon(\eta). \label{eq:limsupbound}
\end{align}
For each fixed $j$, the ergodic theorem gives
$R_{N}(\vartheta_j)\to R(\vartheta_j)$ almost surely, hence the first term on the right side of
\eqref{eq:limsupbound} is zero. Therefore,
\[
\limsup_{N\to\infty}\sup_{\vartheta\in\Theta_m}\big|R_{N}(\vartheta)-R(\vartheta)\big|
\le 2\,\mathbb E\,L_\varepsilon(\eta).
\]
Letting $\varepsilon\downarrow 0$ and using $\mathbb E\,L_\varepsilon(\eta)\to 0$ finishes the proof.
\end{proof}

Returning to the proof of Theorem \ref{th:mcmc}, by Proposition~\ref{prop:ULLN},
\[
\Delta_{N,m}
:=\sup_{\vartheta\in\Theta_m}\big|R_{N}(q)-R(q)\big|
\xrightarrow[N\to\infty]{a.s.}0
\qquad\text{for each fixed }m.
\]

Let $\hat q_m\in\arg\min_{\mathcal Q_m}R(q)$, which exists by Corollary \ref{cor:condvarflow}. Then, for all $N$,
\begin{align*}
R(\hat q_{N,m})
&\le R_{N}(\hat q_{N,m})+\Delta_{N,m}
   \le R_{N}(\hat q_m)+\Delta_{N,m}
   \le R(\hat q_m)+2\Delta_{N,m}.
\end{align*}
Taking $\limsup_{N\to\infty}$
yields
\(
\limsup_{N\to\infty} R(\hat q_{N,m})
\le \inf_{q\in\mathcal Q_m} R(q) \le \varepsilon
\) by Corollary \ref{cor:condvarflow}.

\end{proof}

\section{Proofs for Neural Flow Examples}\label{sec:proofexamples}

We now prove Theorems \ref{th:umnn} and \ref{th:rqs} for specific neural flow classes. 

\subsection{Common Network Notation and Quantitative Bounds}

Let $\eta \in$ some compact $K_\eta \in \mathbb R^d$
and \(\operatorname{diam}_\eta:=\sup_{\eta,\eta'\in K_\eta}\|\eta-\eta'\|_2\).
Write \(\|\,\cdot\,\|_\infty\) for the max norm and $\|\cdot\|_{\text{op}}$ for the operator norm. 
Throughout, clipping uses the one Lipschitz map \(\mathrm{clip}(x;a,b):=\min\{b,\max\{a,x\}\}\).

\subsection{Rational Quadratic Neural Spline Flows}\label{sec:thrqnsf}

\begin{proof}[Proof of Theorem \ref{th:rqs}]
We only need to prove that RQ-NSF satisfies Assumption \ref{asm:pm} as then we can apply Theorem \ref{thm:ckl-explicit}. From the expression of RQS in (\ref{eq:rqnsf}), we have 
\begin{equation}\label{eq:rqnsfderiv}
\begin{aligned}
\partial_z T_\vartheta(\eta,z) &=
\frac{\Delta_i^2\bigl(s_{i+1}t^2+2\Delta_i t(1-t)+s_i(1-t)^2\bigr)}
{\bigl(\Delta_i+(s_{i+1}+s_i-2\Delta_i)t(1-t)\bigr)^2}.
\end{aligned}
\end{equation}

\noindent {\em Assumption \ref{asm:pm}(i)}

For each $\eta\in K_\eta$ and each bin index $i$, the segment map $z\mapsto T_{\vartheta,M}(\eta,z)$ on $[z_i,z_{i+1}]$
is a composition of the affine change of variable $z\mapsto t=(z-z_i)/w$ with a rational function in $t$ whose numerator and denominator are quadratic polynomials
with coefficients depending on $(y_i,y_{i+1},s_i,s_{i+1},\Delta_i)$.
Since $\Delta_i(\eta)>0$, $t(1-t) \leq 1/4$, and $s_i(\eta),s_{i+1}(\eta)>0$, the denominator is strictly positive on $t\in[0,1]$, hence the segment is $C^\infty$ in $z$.
By the defining property of the \cite{durkan2019neural} rational quadratic segment, the right derivative at $z_i$ equals $s_i(\eta)$ and the left derivative at $z_{i+1}$ equals $s_{i+1}(\eta)$.
Therefore the derivative in $z$ matches across adjacent bins, and $z\mapsto T_{\vartheta,M}(\eta,z)$ is $C^1$ on $[z_0,z_{2K}]$.
The linear tails have constant derivatives $s_0(\eta)$ and $s_{2K}(\eta)$, which match the boundary derivatives of the spline at $z_0$ and $z_{2K}$, so
$z\mapsto T_{\vartheta,M}(\eta,z)$ is $C^1$ on $\mathbb R$.
Strict monotonicity holds because each segment is strictly increasing when widths are positive, heights are strictly positive, and endpoint derivatives are positive, and all these are enforced by construction.

Continuity in $(\eta,\vartheta)$ follows because ReLU networks are continuous in $(\eta,\vartheta)$, the map $x\mapsto \clip(x,a,b)$ is continuous,
and exponentiation is continuous. The recursion defining the knot values $y_j(\eta)$ is continuous in the inputs $(\Delta_i(\eta))$ and in $y_K(\eta)$.
The segment formula is continuous in $(y_i,y_{i+1},s_i,s_{i+1},\Delta_i,t)$ whenever the denominator is positive, which is uniform here since all slopes are positive.
Therefore, $(\eta,z,\vartheta)\mapsto T_{\vartheta,M}(\eta,z)$ is continuous on $K_\eta\times\mathbb R\times\Theta_m$.
The explicit expression for $\partial_z T_{\vartheta,M}$ on each bin is a continuous function of the same inputs, and the boundary matching yields continuity across knots and tails,
so $(\eta,z,\vartheta)\mapsto \partial_z T_{\vartheta,M}(\eta,z)$ is continuous as required.

\noindent {\em Assumption \ref{asm:pm}(ii)}

Assume that \(K=K(M)\) is chosen so that \(w=M/K\le w_0\) for a fixed \(w_0>0\).
On a bin \([z_i,z_{i+1}]\) with \(t\in[0,1]\) and \(u:=t(1-t)\in[0,1/4]\), define
\[
N:=s_i(1-t)^2+2\Delta_i u+s_{i+1}t^2,
\qquad
D:=\Delta_i+\bigl(s_i+s_{i+1}-2\Delta_i\bigr)u
=\Delta_i(1-2u)+u s_i+u s_{i+1}.
\]
A direct differentiation of the segment formula gives
\[
\partial_z T_{\vartheta,M}(\eta,z)
=
\frac{\Delta_i(\eta)^2\,N}{D^2}.
\]

Define
\[
\mathcal M_i(\eta):=\max\{\Delta_i(\eta),\,s_i(\eta),\,s_{i+1}(\eta)\},
\qquad
\mathfrak m_i(\eta):=\min\{\Delta_i(\eta),\,s_i(\eta),\,s_{i+1}(\eta)\}.
\]
Since \((1-t)^2+2u+t^2=1\) and \((1-2u)+u+u=1\) with nonnegative weights, both \(N\) and \(D\) are convex combinations of
\(\Delta_i,s_i,s_{i+1}\). Hence, for all \(t\in[0,1]\),
\[
\mathfrak m_i(\eta)\le N \le \mathcal M_i(\eta),
\qquad
\mathfrak m_i(\eta)\le D \le \mathcal M_i(\eta).
\]
Therefore,
\[
\frac{\Delta_i(\eta)^2\,\mathfrak m_i(\eta)}{\mathcal M_i(\eta)^2}
\le
\partial_z T_{\vartheta,M}(\eta,z)
\le
\frac{\Delta_i(\eta)^2\,\mathcal M_i(\eta)}{\mathfrak m_i(\eta)^2}
\le
\frac{\mathcal M_i(\eta)^3}{\mathfrak m_i(\eta)^2},
\]
and also
\[
\partial_z T_{\vartheta,M}(\eta,z)
\ge
\frac{\mathfrak m_i(\eta)^3}{\mathcal M_i(\eta)^2}.
\]

If \(z\in[z_i,z_{i+1}]\) and \(w\le w_0\), then \(|z_i|\vee|z_{i+1}|\vee|\bar z_i|\le |z|+w_0\).
By the clipping in the definition of \(\beta_j\) and \(\Delta_i\), this implies
\[
\mathcal M_i(\eta)\le \exp\!\bigl(H_c(|z|+w_0)\bigr),
\qquad
\mathfrak m_i(\eta)\ge \exp\!\bigl(-H_c(|z|+w_0)\bigr),
\]
where \(H_c(t)=A_c+B_c t\).
Consequently, for all \(z\in[-M,M]\),
\[
\exp\!\bigl(-5H_c(|z|+w_0)\bigr)
\le
\partial_z T_{\vartheta,M}(\eta,z)
\le
\exp\!\bigl(5H_c(|z|+w_0)\bigr),
\]
and therefore
\[
\bigl|\log \partial_z T_{\vartheta,M}(\eta,z)\bigr|
\le
5H_c(|z|+w_0)
=
\bigl(5A_c+5B_cw_0\bigr)+5B_c|z|.
\]

On the tails, \(\partial_z T_{\vartheta,M}(\eta,z)=s_0(\eta)\) for \(z\le -M\) and \(\partial_z T_{\vartheta,M}(\eta,z)=s_{2K}(\eta)\) for \(z\ge M\).
Since \(|\log s_0(\eta)|\le H_c(M)\) and \(|\log s_{2K}(\eta)|\le H_c(M)\) by clipping and \(H_c(M)\le H_c(|z|)\le H_c(|z|+w_0)\) for \(|z|\ge M\),
the same bound holds for all \(z\in\mathbb R\).

Thus Assumption~\ref{asm:pm}(ii) holds with
\[
M^*:=M_0+1,
\qquad
A^*:=5A_c+5B_cw_0,
\qquad
B^*:=5B_c,
\]
namely
\[
\sup_{m\ge1}\ \sup_{\vartheta\in\Theta_m}\ \sup_{\eta\in K_\eta}\ |T_{\vartheta,M}(\eta,0)|\le M^*,
\qquad
\sup_{m\ge1}\ \sup_{\vartheta\in\Theta_m}\ \sup_{(\eta,z)\in K_\eta\times\mathbb R}\ 
\bigl|\log\partial_z T_{\vartheta,M}(\eta,z)\bigr|
\le
A^*+B^*|z|.
\]

\noindent {\em Explicit Lipschitz constants for RQ-NSF log-derivative}

On a bin, write $\beta=(\beta_i,\beta_{i+1})$ and define $q(\beta,t):=\log\partial_z T$ in terms of $N$ and $D$ by
\[
q(\beta,t)=2\log\Delta+\log N-2\log D,
\qquad
\Delta=\exp\Bigl(
\tfrac12(\beta_i+\beta_{i+1})
\Bigr),
\]
with $s_i=e^{\beta_i}$ and $s_{i+1}=e^{\beta_{i+1}}$.

Where the clip is differentiable, one has $|\partial(2\log\Delta)/\partial\beta_i|\le 1$ and similarly for $\beta_{i+1}$.
Also
\[
\frac{\partial N}{\partial\beta_i}=s_i(1-t)^2+2\frac{\partial\Delta}{\partial\beta_i}u
\le 
e^{H_c(M)},
\qquad
\frac{\partial D}{\partial\beta_i}=\frac{\partial\Delta}{\partial\beta_i}(1-2u)+us_i\le \tfrac12 e^{H_c(M)},
\]
and the lower bounds $N\ge e^{-H_c(M)}$ and $D\ge e^{-H_c(M)}$ give
\[
\Bigl|\frac{\partial}{\partial\beta_i}\log N\Bigr|\le \frac{e^{H_c(M)}}{e^{-H_c(M)}}=e^{2H_c(M)},
\qquad
\Bigl|\frac{\partial}{\partial\beta_i}\bigl(2\log D\bigr)\Bigr|\le 2\frac{(1/2)e^{H_c(M)}}{e^{-H_c(M)}}=e^{2H_c(M)}.
\]
Therefore
\[
\Bigl|\frac{\partial q}{\partial\beta_i}\Bigr|\le 1+e^{2H_c(M)}+e^{2H_c(M)}=1+2e^{2H_c(M)},
\qquad
\Bigl|\frac{\partial q}{\partial\beta_{i+1}}\Bigr|\le 1+2e^{2H_c(M)}.
\]
Since clipping is $1$ Lipschitz, these bounds imply global Lipschitz bounds on the full clipped map.
Define
\[
C_2(M):=1+2\exp(2H_c(M)),
\qquad
C_1(M):=2+4\exp(2H_c(M))=2C_2(M).
\]
Then for all $|\beta_i|\vee|\beta_{i+1}|\le H_c(M)$, $|\beta_i'|\vee|\beta_{i+1}'|\le H_c(M)$, and all $t\in[0,1]$,
\[
\bigl|q(\beta,t)-q(\beta',t)\bigr|\le C_1(M)\bigl(|\beta_i-\beta_i'|+|\beta_{i+1}-\beta_{i+1}'|\bigr),
\]
and if $\beta'=(\beta_i,\beta_i)$, then
\[
\bigl|q(\beta,t)-\beta_i\bigr|=\bigl|q(\beta,t)-q(\beta',t)\bigr|\le C_2(M)\,|\beta_{i+1}-\beta_i|.
\]

\noindent {\em Assumption \ref{asm:pm}(iii)}

Recall that the target family satisfies that $u^*(\eta):=T_p(\eta,0)$ is Lipschitz on $K_\eta$ with constant $\mathsf L_a$ and that
$q^*(\eta,z):=\log\partial_z T_p(\eta,z)$ is Lipschitz on $[-M,M]\times K_\eta$ with constant $\mathsf L_b(M)$.
Also $|q^*(\eta,z)|\le H(|z|)$ on $[-M,M]\times K_\eta$.

Fix $\varepsilon>0$. Choose $K=K(M,\varepsilon)$ large enough that $w=M/K\le w_0$ and
\[
(C_2(M)+1)\mathsf L_b(M)\,w\le \varepsilon/2.
\]
Define knot targets
\[
q_j^*(\eta):=q^*(\eta,z_j),\qquad j=0,\dots,2K.
\]
Each $q_j^*$ is Lipschitz on $K_\eta$ with constant at most $\mathsf L_b(M)$ and satisfies $|q_j^*(\eta)|\le H(|z_j|)$.
Let $\delta:=\min\{\varepsilon,\varepsilon/(4C_1(M))\}$.

\cite{yarotsky2017error}'s theorem for ReLU networks on a compact set implies that, after an affine rescaling of $K_\eta$ into $[0,1]^{d_\eta}$,
there exists a ReLU network $\psi_\vartheta$ outputting $(\tilde u(\eta),v(\eta))$ such that
\[
\sup_{\eta\in K_\eta}|\tilde u(\eta)-u^*(\eta)|\le \delta,
\qquad
\max_{0\le j\le 2K}\sup_{\eta\in K_\eta}|v_j(\eta)-q_j^*(\eta)|\le \delta,
\]
and the number of nonzero weights can be bounded as
\[
m \le C_{\mathrm Y}\,(2K+2)\Bigl(\frac{(\mathsf L_a\vee \mathsf L_b(M))\,\diam(K_\eta)}{\delta}\Bigr)^{d_\eta}
\log\!\Bigl(\frac{(\mathsf L_a\vee \mathsf L_b(M))\,\diam(K_\eta)}{\delta}\Bigr),
\]
where $C_{\mathrm Y}$ depends only on $d_\eta$.

Build $T_{\vartheta,M}$ from $(\tilde u,v)$ as in the definition above.
Since $u^*(\eta)\in[-M_0,M_0] \subset [-M^*,M^*]$ and clipping is $1$ Lipschitz,
\[
\sup_{\eta\in K_\eta}|T_{\vartheta,M}(\eta,0)-T_p(\eta,0)|
=
\sup_{\eta\in K_\eta}\bigl|\clip(\tilde u(\eta),-M^*,M^*)-u^*(\eta)\bigr|\le \delta\le \varepsilon.
\]

For the log derivative, fix $(\eta,z)\in[-M,M]\times K_\eta$ and let $z\in[z_i,z_{i+1}]$ with $t=(z-z_i)/w$.
Let $\beta^*_j(\eta):=q_j^*(\eta)$ and let $\beta_j(\eta)$ be the clipped network outputs.
Since $\beta_j^*(\eta)\in[-H(|z_j|),H(|z_j|)]$, clipping leaves $\beta_j^*$ unchanged, and $|\beta_j(\eta)-\beta_j^*(\eta)|\le |v_j(\eta)-q_j^*(\eta)|\le \delta$.
Using the Lipschitz bound with $C_1(M)$ yields
\[
\bigl|\log\partial_z T_{\vartheta,M}(\eta,z)-\log\partial_z T^*_{M}(\eta,z)\bigr|
\le C_1(M)\bigl(|\beta_i(\eta)-\beta_i^*(\eta)|+|\beta_{i+1}(\eta)-\beta_{i+1}^*(\eta)|\bigr)\le 2C_1(M)\delta,
\]
where $T^*_M$ denotes the spline constructed from the exact knot log slopes $\beta^*_j$ and the exact anchor $u^*$.

It remains to bound the within bin error of the exact parameter spline. Since $\log\partial_z T^*_M(\eta,z_i)=\beta_i^*(\eta)=q^*(\eta,z_i)$,
the bound with $C_2(M)$ gives
\[
\bigl|\log\partial_z T^*_M(\eta,z)-q^*(\eta,z_i)\bigr|\le C_2(M)\,|\beta_{i+1}^*(\eta)-\beta_i^*(\eta)|.
\]
By Lipschitzness of $q^*$ in $z$, $|\beta_{i+1}^*(\eta)-\beta_i^*(\eta)|\le \mathsf L_b(M)w$ and $|q^*(\eta,z)-q^*(\eta,z_i)|\le \mathsf L_b(M)w$,
hence
\[
\bigl|\log\partial_z T^*_M(\eta,z)-q^*(\eta,z)\bigr|\le (C_2(M)+1)\mathsf L_b(M)w\le \varepsilon/2.
\]
Combining the bounds and using $\delta\le \varepsilon/(4C_1(M))$ gives
\[
\sup_{(z,\eta)\in[-M,M]\times K_\eta}\bigl|\log\partial_z T_{\vartheta,M}(\eta,z)-\log\partial_z T_p(\eta,z)\bigr|
\le 2C_1(M)\delta + \varepsilon/2 \le \varepsilon.
\]
This is Assumption~\ref{asm:pm}(iii).

Finally, we apply Theorem~1 to convert the Assumption~2(iii) budget into a
Kullback--Leibler guarantee. Set $M=M(\varepsilon):=R\sqrt{-\log\varepsilon}$
with $R$ as in Theorem~1, and recall $\widetilde G(M)=M^{*}+Me^{A^{*}+B^{*}M}$
and $\widetilde L_S(m):=m\,L_S(m)$. By the core bound in the proof of
Theorem~1, $\sup_{\eta\in K_\eta}\mathrm{KL}\bigl(q_T(\cdot\mid\eta)\,\|\,p(\cdot\mid\eta)\bigr)\le\varepsilon$
is guaranteed once the flow attains, in place of $\varepsilon$, the sharper accuracy
\[
\varepsilon':=\frac{\varepsilon}{6\bigl\{1+\widetilde L_S(\widetilde G(M))\bigr\}}
\]
in the $d_M$ metric, i.e.
\[
\sup_{\eta\in K_\eta}\bigl|T_{\vartheta,M}(\eta,0)-T_p(\eta,0)\bigr|\le\varepsilon',
\qquad
\sup_{(z,\eta)\in K_M}\bigl|\log\partial_z T_{\vartheta,M}(\eta,z)-\log\partial_z T_p(\eta,z)\bigr|\le\varepsilon'.
\]
Running the Assumption~2(iii) construction above at this $\varepsilon'$ amounts to taking
$\delta=\min\{\varepsilon',\varepsilon'/(4C_1(M))\}$ and choosing $w=M/K$ with
$(C_2(M)+1)L_b(M)\,w\le\varepsilon'/2$. Substituting these into the within-bin bound
and the parameter count from \cite{yarotsky2017error} yields the stated
\[
K_\star=\left\lceil\frac{C_r\,\mathsf L_b(M)\,M\,\exp\!\bigl(2H_c(M)\bigr)\,\widetilde L_S\!\bigl(\widetilde G(M)\bigr)}{\varepsilon}\right\rceil,
\qquad
m\le C_r\,(2K_\star+2)\,A^{d_\eta}\log A,
\]
with
\[
A=\frac{\exp\,\!\bigl(2H_c(M)\bigr)\,\widetilde L_S\!\bigl(\widetilde G(M)\bigr)\,(\mathsf L_a\vee\mathsf L_b(M))\,\mathrm{diam}_\eta}{\varepsilon},
\]
so that $\sup_{\eta\in K_\eta}\mathrm{KL}\bigl(q_T(\cdot\mid\eta)\,\|\,p(\cdot\mid\eta)\bigr)\le\varepsilon$.
This completes the proof.

\end{proof}

\subsection{Unconstrained Monotonic Neural Networks}\label{sec:umnn}

We next state and prove the next result on uniform KL rates for Unconstrained Monotonic Neural Networks \citep[UMNN;][]{wehenkel2019unconstrained} where the conditional distribution $q_T(\theta \given \eta)$ is specified via $T$ as in (\ref{eq:umnn}) with the functions $a$ and $g$ modeled as ReLU networks. 

\begin{theorem}[Rate, conditional UMNN]\label{th:umnn}
Let $p$ satisfy Assumption~\ref{asm:mono} and let $q_T$ be the law of $T(\eta,Z)$, $Z\sim N(0,1)$, where $T$
is the UMNN (\ref{eq:umnn}) with $a_{\vartheta_a},g_{\vartheta_g}$ ReLU networks.
For any thresholds $M^{*}>M_0$, $A^{*}>A_d$, $B^{*}>B_d$, the class (\ref{eq:umnn}) contains
a realization satisfying Assumption~\ref{asm:pm}(ii) with envelope constants $M^{*},A^{*},B^{*}$. Let
$R=R(M^{*},A^{*},B^{*},\alpha,c)$ be the constant of Theorem~\ref{thm:ckl-explicit} for these. For any
$\varepsilon>0$ put
\[
M=M(\varepsilon):=R\sqrt{-\log\varepsilon},\qquad
\widetilde L_S(M):=2M\,L_S(M),\qquad
\widetilde G(M):=M^{*}+Me^{A^{*}+B^{*}M}.
\]
Then there is a UMNN $q_T$ whose ReLU networks $a_{\vartheta_a},g_{\vartheta_g}$ have parameter
counts
\[
\begin{aligned}
\#\vartheta_a\le& C_a\Bigl(\tfrac{\widetilde L_S(\widetilde G(M))\,L_a\operatorname{diam}_\eta}{\varepsilon}\Bigr)^{d_\eta}
\log\tfrac{\widetilde L_S(\widetilde G(M))}{\varepsilon},\\
\#\vartheta_g\le& C_g\Bigl(\tfrac{\widetilde L_S(\widetilde G(M))\,\mathsf L_b(M)(\operatorname{diam}_\eta+2M)}{\varepsilon}\Bigr)^{d_\eta+1}
\log\tfrac{\widetilde L_S(\widetilde G(M))}{\varepsilon},
\end{aligned}
\]
such that
\[
\sup_{\eta\in K_\eta}\mathrm{KL}\bigl(q_T(\cdot\mid\eta)\,\|\,p(\cdot\mid\eta)\bigr)\le\varepsilon .
\]
Here $C_a,C_g$ depend only on $d_\eta$, and $L_a,\mathsf L_b(\cdot)$ are the local Lipschitz
constants of Lemma~\ref{lem:lips}.
\end{theorem}

\begin{proof}[Proof of Theorem \ref{th:umnn}]
It is enough to show that the class contains a member that satisfies Assumption \ref{asm:pm} as the rest follows from Theorem \ref{thm:ckl-explicit}. 
Define
\begin{equation}\label{eq: umnn defn}
\begin{split}
T_\vartheta(\eta,z)&=a_{\vartheta_a}(\eta)+\int_0^z \exp\!\big(g_{\vartheta_g}(\eta,t)\big)\,dt,\\
\qquad
a_{\vartheta_a}&=\mathrm{clip}(\tilde a_{\vartheta_a},-M^*,M^*),\\
\qquad
g_{\vartheta_g}&=\mathrm{clip}(\tilde g_{\vartheta_g},-A^*-B^*|z|,A^*+B^*|z|),
\end{split}
\end{equation}
where \(\tilde a_{\vartheta_a}\) is a ReLU network on \(K_\eta\), \(\tilde g_{\vartheta_g}\) is a ReLU network on \(K_M\), and $\vartheta = (\vartheta_a,\vartheta_g)^\top$. The clips are pointwise in the displayed variables. Since clipping is one Lipschitz, it preserves continuity and does not increase moduli. 

\paragraph{Assumption \ref{asm:pm}$(i)$.} For every \((\eta,z,\vartheta)\) one has \(\partial_z T_\vartheta(\eta,z)=\exp\big(g_{\vartheta_g}(\eta,z)\big)>0\). Continuity in \((\eta,z,\vartheta)\) follows from continuity of \(\tilde a_{\vartheta_a}\), \(\tilde g_{\vartheta_g}\), exponential, and the integral in the upper limit. Differentiability in \(z\) and continuity of this partial derivative follow directly from the fundamental theorem of calculus, yielding $\partial_z T_\vartheta(\eta,z) = \exp(g_{\vartheta_g}(\eta,z))$.

\paragraph{Assumption \ref{asm:pm}$(ii)$.} Following definition \eqref{eq: umnn defn}, \(|T_\vartheta(\eta,0)|=|a_{\vartheta_a}(\eta)|\le M^*\) and \(|\log\partial_z T_\vartheta(\eta,z)|=|g_{\vartheta_g}(\eta,z)|\le A^*+B^*|z|\). This is precisely the global linear envelope. So, Assumption \ref{asm:pm}$(ii)$ is satisfied, and $R:=R(M^{*},A^{*},B^{*},\alpha,c)$  is well-defined. 

\paragraph{Assumption \ref{asm:pm}$(iii)$.} Lastly, we establish part $(iii)$ with an explicit construction. Let $M:=M(\varepsilon)=R\sqrt{-\log \varepsilon}$. The target quantile map \(T_p\) satisfy Assumption~\ref{asm:mono} on \(K_M\). Define \(u_1(\eta):=T_p(\eta,0)\) and \(u_2(\eta,z):=\log\partial_z T_p(\eta,z)\). The definition yields constants \(M_0>0\) and \(A_d,B_d\ge0\) with
\[
\sup_{\eta\in K_\eta}|u_1(\eta)|\le M_0,
\qquad
\sup_{(z,\eta)\in K_M}|u_2(\eta,z)|\le A_d+B_d|z|.
\]
Choose thresholds \(M^*>M_0\), \(A^*>A_d\), \(B^*>B_d\).
Hence, if we construct \(\tilde a_{\vartheta_a},\tilde g_{\vartheta_g}\) so that
\[
\sup_{\eta\in K_\eta}\big|\tilde a_{\vartheta_a}(\eta)-u_1(\eta)\big|\le \varepsilon,
\qquad
\sup_{(z,\eta)\in K_M}\big|\tilde g_{\vartheta_g}(\eta,z)-u_2(\eta,z)\big|\le \varepsilon,
\qquad
0<\varepsilon\le \varepsilon_{\max},
\]
with \(\varepsilon_{\max}:=\min\{M^*-M_0,\ A^*-A_d,\ B^*-B_d\}\), then the clips are inactive on \(K_M\) and \(a_{\vartheta_a}=\tilde a_{\vartheta_a}\), \(g_{\vartheta_g}=\tilde g_{\vartheta_g}\) there. It follows that
\[
\sup_{\eta\in K_\eta}|a_{\vartheta_a}(\eta)-u_1(\eta)|\le \varepsilon,
\qquad
\sup_{(z,\eta)\in K_M}|g_{\vartheta_g}(\eta,z)-u_2(\eta,z)|\le \varepsilon.
\]

Assumption~\ref{asm:mono} supplies a Lipschitz moduli \(\mathsf L_a\) for \(u_1\) on \(K_\eta\) and \(\mathsf L_b(M)\) for \(u_2\) on $K_M$. That is, we approximate the true maps
\[
u_1(\eta)=T_p(\eta,0)\quad\text{on }K_\eta,
\qquad
u_2(\eta,z)=\log\partial_zT_p(\eta,z)\quad\text{on }K_M
\]
with quantitative budgets that depend on $\mathsf L_a$ and $\mathsf L_b(M)$.

Theorem 1 in \cite{yarotsky2017error} proves that for any $d \ge 1$,
a 1-Lipschitz function $f$ on $[0,1]^d$ with $\|f\|_\infty \leq 1 $ admits a ReLU network approximation with depth $O\left(\log\left(\frac{1}{\varepsilon}\right)\right)$
and number of weights $O\left(\varepsilon^{-d} \log\left(\frac{1}{\varepsilon}\right) \right)$ that achieves sup norm error at most $\varepsilon$ on $[0,1]^d$.
As $u_1$ is $\mathsf L_a$-Lipschitz on $K_\eta$ and $u_2$ is $\mathsf L_b(M)$-Lipschitz on $K_M$, we map each domain to the unit cube by the affine scaling $x \mapsto (x-x_0)/\operatorname{diam}$.
Lipschitz constants scale by the domain diameter, so the rescaled functions are
$\mathsf L_a\,\operatorname{diam}_\eta$ and $\mathsf L_b(M)(\operatorname{diam}_\eta+2M)$ Lipschitz on $[0,1]^d$
up to absolute constants. Applying the theorem of \cite{yarotsky2017error} and scaling the error back gives realizations
$\tilde a_{\vartheta_a}$ and $\tilde g_{\vartheta_g}$ with
\[
\|\tilde a_{\vartheta_a}-u_1\|_{\infty,K_\eta} \le \varepsilon,
\qquad
\|\tilde g_{\vartheta_g}-u_2\|_{\infty,K_M} \le \varepsilon,
\]
for depth $O(\log(1/\varepsilon))$, and parameter counts
\[
\begin{aligned}
\#\tilde a_{\vartheta_a} &\;\le\; C_a\Bigl(\tfrac{\mathsf L_a\,\operatorname{diam}_\eta}{\varepsilon}\Bigr)^{d_\eta}\log\Bigl(\tfrac{\mathsf L_a\,\operatorname{diam}_\eta}{\varepsilon}\Bigr),\\
\#\tilde g_{\vartheta_g} &\;\le\; C_g\Bigl(\tfrac{\mathsf L_b(M)(\operatorname{diam}_\eta+2M)}{\varepsilon}\Bigr)^{d_\eta+1}\log\Bigl(\tfrac{\mathsf L_b(M)(\operatorname{diam}_\eta+2M)}{\varepsilon}\Bigr).
\end{aligned}
\]
Here $C_a$ and $C_g$ are fixed constants depending only on $d_\eta$. 
This completes the verification of Assumption \ref{asm:pm} for UMNN. Then Theorem \ref{th:umnn} is proved by simply using $\varepsilon / O(\tilde L_S(\tilde G(M(\varepsilon)))$ which is the required scaling from Theorem \ref{thm:ckl-explicit}.
\end{proof}

\section{Proofs for rates in Table \ref{tab:envelopes}}\label{sec:proofrates}

We now derive the rates stated in Table \ref{tab:envelopes}. In all the examples, $K_\eta\subset\mathbb R$ is compact and all parameter maps
(location, scale, degrees of freedom, etc.) depend smoothly on $\eta$ and are thus uniformly (free of $M$) bounded. For simplicity of notation, we use $F_\eta(\cdot)$ to denote the conditional cdf $F_p(\cdot \given \eta)$.
Write
\[
Q_\eta(u)=F_\eta^{-1}(u),\qquad T(\eta,z)=Q_\eta(\Phi(z)),\qquad
G(M)=\sup_{\eta\in K_\eta,\ |z|\le M}|T(\eta,z)|.
\]
For $G(M)$ we evaluate upper envelopes of $Q_\eta$ at the \emph{Gaussian}
tail levels $\bar\Phi(M)=1-\Phi(M)\sim \phi(M)/M$.

\paragraph{Normal $N(\mu(\eta),\sigma^2(\eta))$.}
Assume $\mu,\sigma\in C^1(K_\eta)$ with $K_\eta$ compact and
$\underline\sigma:=\inf_{\eta\in K_\eta}\sigma(\eta)>0$.
Let $\overline\mu:=\sup_{\eta\in K_\eta}|\mu(\eta)|<\infty$ and
$\overline\sigma:=\sup_{\eta\in K_\eta}\sigma(\eta)<\infty$.

\emph{Quantile growth.} The normal quantile growth $G(M)$ is trivially $O(M)$. 

\emph{Spikiness.}
\[
\partial_\theta \log p(\theta\mid\eta)= -\frac{\theta-\mu(\eta)}{\sigma^2(\eta)} = O(M).
\]

\emph{Conditional perturbation.}
\[
\log p(\theta\mid\eta)=\mathrm{const}-\log\sigma(\eta)-\frac{(\theta-\mu(\eta))^2}{2\sigma^2(\eta)}.
\]
The $\mu$ sensitivity is linear in $(\theta-\mu)$ while the $\sigma$ sensitivity contains $(\theta-\mu)^2$,
hence on $|\theta|\le M$ one has $\sup_{\eta}\|\nabla_\eta \log p(\theta\mid\eta)\|=O(M^2)$, i.e.,  
$L_{CP}(M)=O(M^2)$.

\paragraph{Laplace $\mathrm{Lap}(\mu(\eta),b(\eta))$.}
Assume $\mu,b\in C^1(K_\eta)$ with $K_\eta$ compact and
$0<\inf_{\eta\in K_\eta} b(\eta)\le \sup_{\eta\in K_\eta} b(\eta)<\infty$.

\emph{Quantile growth.}
The Laplace tail is exponential:
$1-F_\eta(x)=\tfrac12\exp(-(x-\mu)/b)$ for $x\ge \mu$,
so $Q_\eta(1-\varepsilon)=\mu+b\log(1/(2\varepsilon))$.
With Gaussian base $u=\Phi(z)$, one has $1-\Phi(z)\asymp e^{-z^2/2}/z$,
hence $\log(1/(1-\Phi(z)))=O(z^2)$.
Therefore $T(\eta,z)=Q_\eta(\Phi(z))=O(M^2)$ on $|z|\le M$, so $G(M)=O(M^2)$.

\emph{Spikiness.}
$\log p(\theta\mid\eta)=\mathrm{const}-\log b-|\theta-\mu|/b$,
so $\partial_\theta\log p=-\mathrm{sgn}(\theta-\mu)/b$ is uniformly bounded.
Hence $L_S(M)=O(1)$.

\emph{Conditional perturbation.}
$\eta$ only enters through $\mu(\eta)$ and $b(\eta)$.
The $\mu$ sensitivity is bounded, while the $b$ sensitivity carries a factor
$|\theta-\mu|/b^2$, which is $O(M)$ on $|\theta|\le M$.
With bounded $\nabla_\eta\mu,\nabla_\eta b$ on compact $K_\eta$, this gives
$L_{CP}(M)=O(M)$.

\paragraph{Generalized Gaussian $\mathrm{GGD}(\mu(\eta),\alpha(\eta),\beta(\eta))$ with $1\le \beta(\eta)\le 2$.}
Assume $\mu,\alpha,\beta\in C^1(K_\eta)$ with $K_\eta$ compact,
$0<\inf_\eta \alpha(\eta)\le \sup_\eta \alpha(\eta)<\infty$, and
$1\le \underline\beta:=\inf_\eta \beta(\eta)\le \sup_\eta \beta(\eta)=:\overline\beta\le 2$.

\emph{Quantile growth.}
The upper tail satisfies
\[
1-F_\eta(x)\sim \frac{1}{2\,\Gamma(1/\beta(\eta))}\,
\Big(\frac{x-\mu(\eta)}{\alpha(\eta)}\Big)^{1-\beta(\eta)}
\exp\!\Big(-\Big(\frac{x-\mu(\eta)}{\alpha(\eta)}\Big)^{\beta(\eta)}\Big),
\qquad x\to\infty,
\]
so $Q_\eta(1-\varepsilon)\asymp \mu(\eta)+\alpha(\eta)\{\log(1/\varepsilon)\}^{1/\beta(\eta)}$.
With Gaussian base $u=\Phi(z)$, $\log(1/(1-\Phi(z)))=O(z^2)$, hence on $|z|\le M$,
$T(\eta,z)=Q_\eta(\Phi(z))=O(M^{2/\underline\beta})$, so $G(M)=O(M^{2/\underline\beta})$.

\emph{Spikiness.}
$\log p(\theta\mid\eta)=\mathrm{const}-(|\theta-\mu|/\alpha)^{\beta}+\text{terms independent of }\theta$, so
$\partial_\theta\log p$ scales like $|\theta-\mu|^{\beta-1}/\alpha^{\beta}$.
Therefore on $|\theta|\le M$, $\sup_\eta|\partial_\theta\log p|=O(M^{\overline\beta-1})$, i.e.\ $L_S(M)=O(M^{\overline\beta-1})$.

\emph{Conditional perturbation.}
$\eta$ enters only through $\mu(\eta),\alpha(\eta),\beta(\eta)$.
The $\mu$ sensitivity is $O(M^{\beta-1})$, the $\alpha$ sensitivity is $O(M^\beta)$, and the $\beta$ sensitivity picks up an extra $\log M$ factor, $O(M^\beta\log M)$.
With bounded $\nabla_\eta\mu,\nabla_\eta\alpha,\nabla_\eta\beta$ on compact $K_\eta$, the worst case over $\beta\le\overline\beta$ gives $L_{CP}(M)=O(M^{\overline\beta}\log M)$.

\paragraph{Cauchy $C(\mu(\eta),\gamma(\eta))$.}
The density is
\[
p(\theta\mid\eta)=\frac{1}{\pi\,\gamma(\eta)}\frac{1}{1+\big((\theta-\mu(\eta))/\gamma(\eta)\big)^2}.
\]
Its upper tail satisfies, as $x\to\infty$,
\[
\bar F_\eta(x)=\Pr_\eta(\Theta>x)
=\frac{1}{\pi}\Big(\frac{\pi}{2}-\arctan\frac{x-\mu(\eta)}{\gamma(\eta)}\Big)
=\frac{\gamma(\eta)}{\pi(x-\mu(\eta))}\bigl(1+o(1)\bigr).
\]
Hence, as $\varepsilon\downarrow 0$,
\[
Q_\eta(1-\varepsilon)=\mu(\eta)+\frac{\gamma(\eta)}{\pi\,\varepsilon}\bigl(1+o(1)\bigr).
\]
For $Z\sim\mathcal N(0,1)$, Mills ratio gives $\bar\Phi(M)\sim \phi(M)/M$, so
\[
Q_\eta(\Phi(M))
= \mu(\eta)+\frac{\gamma(\eta)}{\pi\,\bar\Phi(M)}\bigl(1+o(1)\bigr)
=O\!\left(\frac{1}{\bar\Phi(M)}\right)
=O\!\left(M e^{M^{2}/2}\right),
\]
and one may take $G(M)=O\!\left(M e^{M^{2}/2}\right)$ (uniformly over $\eta$ on compact $K_\eta$ with $\gamma(\eta)\ge \underline\gamma>0$).
Moreover,
\[
\log p(\theta\mid\mu,\gamma)=\text{constant}+\log\gamma-\log\!\big(\gamma^2+(\theta-\mu)^2\big),
\]
so
\[
\partial_\theta \log p=-\frac{2(\theta-\mu)}{\gamma^2+(\theta-\mu)^2},\qquad
\partial_\mu \log p=\frac{2(\theta-\mu)}{\gamma^2+(\theta-\mu)^2},\qquad
\partial_\gamma \log p=\frac{1}{\gamma}-\frac{2\gamma}{\gamma^2+(\theta-\mu)^2},
\]
and $\big|\partial_\theta \log p\big|\le 1/\gamma$, $\big|\partial_\mu \log p\big|\le 1/\gamma$, $\big|\partial_\gamma \log p\big|\le 1/\gamma$.

\paragraph{Student's-$t_{\nu(\eta)}$ with location $\mu(\eta)$ and scale $s(\eta)$.}
Assume $\inf_{\eta\in K_\eta}s(\eta)>0$ and $0<\nu_{\min}:=\inf_{\eta\in K_\eta}\nu(\eta)$.

\emph{Quantile growth.}
The $t_\nu$ tail is polynomial: $\Pr(T>x)\asymp x^{-\nu}$ as $x\to\infty$, hence for $X=\mu+sT$,
$\Pr(X>x)\asymp x^{-\nu(\eta)}$ uniformly up to constants.
Therefore $Q_\eta(1-\varepsilon)\asymp \varepsilon^{-1/\nu(\eta)}$, and with Gaussian base
$\varepsilon=1-\Phi(M)\asymp \phi(M)/M$,
\[
G(M)=O\!\left(\left(\frac{M}{\phi(M)}\right)^{1/\nu_{\min}}\right)
=O\!\left(\left(M e^{M^2/2}\right)^{1/\nu_{\min}}\right).
\]

\emph{Spikiness.}
\[
\partial_\theta\log p(\theta\mid\eta)
=-\frac{(\nu(\eta)+1)(\theta-\mu(\eta))}{\nu(\eta)s^2(\eta)+(\theta-\mu(\eta))^2}
\]
is uniformly bounded (using $\inf_\eta s(\eta)>0$), so $L_S(M)=O(1)$.

\emph{Conditional perturbation.}
Sensitivities in $\mu$ and $s$ are bounded on $|\theta|\le M$.
If $\nu$ depends on $\eta$, then $\partial_\nu\log p$ contains
$\log\!\big(1+(\theta-\mu)^2/(\nu s^2)\big)$, which is $O(\log M)$ on $|\theta|\le M$.
Thus, $L_{CP}(M)=O(\log M)$ when $\nu$ varies with $\eta$, and $L_{CP}(M)=O(1)$ if $\nu$ is $\eta$-independent.

\paragraph{Finite mixtures.}
Let $p(\theta\mid\eta)=\sum_{k=1}^K \pi_k(\eta)\,p_k(\theta\mid\eta)$ with $\pi_k(\eta)\ge0$ and $\sum_k\pi_k(\eta)=1$.
At extreme levels, the mixture tail is governed by the heaviest component tail: if components outside the heaviest--tail set are negligible and the heaviest ones are comparable, then
\[
\bar F(x\mid\eta)\asymp \max_k \bar F_k(x\mid\eta),
\]
uniformly in $\eta$. Consequently, the upper quantile is of the same order as the largest component quantile, so at Gaussian tail levels
\[
G(M)\asymp \max_k G_k(M).
\]

For derivatives, write the posterior weights
\[
\tau_k(\theta,\eta)=\frac{\pi_k(\eta)p_k(\theta\mid\eta)}{\sum_j \pi_j(\eta)p_j(\theta\mid\eta)},\qquad \sum_k\tau_k=1.
\]
Then $\partial_\theta\log p=\sum_k \tau_k\,\partial_\theta\log p_k$, hence
\[
L_S(M)\le \max_k L_{S,k}(M).
\]
Similarly,
\[
\partial_\eta\log p=\sum_k \tau_k\big(\partial_\eta\log p_k+\partial_\eta\log\pi_k\big),
\]
so
\[
L_{CP}(M)\le \max_k L_{CP,k}(M)+\sup_k\|\nabla_\eta \log \pi_k\|_{L^\infty(K_\eta)}.
\]
If $\pi_k\in C^1(K_\eta)$ and $\inf_{\eta,k}\pi_k(\eta)\ge \pi_{\min}>0$, then
$\sup_k\|\nabla_\eta \log \pi_k\|_{L^\infty(K_\eta)}<\infty$, so the mixture has the same $M$-growth rates as the worst component (up to an additive constant from the weights).

\subsection{Verification of Assumptions in NeVI-Cut Applications}\label{sec:asmver}

We briefly discuss how some of the assumptions can be verified when applying NeVI-Cut. The target conditional cut posterior is $p(\theta \given \eta, D_2) = p(D_2 \given \theta,\eta) \times p(\theta \given \eta) / A(\eta,D_2)$ where $A(\eta,D_2)=\int p(D_2 \given \theta,\eta) p(\theta \given \eta) d\theta$ is the normalizing constant. The Lipschitz modulus $L_S$ in Assumption \ref{asm:mono}(i) exists if both $p(D_2 \given \theta,\eta)$ and $p(\theta \given \eta)$ are continuously differentiable in $\theta$, which can be directly verified from the functional form of the likelihood and the prior. Assumption \ref{asm:mono}(ii) can be verified by justifying differentiation (in $\eta$) under integration (in $\theta$) to obtain the $\partial_\eta A(\eta,D_2)$. A sufficient condition is: choosing an independent positive prior for $\theta$, i.e., $p(\theta \given \eta) = p(\theta) > 0$, and showing that $\log p(D_2 \mid \theta,\eta)$ is $C^1$ in $\eta$, with envelopes $h_i(\theta)\ge 0, i=1,2$ ($\int h_i(\theta)\,p(\theta)\,d\theta<\infty$) satisfying
\[
\sup_{\eta}p(D_2\mid\theta,\eta)\le h_1(\theta),\, \sup_{\eta}\Bigl|p(D_2\mid\theta,\eta)\,\partial_\eta \log p(D_2\mid\theta,\eta)\Bigr|\le h_2(\theta)
\quad \text{for all }\theta.
\]

Assumption \ref{asm:mono}(iii)
holds for finite mixtures of many common families (e.g., Gaussian, Laplace, generalized Gaussian) as shown in Table \ref{tab:envelopes}, and therefore approximately holds for the large class of distributions that can be well approximated by such mixtures. The assumption can be verified by checking the tail behavior of both the likelihood and prior for $\theta$ and using Mills ratio, as we show in Section \ref{sec:proofrates} when deriving the rates of Table \ref{tab:envelopes}, and in the applications.
Similarly, Assumption \ref{asm:mono}(iv) on the tail thinness lower bound can be verified directly from the analytical expressions of the likelihood and prior.
Finally, since the theory is developed for a univariate $\theta$, it cannot be directly applied in applications involving a multivariate $\theta$. However, one can verify the assumptions for a univariate analog of the problem, as will be demonstrated in the data analysis.

\section{Simulation Study}
\label{sec:sim}

This section examines the performance of NeVI-Cut in simulation studies. Section~\ref{sec:express} compares theoretical predictions regarding the expressiveness of NeVI-Cut with empirical findings. Section~\ref{sec:propscore} evaluates NeVI-Cut in a propensity score simulation setting that is widely studied in the literature on cutting feedback. An additional simulation experiment under a misspecified prior that is widely studied in the literature \citep{Bayarri2009, jacob2017better, yu2023variational}, and compares NeVI-Cut with the openBUGS cut, the tempered cut of \cite{plummer2015cuts}, and the parametric variational cut of \cite{yu2023variational} is presented in Section~\ref{app:biased}.

\subsection{Expressiveness of NeVI-Cut}\label{sec:express}

\begin{figure}[!b]
\centering
\begin{subfigure}[b]{0.45\textwidth}
    \centering
    \includegraphics[width=\textwidth]{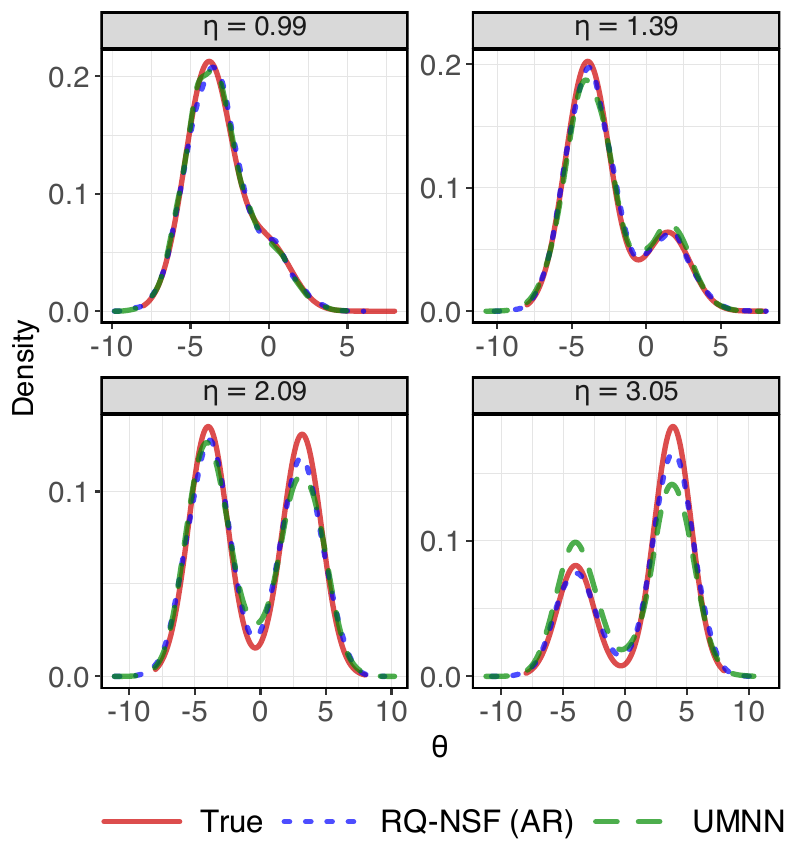}
    \caption{Conditional density}
    \label{fig:mono_conditional}
\end{subfigure}
\hfill
\begin{subfigure}[b]{0.5\textwidth}
    \centering
    \includegraphics[width=\textwidth]{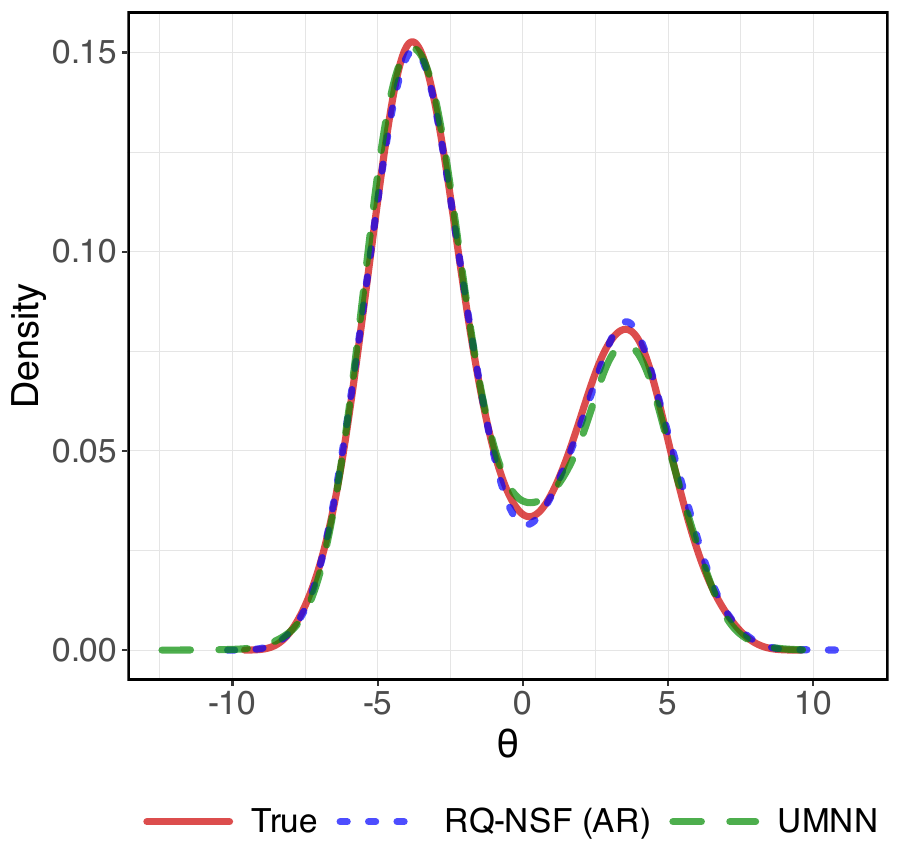}
    \caption{Marginal density}
    \label{fig:mono_marginal}
\end{subfigure}
\caption{Comparison of NeVI-Cut estimates using RQ-NSF(AR) and UMNN flows with the true conditional cut density $p(\theta \given \eta)$ (left) and marginal cut density $p(\theta)$ (right).}
\label{fig:monotone}
\end{figure}

This section illustrates NeVI-Cut's representational capacity for conditional distributions with complex features. For illustration, we consider a true joint density $p(\theta,\eta)$ with marginal $p(\eta)=\text{Gamma}(2,1)$ and conditional
\begin{equation*}
p(\theta \given \eta) = \pi(\eta) \,\, N \left(\mu_1(\eta),\sigma_1^2 \right) + \left(1-\pi(\eta)\right) \,\, N \left(\mu_2(\eta),\sigma_2^2 \right),
\end{equation*}
where $\pi(\eta) = 0.2 + 0.5\,\sigma\big(4(\eta - 2)\big)$, $\sigma (x) = 1/(1 + e^{-x})$, $\mu_1(\eta) = 4 \tanh(\eta - 1)$, $\mu_2(\eta) = -4 \tanh(\eta + 1)$, and $\sigma_1^2 = \sigma_2^2 = 1.5$. To mimic the NeVI-Cut setup, we draw random samples $\left\{\eta_i\right\}$ from $\text{Gamma}(2,1)$ and assume access only to these samples. The true $p(\theta \given \eta)$ displays varying multimodality and skewness across $\eta$ (red curves in Figure~\ref{fig:monotone}), which we aim to recover. Following Section~\ref{sec:theory}, we implement NeVI-Cut with two flow choices: UMNN \citep{wehenkel2019unconstrained} and RQ-NSF(AR) \citep{durkan2019neural}. In UMNN (\ref{eq:umnn}), we parameterize $a(\cdot)$ using a fully connected MLP with two hidden layers and $g(\cdot)$ using an MLP with three hidden layers, both with LeakyReLU activations. As the conditional distribution $\theta \given \eta$ is mixture normal with both the positive weights and mean being continuously differentiable functions of $\eta$, we know from Table \ref{tab:envelopes} that parts (i) - (iii) of Assumption \ref{asm:mono} are satisfied. Part (iv) is also satisfied as both components are normal and hence so is the mixture (conditional on $\eta$) thereby meeting the log-quadratic lower bound with $c$ and $\alpha$ depending on $\min_\eta \{\pi(\eta),1-\pi(\eta)\} > 0$ and $\max_\eta \{\mu_1(\eta),\mu_2(\eta)\} < \infty$.

Figure \ref{fig:monotone} compares the estimated conditional and marginal cut densities, $q_{\hat{\vartheta}}(\theta \given \eta)$ and $\hat p_{\text{cut}}(\theta)$, with the corresponding true densities $p(\theta \given \eta)$ and $p(\theta)$. To demonstrate the expressive power of NeVI-Cut, we consider $\eta \in \left\{0.99, 1.39, 2.09, 3.05 \right\}$ that induce distinct conditional densities (left panel). At $\eta=0.99$, the density is unimodal with a slight shoulder; at $\eta=1.39$, a weak secondary mode emerges. The remaining cases are clearly bimodal, with $\eta=2.09$ yielding roughly equal modes and $\eta=3.05$ yielding unequal modes. The marginal cut-posterior (right panel) is bimodal with unequal modes.

We see that NeVI-Cut is highly accurate with either flow family (UMNN or NSF). They closely match the true conditional and marginal cut densities, and demonstrate strong flexibility across density shapes. However, UMNN can be computationally expensive due to the evaluation of integrals. Given the near-identical accuracy of the two variants, we adopt the faster RQ-NSF (AR) for all subsequent analyses and simply refer to it as NeVI-Cut.

\subsection{Using Propensity Scores in Misspecified Outcome Model}\label{sec:propscore}

Propensity scores for treatment assignment are central to causal analysis in observational and other non-randomized settings, offering a principled way to adjust for measured confounding by balancing covariates.
In Bayesian formulations, joint modeling of treatment assignment and outcomes can allow undesirable feedback from a misspecified outcome model to the propensity score model, compromising causal conclusions \citep{zigler2016central}.
This motivates modular approaches that prevent the feedback, preserving propensity scores as a design-stage summary of treatment assignment while still propagating their uncertainty into downstream causal estimates.

Following \cite{jacob2017better}, we revisit the same setting of \cite{zigler2016central} to examine this issue. We study the effect of a binary treatment indicator $X$ on a binary outcome $Z$ in an observational setting, where individual-level covariates $\bc\in\bbR^p$ are available that may confound the relationship between $X$ and $Z$. Given observed data $\left\{ \left( X_i, Z_i, \bc_i \right) \right\}_i$ for $n$ individuals, we first model treatment assignment using a logistic regression
\begin{equation}\label{eq:treatment}
\text{logit} \, \bbP \left(X_i=1\given \bc_i\right)=\alpha_{10}+\sum_{j=1}^p \alpha_{1j} c_{ij}, \quad \forall i = 1,\dots,n
\end{equation}
with parameter $\alpha_1 = \left( \alpha_{10}, \dots, \alpha_{1p}\right)^\top$. The propensity score, defined as $e_i=\bbP \left(X_i=1\given \bc_i\right)$, summarizes the covariate-treatment relationship. 
Rather than modeling outcomes within propensity score strata, we simplify the outcome model by conditioning directly on the fitted propensity scores. The analysis model for the outcome is then 
\begin{equation}\label{eq:outcome}
\text{logit} \, \bbP(Z_i=1\given X_i,q_i)
=\theta_{20}+\theta_{21}X_i+\theta_{22}q_i,
\end{equation}
with parameters $\theta_2=(\theta_{20},\theta_{21},\theta_{22})^\top$. We assign independent priors $\alpha_{10},\theta_{20}\iid N(0,800)$ and $\alpha_{1j},\theta_{2j}\iid N(0,50)$ for all $j\ge 1$. 

For the true data generation process (DGP), we assume a correctly specified treatment assignment model with $\alpha_1^*=(0,0.1,0.2,0.3,0.4,0.5,0.6)^\top$, and a misspecified outcome model:
\begin{equation}\label{eq:outcometrue}
\text{logit} \, \bbP(Z_i=1\given X_i,\bc_i)=
\gamma_1^* c_{i1}+\gamma_2^* e^{c_{i2}-1}
+\gamma_3^* c_{i3}+\gamma_4^* e^{c_{i4}-1}
+\gamma_5^* \abs{c_{i5}} +\gamma_6^* \abs{c_{i6}},
\end{equation}
with $\gamma^*=(0.6,0.5,0.4,0.3,0.2,0.1)^\top$. We set $n=1000$, $p=6$, and $c_{ij}\iid N(0,1)$. 

The parameter of interest is the effect of the exposure $X$ on the outcome $Z$. This is  $\theta_{21}$ in the downstream analysis model (\ref{eq:outcome}). As in truth (Eq. \ref{eq:outcometrue}), $Z$ is independent of $X$ given $\bc$, we expect no treatment effect, i.e., the true effect size is $0$.
The performance metrics are thus defined in terms of the accuracy of the estimates of $\theta_{21}$ from the different methods relative to the true value. 

We first consider a simpler, smoother version of the problem (with a univariate downstream parameter) to see how the setup aligns with Assumption \ref{asm:mono}. We consider the outcome model $\text{logit} \, \bbP(Z_i=1\given X_i,e_i)
=\theta X_i+ f_\lambda(e_i), \quad \forall i\in{1,\ldots,n}$, where $f_\lambda$ is a flexible smooth function of the propensity scores, defined via a set of nuisance parameters $\lambda$, which we assume to be known, so that the downstream parameter is the univariate $\theta$. Here $\eta$ is the collection of propensity scores $\{e_i\}_i$. Assumption \ref{asm:mono}(i) is immediately met as the logistic regression likelihood and the normal prior are continuously differentiable. Assumption \ref{asm:mono}(ii) also holds because both the likelihood and the $\eta$-score of the log likelihood are uniformly bounded (logistic derivatives are bounded, and $f_\lambda$ is $C^1$ on compact $K_\eta$), which is sufficient as discussed in Section \ref{sec:asmver}. 
Since $p(\theta)$ has Gaussian tails (normal prior) and the likelihood is uniformly bounded above (by $1$) and below by an exponential decay (logistic likelihood), the tails behave like a Gaussian, thereby yielding a linearly bounded log-quantile derivative and satisfying the quantile bound \ref{asm:mono}(iii). Similarly, the tail thinness lower bound (iv) holds, as the tails of the log-posterior are bounded below by an exponential plus Gaussian decay, which is thicker than a Gaussian decay. Hence, all parts of Assumption \ref{asm:mono} are met for the smooth, univariate version of the problem. 

We return to the actual data analysis, where the downstream parameter set is multivariate, including both the parameter of interest, $\theta_{21}$, and the nuisance parameters, $\theta_{2j}, j \neq 1$, and where discretized propensity scores are used in the downstream model. We compare the marginal posterior distributions of $\theta_{21}$ from full Bayes, NeVI-Cut, and the multiple-imputation-based nested MCMC, Tempered-cut \citep{plummer2015cuts}, Gaussian VI-Cut \citep{yu2023variational} and SMI-Cut \citep{carmona2022scalable}. Multiple imputation-based nested MCMC and the full Bayes are run in \texttt{rstan}, while NeVI-Cut, Gaussian VI-Cut and SMI-Cut are implemented in Python 3.10. To evaluate each method, we compute the mean squared error (MSE) of the posterior mean, frequentist coverage and weighted interval score of 95\% credible interval, 
and continuous ranked probability score (CRPS). 
Frequentist coverage is the Monte Carlo proportion of times the 95\% credible interval contains the true value 0. The interval score evaluates the accuracy of the credible interval in covering the truth. The CRPS assesses the accuracy of the entire distribution relative to the true value.

\begin{table}[!t]
\centering
\caption{Performance comparison for estimating $\theta_{21}$, whose true value is $0$, in the propensity score simulation. We report the mean squared error (MSE) of the posterior mean, frequentist coverage and weighted interval score (WIS) of the 95\% credible interval, continuous ranked probability score (CRPS) of the marginal posterior, and average total runtime in seconds, including both upstream and downstream computations. Lower values of MSE, WIS, and CRPS indicate better performance.}
\label{tab:propensity}
\resizebox{\textwidth}{!}{
\begin{tabular}{lccccc}
\noalign{\hrule height 1.5pt}
\textbf{Method} 
& \textbf{MSE} 
& \makecell{\textbf{Frequentist}\\\textbf{Coverage}} 
& \textbf{WIS} 
& \textbf{CRPS} 
& \makecell{\textbf{Averaged Runtime}\\\textbf{(seconds)}} \\
\noalign{\hrule height 1.5pt}
Full Bayes 
& 0.052 (0.006, 0.227) 
& 0.790 
& 0.033 (0.015, 0.188) 
& 0.136 (0.052, 0.389) 
& 863 \\

Nested MCMC 
& 0.022 (0.003, 0.100) 
& 0.950 
& 0.018 (0.015, 0.044) 
& 0.083 (0.043, 0.234) 
& 3168 \\

Tempered-Cut 
& 0.022 (0.003, 0.102) 
& 0.960 
& 0.018 (0.015, 0.041) 
& 0.083 (0.043, 0.236) 
& 1054 \\

NeVI-Cut 
& 0.022 (0.003, 0.098) 
& 0.950 
& 0.018 (0.015, 0.031) 
& 0.083 (0.042, 0.230) 
& 47 \\

Gaussian VI-Cut 
& 0.022 (0.003, 0.095) 
& 0.970 
& 0.018 (0.016, 0.017) 
& 0.082 (0.045, 0.222) 
& 18 \\

SMI-Cut 
& 0.022 (0.003, 0.099) 
& 0.940 
& 0.018 (0.015, 0.033) 
& 0.083 (0.042, 0.232) 
& 521 \\
\noalign{\hrule height 1.5pt}
\end{tabular}
}
\end{table}

The results are summarized in Table~\ref{tab:propensity}. Outcome model misspecification adversely affects the full-Bayes posterior, yielding biased estimates and poor coverage, underscoring the need to cut feedback. In contrast, NeVI-Cut closely matches nested MCMC, the gold standard for cutting feedback, in terms of both accuracy and uncertainty quantification. The frequentist coverages for the cut-Bayes methods are approximately 96\%, which is close to the nominal level of 95\%. However, note that our theory only guarantees that the frequentist coverage of $95\%$ credible intervals from NeVI-Cut will approximate that from the true cut-posterior. In general, cut-posteriors do not guarantee correct frequentist coverage \citep{pompe2021asymptotics}, and so we do not expect this coverage to be well calibrated always.
Finally, NeVI-Cut offers a substantial computational advantage, with runtimes nearly 70 times lower than those of nested MCMC.

\subsection{Simulation Experiments under Misspecified Prior}\label{app:biased}

We revisit a commonly used illustrative example \citep{Bayarri2009, jacob2017better, yu2023variational} that demonstrates the challenges of full Bayesian inference when there is model misspecification or biased data. This example constitutes a case of a misspecified prior in the downstream analysis. The upstream dataset consists of a small sample $z = (z_1, \ldots, z_{n_1})^\top$, where $z_i \overset{\text{iid}}{\sim} N(\phi, 1)$.
The downstream dataset is a large sample $w = (w_1, \ldots, w_{n_2})^\top$, with $w_i \overset{\text{iid}}{\sim} N(\phi + \eta, 1)$.
This dataset has a mean $\phi + \eta$, where $\eta$ represents a downstream bias parameter. We use the same priors as in prior research, i.e., $\phi \sim N(0, \delta_1^{-1})$ and $\eta \sim N(0, \delta_2^{-1})$, with $\delta_2$ set to a large value to reflect high prior confidence in the absence of bias. When the true bias is large, this prior leads to a biased posterior under the full Bayesian model due to the overconfident assumption that $\eta \approx 0$.

This simulation setting illustrates how posterior samples from the small upstream dataset can be leveraged to inform inference on the bias parameter $\eta$ for the large dataset. Following previous studies, we set $n_1=100$ and $n_2=1000$, with true values $\phi = 0$ and $\eta = 1$. The inverse variance parameters are set to $\delta_1 = 1$ and $\delta_2 = 100$. For this setting, the full Bayes posterior and the cut-posterior
for $\eta$ both have closed-form expressions. We first derive these. 

Suppose we have the upstream analysis based on a relatively small sample size $n_1$, and a following downstream analysis with a large sample size $n_2$; the model is as follows:
\begin{align*}
 z_i & \overset{\text{iid}}{\sim} N(\phi, 1), \quad i = 1, \dots, n_1, \\
 w_i \given \phi, \eta & \overset{\text{iid}}{\sim} N(\phi + \eta, 1), \quad i = 1, \dots, n_2,
\end{align*}
with Gaussian priors 
\begin{align*}
 \phi \sim N(0, \delta_1^{-1}), \qquad \eta \sim N(0, \delta_2^{-1}).
\end{align*}
Let $S_z = \sum_{i=1}^{n_1} z_i, S_w = \sum_{i=1}^{n_2} w_i$.
\paragraph{Full Bayes Posterior.} The joint distribution of $(\phi,\eta)$ is given by
\begin{align*}
    p(\phi, \eta \given z, w) \,&\propto\, p(z \given \phi)\, p(w \given \phi, \eta)\, p(\phi) \,p(\eta)\, \\
    &\propto\,\exp \left(-\frac{1}{2}\left[\sum_{i=1}^{n_1}(z_i-\phi)^2 + \sum_{i=1}^{n_2}(w_i-\phi-\eta)^2\right]-\frac{\delta_1}{2}\phi^2-\frac{\delta_2}{2}\eta^2\right)\\
    &\propto\exp \left(-\frac{1}{2}\left[
(n_1 + n_2 + \delta_1) \phi^2 +
(n_2 + \delta_2) \eta^2 +
2n_2 \phi \eta
- 2\phi(S_z + S_w) - 2\eta S_w
\right]\right).
\end{align*}

Thus $(\phi,\eta)^\top$ follows a bivariate normal distribution: $(\phi, \eta)^\top \given z, w \sim N({\mu}, {\Sigma})$ where the covariance and mean vector are given as follows:
\begin{align*}
 {\Sigma} 
= \frac{1}{D}
\begin{pmatrix}
n_2 + \delta_2 & -n_2 \\
-n_2 & n_1 + n_2 + \delta_1
\end{pmatrix}, \quad D = (n_1 + n_2 + \delta_1)(n_2 + \delta_2) - n_2^2.
\end{align*}

The posterior mean vector is given by
\begin{align*}
 {\mu} = {\Sigma}
\begin{pmatrix}
S_z + S_w \\
S_w
\end{pmatrix}
= \frac{1}{D}
\begin{pmatrix}
(n_2 + \delta_2)(S_z + S_w) - n_2 S_w \\
- n_2 (S_z + S_w) + (n_1 + n_2 + \delta_1) S_w
\end{pmatrix}
\end{align*}

Explicitly,
\begin{align*}
 \mu_\phi = \frac{(n_2 + \delta_2) S_z + \delta_2 S_w}{(n_1 + n_2 + \delta_1)(n_2 + \delta_2) - n_2^2}, \quad
\mu_\eta = \frac{(n_1 + \delta_1) S_w - n_2 S_z}{(n_1 + n_2 + \delta_1)(n_2 + \delta_2) - n_2^2},
\end{align*}
and
\begin{align*}
 \operatorname{Var}_{\phi} = \frac{n_2 + \delta_2}{(n_1 + n_2 + \delta_1)(n_2 + \delta_2) - n_2^2}, \quad
\operatorname{Var}_{\eta} = \frac{n_1 + n_2 + \delta_1}{(n_1 + n_2 + \delta_1)(n_2 + \delta_2) - n_2^2}.
\end{align*}

\paragraph{True Cut Posterior.} Given the model settings, the cut-posterior for $\phi$ is defined as $p_{\text{cut}}(\phi \given z,w) = p(\phi \given z)$, where
${\phi} \given z$:
\begin{align*}
{\phi} \given z \sim N \left(
\frac{S_z}{n_1 + \delta_1}, \;
\frac{1}{n_1 + \delta_1}
\right).
\end{align*}

Conditional on $\phi$, the downstream posterior of $\eta$ is
\begin{align*}
\eta \given w, \phi \sim N \left(
\frac{S_w - n_2 \phi}{n_2 + \delta_2}, \;
\frac{1}{n_2 + \delta_2}
\right).
\end{align*}

Marginalizing over $\eta \given z$:
\begin{align*}
\mathbb{E}[\eta \given w]= \mathbb{E}[\mathbb{E}[\eta \given \phi, w]]=
\frac{(n_1 + \delta_1)S_w - n_2S_z}{(n_1 + \delta_1)(n_2 + \delta_2)}.
\end{align*}

The variance of the marginal posterior for $\eta$ is
\begin{align*}
    \operatorname{Var}({\eta} \given w)
&= \mathbb{E} \left[
\operatorname{Var}({\eta} \given w, {\phi})
\right]
+ \operatorname{Var} \left[
\mathbb{E}({\eta} \given w, {\phi})
\right]\\
 &= \frac{1}{n_2 + \delta_2} + \left( \frac{n_2}{n_2 + \delta_2} \right)^2 \, \frac{1}{n_1 + \delta_1}\\
& = \frac{1}{n_2 + \delta_2}
+ \frac{n_2^2}{(n_2 + \delta_2)^2 (n_1 + \delta_1)}.
\end{align*}

We can thus compare the marginal distribution of $\eta$ across four methods: the Full Bayes posterior, true cut-posterior,
NeVI-Cut estimate of the cut-posterior, and estimated cut-posterior using the Gaussian variational approximation of \citet{yu2023variational} ({\em Gaussian VI-Cut}).

We evaluate each method using 95\% weighted interval score (WIS), Continuous Ranked Probability Score \citep[CRPS;][]{gneiting2007strictly}, mean squared error (MSE), and 95\% coverage. CRPS measures the accuracy of the predictive distribution by comparing it to the observed value. The 95\% weighted interval score assesses the accuracy of prediction based on the quantiles of the predicted samples. The results are presented in Table \ref{tab:biased}.

 \begin{table}[h]
 \centering
 \caption{Predictive performance metrics of methods in simulation under a misspecified prior. Lower values of interval score, Continuous Ranked Probability Score (CRPS), and Mean Squared Error (MSE) indicate better performance.}
 \label{tab:biased}
 \resizebox{\textwidth}{!}{
 \begin{tabular}{lcccc}
\noalign{\hrule height 1.5pt}
 \textbf{Method} & \textbf{WIS} & \textbf{CRPS} & \textbf{MSE} & \textbf{Frequentist Coverage} \\
\noalign{\hrule height 1.5pt}
 Full Bayes & 0.387 (0.346, 0.480) & 0.481 (0.440, 0.574) & 0.275 (0.231, 0.378) & 0.00 \\
 \midrule
 True Cut & 0.016 (0.009, 0.090) & 0.077 (0.029, 0.214) & 0.017 (0.002, 0.072) & 0.84 \\
 \midrule
 Gaussian VI-Cut & 0.062 (0.003, 0.213) & 0.092 (0.024, 0.250) & 0.017 (0.001, 0.071) & 0.35 \\
 \midrule
 NeVI-Cut & 0.016 (0.009, 0.092) & 0.077 (0.028, 0.213) & 0.017 (0.001, 0.071) & 0.83 \\
\noalign{\hrule height 1.5pt}
 \end{tabular}}
 \end{table}

\section{Applications to Real Data}\label{app:real_data}
All analyses were conducted on the \href{https://jhpce.jhu.edu/}{Joint HPC Exchange cluster} utilizing the same CPU compute node 163 with an Intel Xeon Platinum 8558U processor.

\subsection{Epidemiological Study of HPV and Cervical Cancer}\label{sec:hpv}

In the upstream analysis, posterior samples of HPV prevalence are derived using prevalence data $(D_1)$ from 13 countries. Let $z = (z_1, \ldots, z_{13})^\top$, where $z_i$ denotes the number of individuals infected with high-risk HPV out of a sample of size $n_i$ in country $i$. The data are modeled as $z_i \con \gamma_i \sim \text{Binomial}(n_i, \gamma_i)$, where $\gamma_i$ is the parameter representing the true HPV prevalence in country $i$. Independent Beta priors are placed on each prevalence parameter, $\gamma_i \sim \text{Beta}(1, 1)$. By conjugacy, the posterior distribution is
$$
\gamma_i | z_i \sim \text{Beta}(1 + z_i, 1 + n_i - z_i), \quad i = 1, \ldots, 13.
$$  

Upstream posterior samples of $\gamma_i$ are then passed to the downstream analysis, which models the relationship between HPV prevalence and cervical cancer incidence. The downstream data ($D_2$) consists of $w = (w_1, \ldots, w_{13})^\top$ denoting the number of observed cervical cancer cases. The downstream analysis model specifies
$$
w_i \con \theta, \gamma_i \sim \text{Poisson} \left(T_i \exp(\theta_1 + \theta_2 \gamma_i)\right),
$$
where $T_i$ is the number of years at follow-up for country $i$, and $\theta = (\theta_1, \theta_2)^\top$ are regression parameters of interest. A flat multivariate normal prior $\theta \sim N \left( 0, 1000 I_2 \right)$ is used. Because the downstream model may be misspecified (cervical cancer rates depend on other individual factors as well), cutting feedback methods are particularly relevant in this setting. Thus the upstream parameters are $\eta=\{\gamma_i\}$ while the parameters in the downstream are $\theta = (\theta_1, \theta_2)^\top$.

Our target is the joint and marginal posterior distributions of $\theta = (\theta_1, \theta_2)^\top$. We compare the full Bayes and six different cut-posterior approaches: multiple imputation-based nested MCMC, {\em OpenBUGS-Cut} which obtains cut-posteriors via the cut function in OpenBUGS with adaptive Metropolis (1D adapter), the {\em tempered-Cut algorithm} of \citet{plummer2015cuts}, the parametric (Gaussian) VI-Cut method \citep{yu2023variational}, SMI-Cut (SMI with influence parameter fixed at 0) \citep{carmona2022scalable} and NeVI-Cut (using a fully autoregressive RQ-NSF).

\begin{figure}[!t]
\centering
\begin{subfigure}[b]{0.8\textwidth}
    \centering
    \includegraphics[width=\textwidth]{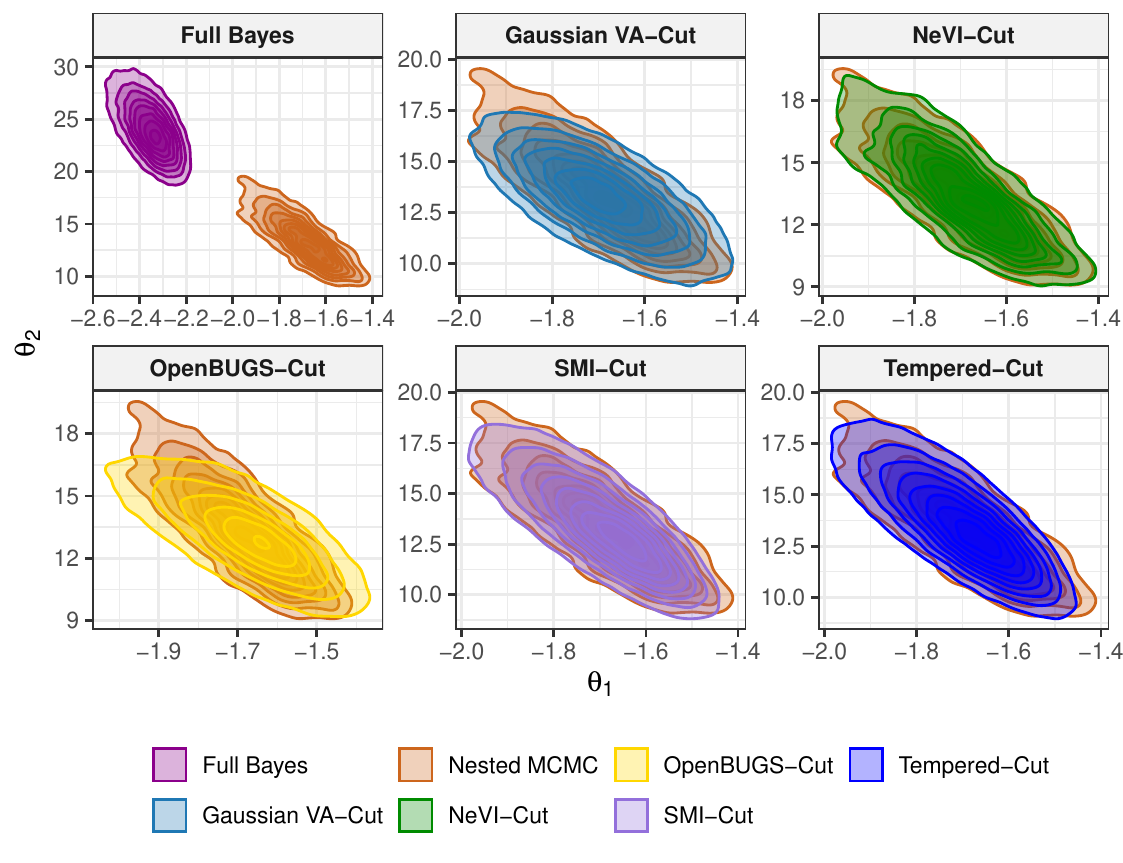}
    \caption{Joint posterior of $(\theta_1, \theta_2)$}
    \label{fig:hpv_joint_joint}
\end{subfigure}
\vfill
\begin{subfigure}[b]{\textwidth}
    \centering
    \includegraphics[width=\textwidth]{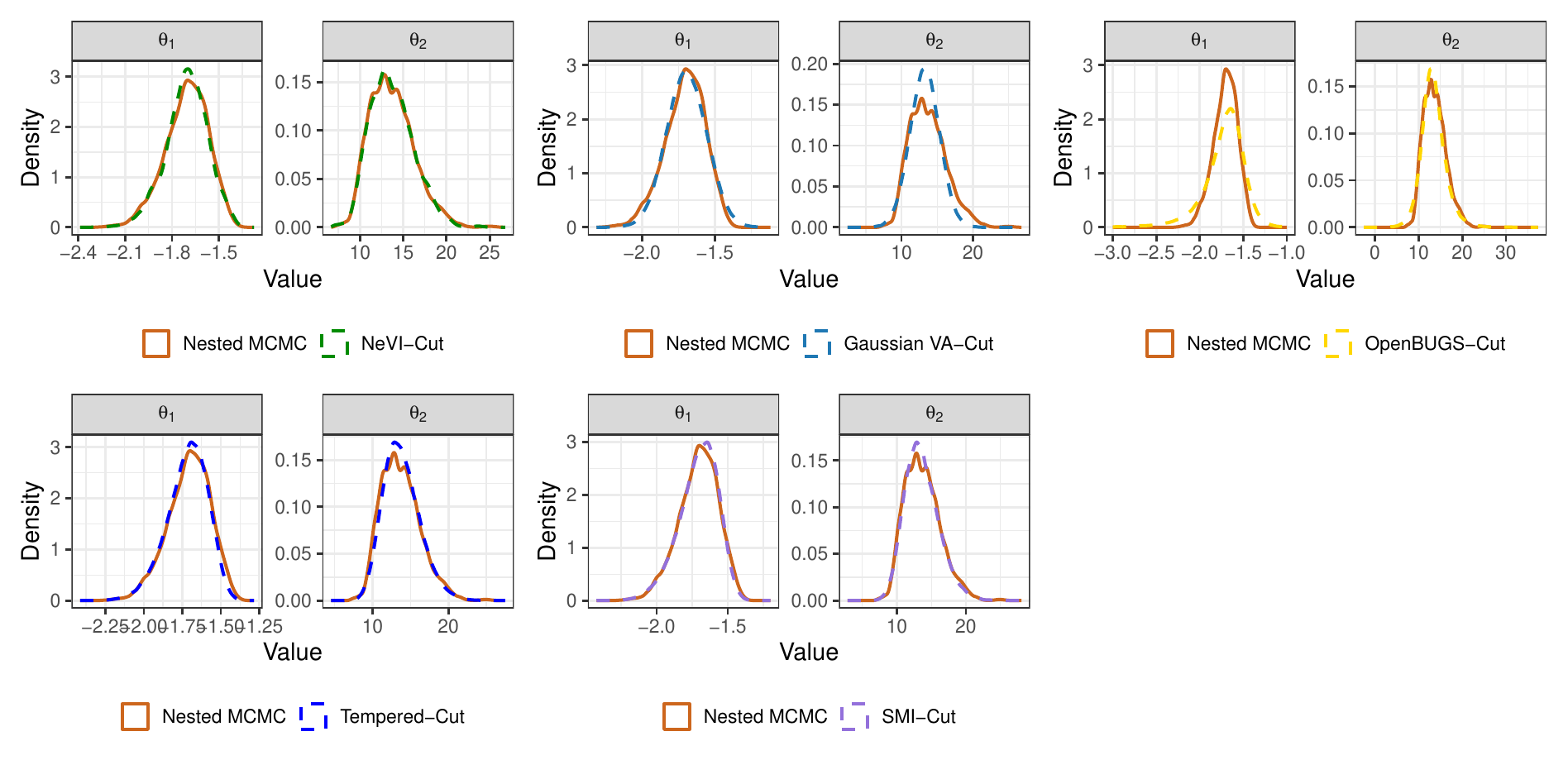}
    \caption{Marginal posterior of $(\theta_1, \theta_2)$}
    \label{fig:hpv_joint_marginal}
\end{subfigure}
\caption{\textbf{HPV Prevalence in Cervical Cancer.} Posterior distributions for the downstream model parameters $\theta = \left( \theta_{1} , \theta_2 \right)^\top$ in the HPV data analysis. The top panel shows their joint posteriors under full Bayes and different cut-Bayes methods compared with multiple imputation-based nested MCMC. The bottom panel presents their marginal posteriors using different cut-Bayes methods compared with multiple imputation-based nested MCMC.}
\label{fig:hpv_joint}
\end{figure}

The top panel of Figure~\ref{fig:hpv_joint} presents the joint posterior distributions of $\theta$ under the full Bayes and cut-Bayes models. The bottom panel shows pairwise comparisons against the baseline nested MCMC of the other five cut-Bayes methods. As noted in prior research \citep[such as][]{plummer2015cuts, yu2023variational}, the full Bayes and cut-posteriors differ substantially, underscoring the sensitivity of inference to model misspecification. The figures on the bottom panel show that all methods generally produce a reasonable approximation to the nested MCMC cut-posterior. This is not surprising as the cut-posterior is clearly unimodal and reasonably Gaussian-shaped. Overall, NeVI-Cut produces results that most closely align with nested MCMC.
Gaussian VI-Cut, SMI-Cut and tempered-Cut tend to overestimate the spike in $\theta_2$, while OpenBUGS-Cut underestimates the spike in $\theta_1$. 

We compute the empirical 2-Wasserstein distance between the joint posterior samples of $(\theta_{1}, \theta_{2})$ obtained from each method and those obtained from nested MCMC. Since solving the full optimal transport problem is computationally expensive for large posterior sample sizes, e.g., $100000$ draws, we estimated the Wasserstein distance using repeated random subsampling, with $B = 10$ repetitions and $m = 10000$ samples per repetition. We report the mean and standard deviation of the resulting Wasserstein distances, together with the running time for each method in Table \ref{tab:hpv_metrics}. All methods, except OpenBUGS, are run on the same node of a high-performance computing cluster (HPC). OpenBUGS is run on a local computer to satisfy the requirements of the BUGS package. We report both upstream and downstream computation times for nested MCMC, NeVI-Cut, and Gaussian VI-Cut when applicable, and the total computation time when the two components are not timed separately. Since the upstream posterior in the MCMC approach is available in closed form, the upstream computation time is negligible for Nested MCMC and NeVI-Cut. Consistent with the density plots, the full Bayes model has the largest Wasserstein distance from nested MCMC, while the cut-Bayes methods are closer to the reference distribution. NeVI-Cut achieves the smallest Wasserstein distance.
Sensitivity analyses for the model configuration are provided in Section~\ref{app:sensitivity}.

\begin{table}[ht] 
\centering
\caption{\textbf{HPV Prevalence in Cervical Cancer.} Runtime and posterior discrepancy relative to nested MCMC for the joint posterior of $\theta = (\theta_1, \theta_2)$. Wasserstein distances are computed using repeated random subsampling.} \label{tab:hpv_metrics}
\resizebox{\textwidth}{!}{
\begin{tabular}{lcccc}
\noalign{\hrule height 1.5pt}
\multirow{2}{*}{\textbf{Method}} & \multicolumn{2}{c}{\textbf{Runtime (in seconds)}} & \multicolumn{2}{c}{\textbf{Wasserstein distance}} \\
\cmidrule[1.5pt](lr){2-3} \cmidrule[1.5pt](lr){4-5}
& \textbf{Upstream} & \textbf{Downstream} & \textbf{Mean} & \textbf{SD} \\
\noalign{\hrule height 1.5pt}
Nested MCMC     & 0.02 & 88  & --    & -- \\
NeVI-Cut        & 0.02 & 73  & 0.119 & 0.005 \\
SMI-Cut \citep{carmona2022scalable}            & \multicolumn{2}{c}{771} & 0.167 & 0.017 \\
Tempered-Cut \citep{plummer2015cuts}   & \multicolumn{2}{c}{382} & 0.215 & 0.014 \\
OpenBUGS-Cut    & \multicolumn{2}{c}{208} & 0.550 & 0.053 \\
Gaussian VI-Cut \citep{yu2023variational} & 3    & 63  & 0.779 & 0.021 \\
Full Bayes      & \multicolumn{2}{c}{34}  & 10.60 & 0.014 \\
\noalign{\hrule height 1.5pt}
\end{tabular}
}
\end{table}

\subsection{Linguistic Profiles}
\label{app:LP}

In this section, we provide implementation details for reproducing the linguistic profile (LP) validation analysis in
Section~\ref{sec:realdata} of the main article.
We denote the SMI influence parameter by $\kappa$.

\subsubsection{Dividing Dataset into Training and Validation Sets}

The full dataset \texttt{coarsen\_all\_items} is available with
\citet{battaglia2025amortising}'s \href{https://github.com/llaurabat91/amortising-variational-meta-posteriors}{GitHub} repository.  Following their validation pipeline in Section~6.2, we model 120 anchor and 247 floating profiles including all $71$ items, partition the 120 anchor profiles into $80$ train and $40$ test/held-out sets for validation, and induce an $11 {\times}11$ grid for the
shared latent Gaussian-process basis.  The $40$ held-out anchor LPs
are exactly \citet{battaglia2025amortising}'s anchor-validation set. They
are stored in the \texttt{LP\_anchor\_val} field of every output
\texttt{.npz} we generate. The split is fixed by seeds in
\texttt{configs/all\_items\_SMI\_amortized.py}, so SMI-Cut and NeVI-Cut are evaluated on the same 40 held-out profiles.

\subsubsection{Reproducing SMI-Cut}

Statistical modeling details for this application are detailed in \citet{battaglia2025amortising} and in the main article; where upstream parameters are $\eta=(\mu=\{\mu_i\},\zeta=\{\zeta_i\},a=\{a_i\},W=\{w_{if,b}\},\Gamma=\{\gamma_b\})$ and parameters in the downstream are the unknown locations $\theta=X_M=\{x_p\}_{p \,\in \,M}$. We use \citet{battaglia2025amortising}'s training driver
\texttt{train\_vmp\_flow\_allhp\_smallcondval.py} available from their \href{https://github.com/llaurabat91/amortising-variational-meta-posteriors}{GitHub} repository.
Following their choice,
the neural architecture is a 6-layer RQ-NSF with coupling flow with $10$ bins and conditioner MLPs of hidden sizes 30 and depth 5. The priors are $\calN \left( 0,I_{d_\eta} \right)$ base distribution on $\eta$, $\text{Uniform} \left( [0,1] \times [0,1] \right)$ base distribution on $X_M$. The unit-square output $\theta$ is then pushed through a blockwise affine to the geographic rectangle $[0,1] \times [0, 0.89]$ for each pair. The
five-dimensional context
$c=(\sigma_a,\sigma_w,\sigma_k,\ell_k,\kappa)$ stacks the four hyperparameters $\psi = (\sigma_a, \sigma_w, \sigma_k, \ell_k)$ with the SMI influence parameter $\kappa$.  The
hyperpriors are
\begin{gather*}
\sigma_a \sim \text{Gamma}(10,1), \quad \sigma_w \sim \text{Gamma}(5,1), \quad \sigma_k \sim \text{Uniform}(0.1,0.4),\\
\ell_k \sim \text{Uniform}(0.2,0.5), \quad
\text{and} \quad \kappa\sim\mathrm{Beta}(0.5,0.5) .
\end{gather*}
By specifying a prior on the influence parameter, SMI yields a collection of posteriors, indexed by $\kappa$. We subsequently set $\kappa=0$ to draw samples from the cut-posterior, as detailed below. 

Training runs are conducted for $100,000$ ELBO
steps with $3$ ELBO Monte-Carlo samples per step, $5$
$\gamma$-profile samples, and the AdaBelief optimiser with
global-norm gradient clipping at $1.0$ and a
warmup-exponential-decay learning-rate schedule (warmup $3{,}000$
steps, transition $10{,}000$ steps, peak $3\times 10^{-4}$, decay
rate $0.6$).  We deviate from the released config in only two places:
$(i)$~we set \texttt{use\_wandb=False} to remove the dependency on the
Weights~\&~Biases service, and $(ii)$~we write \texttt{peak\_value} and
\texttt{decay\_rate} directly into \texttt{lr\_schedule\_kwargs}. The
values come from \texttt{fixed\_configs\_wandb} in
\texttt{train\_vmp\_flow\_allhp\_smallcondval.py}, and
the resulting LR schedule is identical to \cite{battaglia2025amortising}'s.
The seed is fixed at $1$.  Checkpoints are written every $20{,}000$
steps.  The latest checkpoint is then used to draw $10{,}000$
posterior samples of $X_M$ via \texttt{sample\_at\_c.py}, at the
``optimized bound''
$\psi = (\sigma_a, \sigma_w, \sigma_k, \ell_k) = (10,5,0.4,0.2)$
recorded in \texttt{prior\_hparams\_plot\_optim} of
\texttt{configs.yaml}, with $\kappa = 0$ for the full cut. The posterior samples are saved as \texttt{samples\_c\_5\_10\_0.4\_0.2\_eta0.0.npz}.
Total wall-clock time was approximately 8.08 hours, recorded in
\texttt{timing\_info.txt}.  Per-step negative ELBO is recovered
post-hoc from the TensorBoard event files via our
\texttt{extract\_smi\_loss.R}, which calls 
\texttt{diagnostics/extract\_smi\_loss\_from\_tb.py} and
caches the result in the SMI working directory as \texttt{smi\_loss\_trace.npz}.

\subsubsection{Implementing NeVI-Cut for Downstream Analysis}

NeVI-Cut only requires samples from the upstream posterior (obtained through any upstream modeling of the anchor data) together with the downstream likelihood and prior.
As the purpose of this analysis is primarily to compare the posteriors of locations of floating profiles between SMI-Cut and NeVI-Cut, to run the latter, we simply utilize the 10,000 posterior samples of $\eta$ already available from SMI-Cut (as the cut posterior for $\eta$ is the same as its upstream posterior).

Conditional on $\eta$, the
NeVI-Cut downstream flow $q_\vartheta(\theta\mid\eta)$ is modeled using the same flow architecture as SMI-Cut with $\text{Uniform} \left( [0,1] \times [0,1] \right)$ base distribution on $\theta$. The unit-square output $\theta$ is similarly mapped to the geographic rectangle $[0,1] \times [0, 0.89]$ for each pair. The full configuration is available at
\texttt{flow\_config.json}. Training minimizes the negative ELBO using Adam (default JAX/Optax
hyperparameters) with mini batch sizes matching the SMI per-step Monte-Carlo budget (3 upstream $\eta$-samples, 5 $\gamma$-profile samples, and 1 $Z$-sample). We track the negative ELBO at every step and write the full per-step trace to \texttt{training\_loss.npy}.  Early stopping is implemented in \texttt{run\_nevi\_cut.py} after every $\mathtt{log\_every}$ steps where the driver computes the trailing mean of negative ELBO of the last $\mathtt{log\_every}$ steps and triggers stopping if it fails to improve by \texttt{early\_stopping\_tol} for \texttt{early\_stopping\_patience} consecutive log events.  We use $\mathtt{log\_every}=100$, $\mathtt{early\_stopping\_tol}=10^{-3}$, and desired $\mathtt{early\_stopping\_patience}$. To compare with SMI-Cut, we disable early stopping by setting $\mathtt{early\_stopping\_patience}=0$ and trace 100,000 steps. We independently repeat training across four seeds
$\{0,1,2,3\}$.  After training, each flow draws
10,000 posterior samples of $\theta$, written to \texttt{samples\_for\_R.npz} and \texttt{samples\_XM.npz} for the larger $X_M$-only array.

\subsubsection{Held-out Evaluation}

Given 10,000 posterior samples of locations for 40 held-out profiles from both methods, we treat their observed locations as the truth and compute the 2-Wasserstein distance using the \href{https://cran.r-project.org/package=transport}{\texttt{transport}} R package
\citep{transport2024}, root mean squared error (RMSE) of the posterior mean, the energy score or multivariate continuous ranked probability score (CRPS) using the \href{https://cran.r-project.org/package=scoringRules}{\texttt{scoringRules}} R package \citep{scoringRules2019}. All metrics are computed by directly passing the output in \texttt{evaluate\_held\_out.R}. The output is available as \texttt{eval/eval.csv} in the working directory, with one row per (method, profile) pair.  An aggregated
per-method summary over different seeds is written to \texttt{eval/eval\_summary.csv} using \texttt{evaluate\_seeds.R}.

\begin{figure}[!ht]
\centering
\begin{subfigure}[b]{\textwidth}
    \centering
    \includegraphics[width=.99\textwidth]{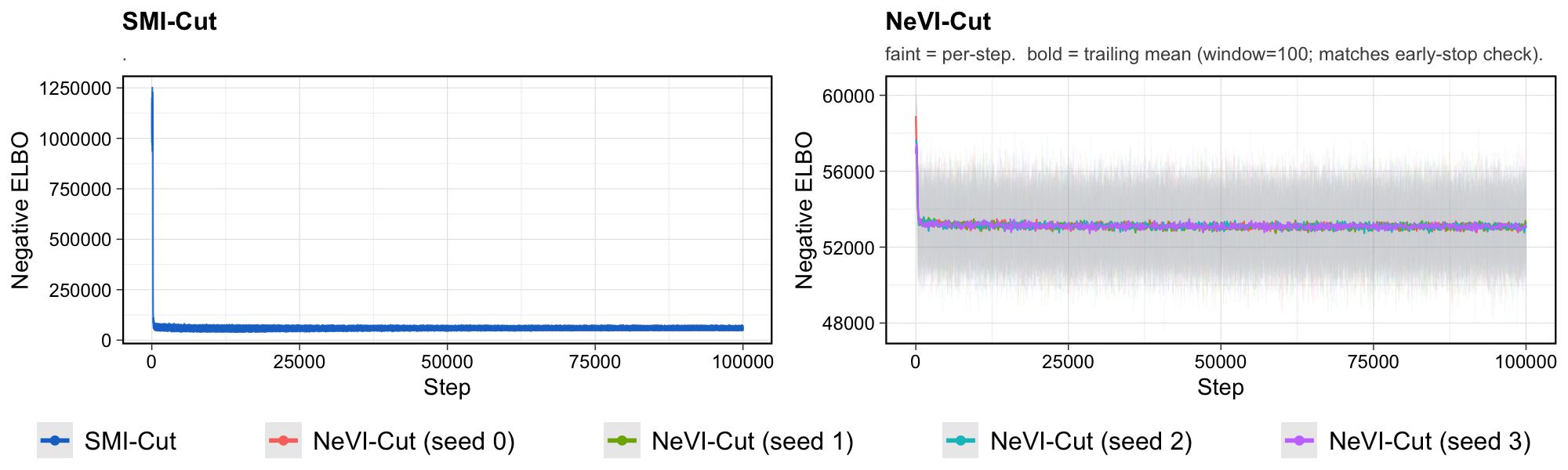}
    \caption{Negative ELBO convergence}
    \label{fig:LP seed nelbo}
\end{subfigure}
\vfill
\begin{subfigure}[b]{\textwidth}
    \centering
    \includegraphics[width=.99\textwidth]{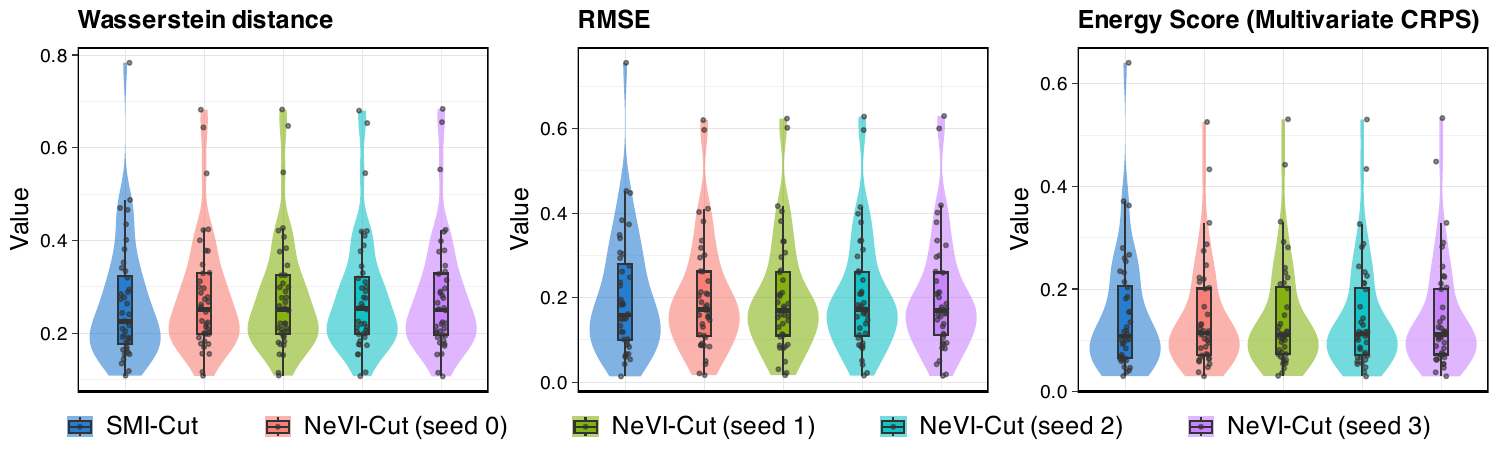}
    \caption{Performance diagnostics}
    \label{fig:LP seed diagnostic}
\end{subfigure}
\vfill
\begin{subfigure}[b]{\textwidth}
    \centering
    \includegraphics[width=.99\textwidth]{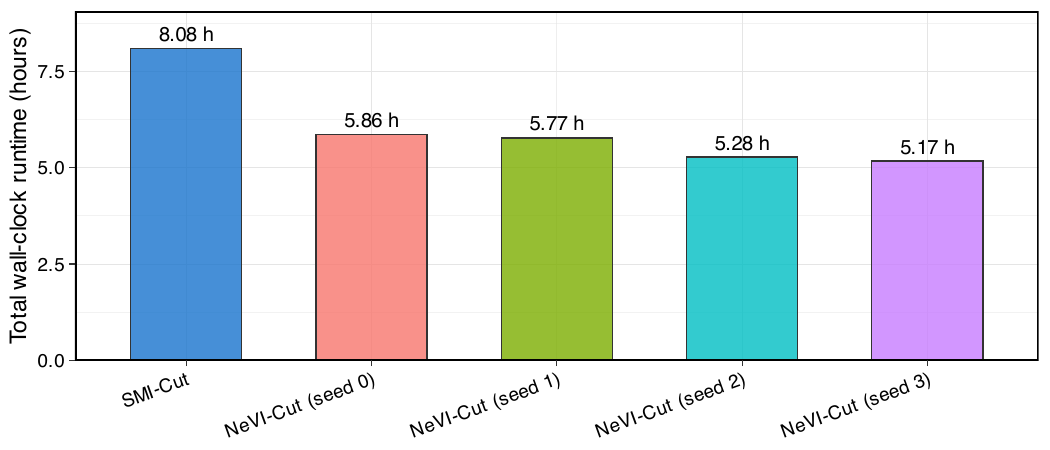}
    \caption{Runtime}
    \label{fig:LP seed runtime}
\end{subfigure}
\caption{\textbf{Linguistic Profile Analysis.} This is a summary of linguistic profile validation analysis.
\ref{fig:LP seed nelbo} presents negative ELBO across optimization steps. Faint lines in NeVI-Cut indicate each step, while solid lines are the trailing means. \textit{Note: The target posteriors in SMI-Cut and NeVI-Cut are different, so their negative ELBOs are not comparable.} \ref{fig:LP seed diagnostic} presents box and violin plots of the 2-Wasserstein distance of the posterior of $\theta$, the root mean squared error (RMSE) of the posterior mean of $\theta$, and the energy score of the posterior of $\theta$. The \ref{fig:LP seed runtime} reports the total wall clock time, including training and sampling. \textit{Note:
The runtimes reported here are not directly comparable as SMI-Cut produces an entire collection of SMI posteriors indexed by $\kappa$ while the NeVI-Cut is using the upstream posterior samples of $\eta$ from the SMI-Cut run.}}\label{fig:LP seed}
\end{figure}

\subsubsection{Diagnostics and Figures}

\paragraph{Average Performance Over Multiple Seeds and Held-Out Profiles.} Figure~\ref{fig:LP seed} provides an overview of the convergence and performance diagnostics for NeVI-Cut across the four seeds, and compares them with SMI-Cut. Figure~\ref{fig:LP seed nelbo} illustrates the convergence of the negative ELBO with increasing optimization steps. In the case of NeVI-Cut, the faint lines depict trace plots for each step, while the solid lines represent trailing means, indicating a favorable convergence of the negative ELBO. It is important to note that the target posteriors for SMI-Cut and NeVI-Cut differ; thus, they are not directly comparable, each having their own vertical axis, and are presented here for illustrative purposes.

Based on the 40 held-out profiles, Figure~\ref{fig:LP seed diagnostic} presents a comparison of the validation metrics, namely, Wasserstein distance, RMSE, and energy score. The average performance metrics across the seeds for NeVI-Cut are compiled in Table~\ref{tab:LP diagnostic table}. Collectively, these findings suggest that all methods yield comparable validation measures, indicating that they estimate the posterior distribution of $\theta$ with equal effectiveness.

Figure~\ref{fig:LP seed runtime} illustrates the runtimes, encompassing both the training time of the flows and the subsequent sampling from the fitted flow. SMI-Cut is a joint model of both components, whereas NeVI-Cut is a downstream model; therefore, their runtimes are not directly comparable. The provided runtimes support reproducibility and transparency in reporting. The per-seed wall-clock times, as documented in \texttt{runtime.json}, were $5.86,5.77,5.28,5.17$ hours, respectively, with $99.6\%$ attributed to training, while the remaining time of approximately $80$ seconds was spent on sampling. Despite this, it is encouraging to note the observed reduction in runtime for NeVI-Cut.

\begin{table}[!t]
\centering
\caption{\textbf{Linguistic Profile Analysis.} Average Diagnostic Summary Across Seeds in Linguistic Profile Validation}
\label{tab:LP diagnostic table}
\begin{tabular}{lccc}
\noalign{\hrule height 1.5pt}
\textbf{Method} 
& \textbf{Wasserstein Distance} & \textbf{RMSE} & \textbf{Energy Score} \\
\noalign{\hrule height 1.5pt}
SMI-Cut 
& 0.266 
& 0.203 
& 0.150 \\
\noalign{\hrule}

NeVI-Cut 
& 0.280 
& 0.202 
& 0.146 \\
\noalign{\hrule height 1.5pt}
\end{tabular}
\end{table}

\begin{figure}[!b]
\centering
\includegraphics[width=\textwidth]{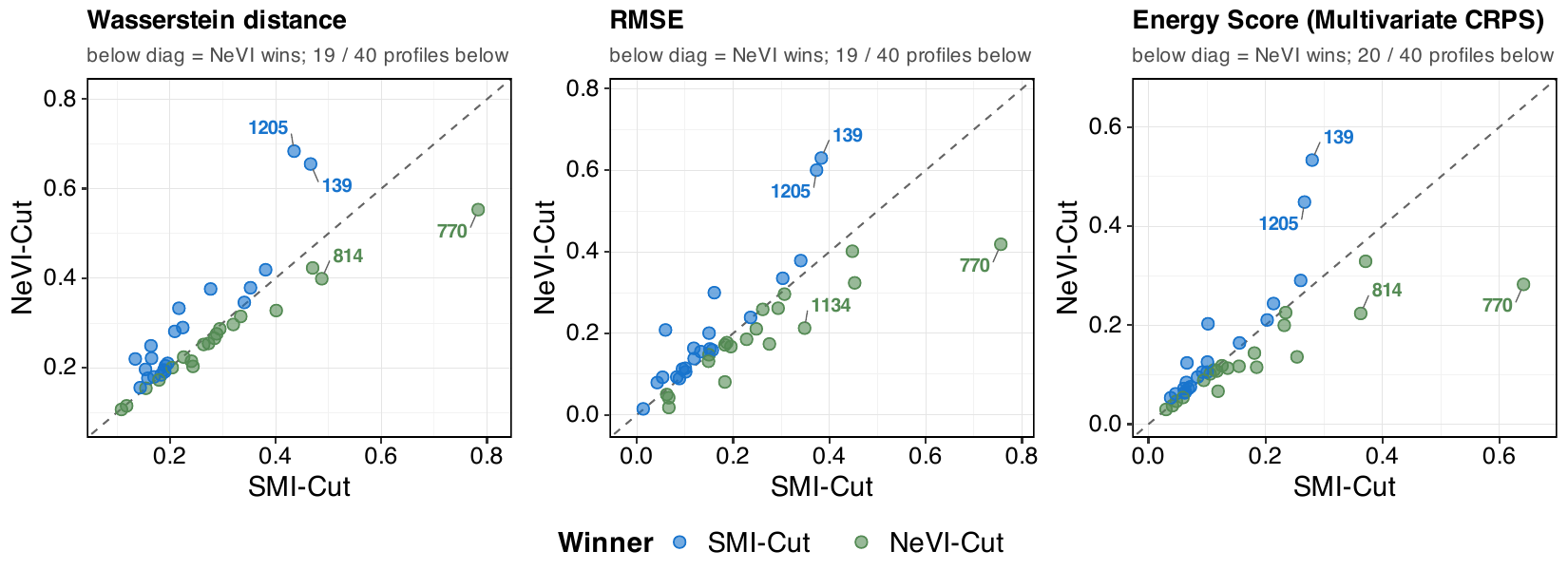}
\caption{\textbf{Linguistic Profile Analysis.} This compares diagnostic measures in Figure~\ref{fig:LP seed} for 40 held-out profiles based on SMI-Cut and NeVI-Cut with seed 3.}\label{fig:LP seed 3 diagnostic}
\end{figure}

\begin{figure}[!t]
\centering
\begin{subfigure}{\textwidth}
    \centering
    \includegraphics[width=.98\textwidth]{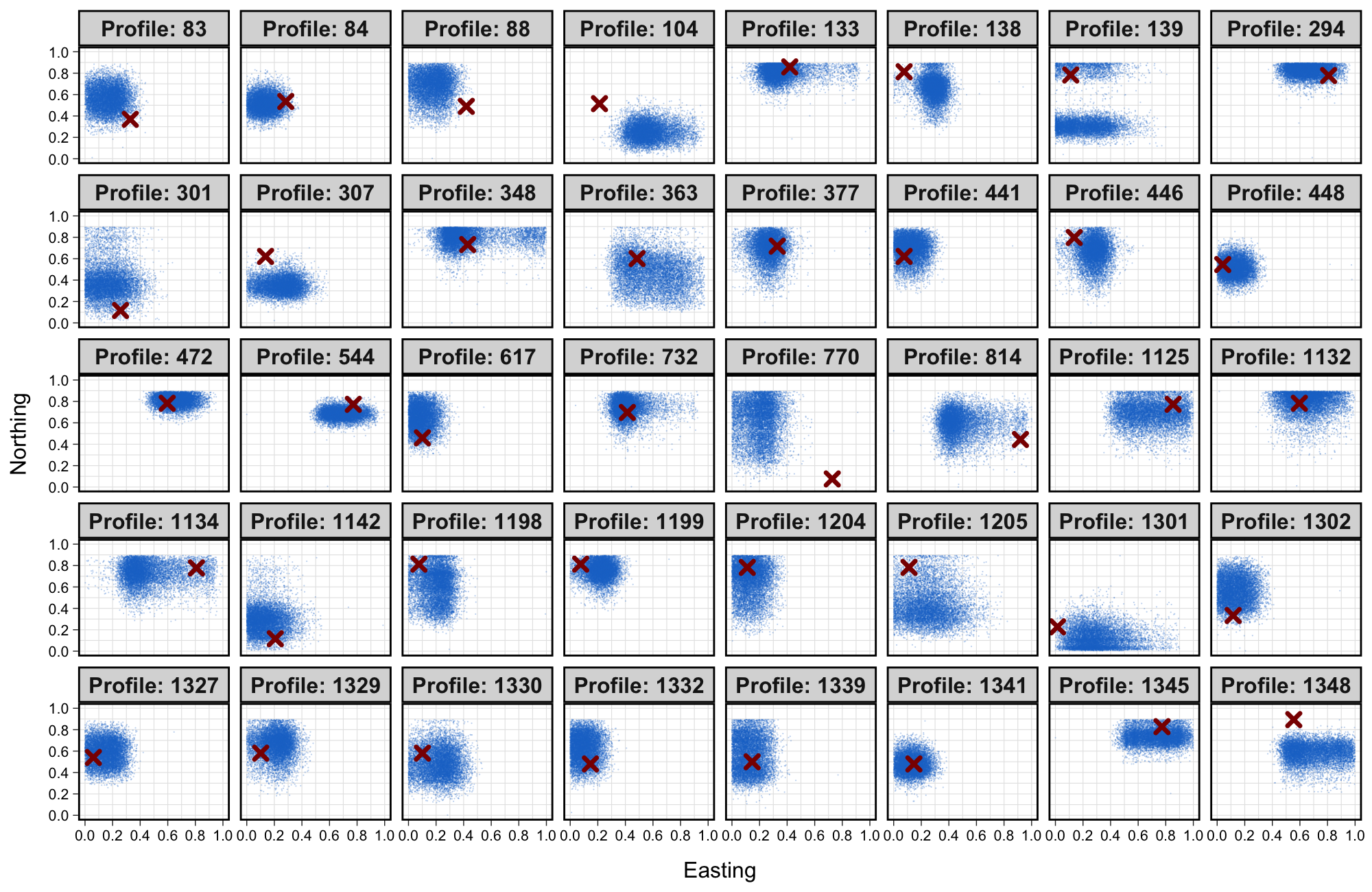}
    \caption{SMI-Cut}
    \label{fig:LP seed 3 location smi-cut}
\end{subfigure}
\vfill
\begin{subfigure}{\textwidth}
    \centering
    \includegraphics[width=.98\textwidth]{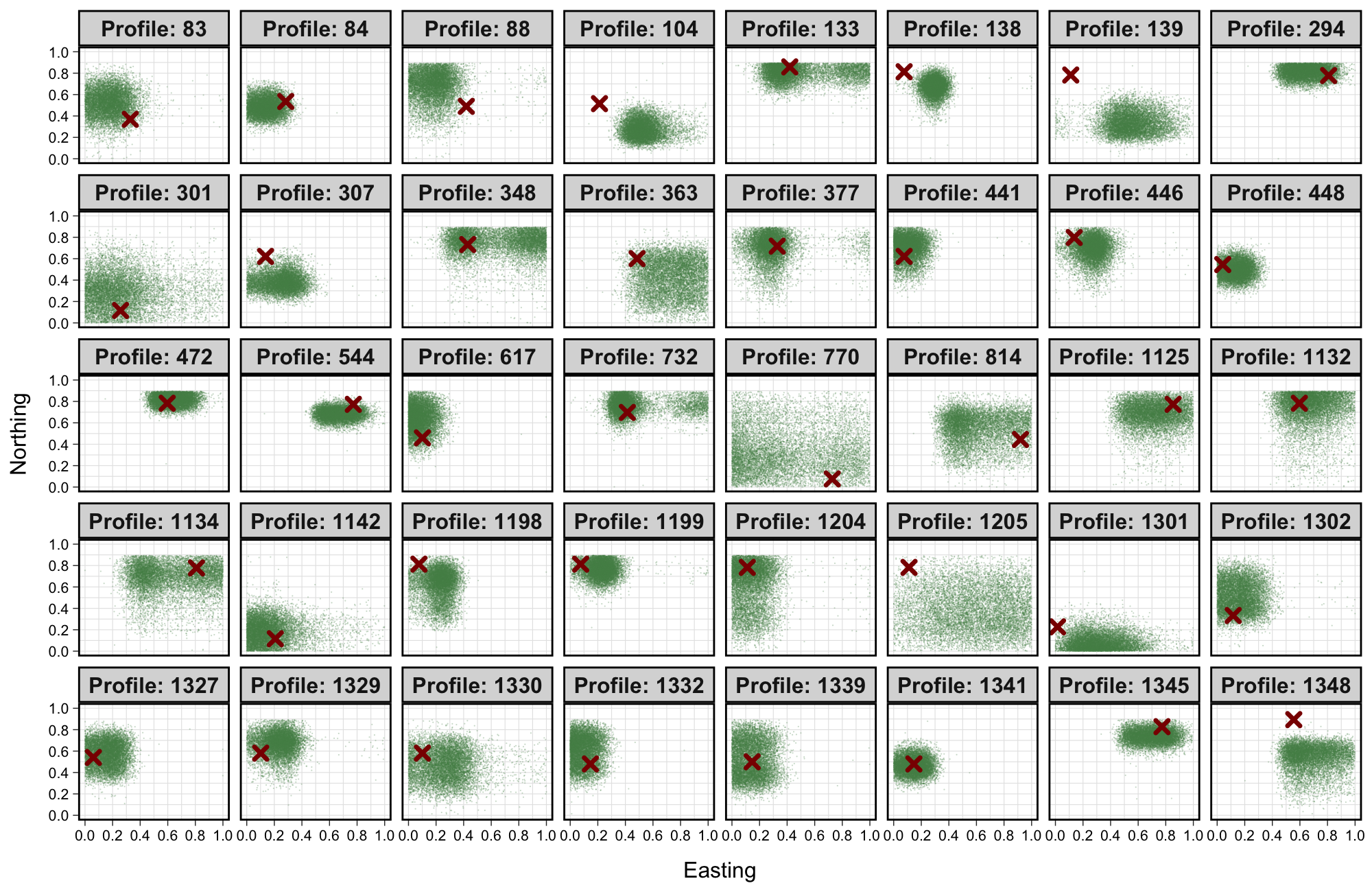}
    \caption{NeVI-Cut}
    \label{fig:LP seed 3 location nevi-cut}
\end{subfigure}
\caption{\textbf{Linguistic Profile Analysis.} Posterior distributions of locations for 40 held-out anchor profiles using SMI-Cut and NeVI-Cut. Red crosses indicate their observed locations, which are considered true for validation analyses.}\label{fig:LP seed 3 location}
\end{figure}

\paragraph{Performance Comparison Per Held-Out Profile.} Figure~\ref{fig:LP seed} presents a comparison of average performance across the held-out profiles across different seeds for NeVI-Cut and SMI-Cut.
Given the visually comparable performance of NeVI-Cut across seeds, for the other comparisons, we include only the results for one of the seeds here for simplicity.
Table~\ref{tab:LP diagnostic table} and Figure~\ref{fig:LP seed 3 diagnostic} show that both on average and for most individual profiles, both methods yield similar results in terms of Wasserstein distance (0.266 for SMI-Cut and 0.280 for NeVI-Cut), RMSE of posterior means (0.203 for both methods), and energy score (0.15 for both SMI-Cut and NeVI-Cut).

Figures~\ref{fig:LP seed 3 location smi-cut}--\ref{fig:LP seed 3 location nevi-cut} also provide similar findings. For most profiles, the posteriors of the profile locations from both methods are close to the observed values (e.g., 472, 1341), whereas for a few others, both of them miss the truth (e.g., 104, 1348). While for most of the profiles, the posteriors concentrate on the same region of the unit rectangle (e.g., profiles
83, 84), there are some profiles for which the posteriors look different (e.g., 139, 770, 814, 1134, 1205). Among the profiles where they differ, SMI-Cut posteriors for some profiles are closer to the observed location (e.g., 139, 1205), and for others, NeVI-Cut's posterior better represents the observed location (e.g., 770, 814, 1134). For a small subset (363, 770, 1205), the posteriors are diffused over a large
fraction of the unit rectangle. This reflects limited information, or potentially misspecified likelihood and/or prior, or potential LPs whose linguistic form
realizations are too sparse or item-poor to constrain location through the latent Gaussian-process basis. Both methods default to
broad uncertainty for these. Thus, on average across the held-out profiles, SMI-Cut and NeVI-Cut perform similarly, and this is reassuring given that they both infer the same target posterior using normalizing flows.

Three classes of diagnostic figures are produced above.  All are generated by the script \texttt{plot\_validation.R} (single-seed view) and \texttt{plot\_seeds.R} (multi-seed view); both scripts are organized so that each figure can be regenerated by stepping line-by-line through a single self-contained section in RStudio.

\subsection{Cause-Specific Mortality Fractions in Mozambique}\label{app:comsa}

In this section, we compare cut methods in a global health application where posteriors are more irregularly shaped, and where upstream data is not available. We focus on predicting cause-specific mortality fractions (CSMFs), the proportion of deaths attributable to each cause, among neonates (0-27 days) and children (1-59 months) in Mozambique using algorithm-predicted cause-of-death (COD) data. The data are collected in the Countrywide Mortality Surveillance for Action program \citep[\href{https://comsamozambique.org/}{COMSA-Mozambique};][]{Macicame2023} and are publicly available in the \href{https://github.com/sandy-pramanik/vacalibration/tree/main/data}{vacalibration} R package. \href{https://comsamozambique.org/}{COMSA-Mozambique} collects \textit{verbal autopsy (VA)} records, a WHO-standardized interview of the caregiver, and they are widely used for population-level mortality surveillance in low- and middle-income countries (LMICs). These VA records are then passed through pre-trained classifiers, known as \textit{computer-coded verbal autopsy (CCVA)} algorithms, to predict individual-level COD. However, as these classifiers are imperfect, the estimation of CSMFs from CCVA-predicted labels must account for algorithmic misclassification using the confusion matrix (misclassification rates). This post-prediction inference problem is known as VA-calibration \citep{datta2020,fiksel2022,fiksel2023correcting,gilbert2023multi}.

For an age group with $C$ causes and a CCVA classifier, let $\Phi=(\phi_{ij})$ denote the $C\times C$ confusion (misclassification) matrix, where
$$
\phi_{ij}=\bbP(\text{CCVA predicts cause } j \given \text{True cause is } i).
$$
Let $\bv=(v_1,\ldots,v_C)^\top$ denote the COMSA data, where $v_j$ deaths are predicted to be from cause $j$ by the classifier. The VA-calibration downstream model is given by
$$
v \con \theta, \Phi \sim \text{Multinomial}(n,q)=\text{Multinomial}(n,\Phi^\top \theta),
$$
where $n=\sum_j v_j$ is the sample size, and it aims to estimate the vector of CSMFs $\theta=(p_1,\ldots,p_C)^\top$, where $p_i=\bbP(\text{True cause is } i)$. The CSMFs $\theta$ are assigned a Dirichlet prior.

Following \cite{pramanik2025modeling}, the upstream analysis uses a secondary labeled dataset ($D_1$) from the \href{https://champshealth.org/}{CHAMPS project}, which includes COD from both VA and a minimally invasive autopsy, and estimates the confusion matrices $\Phi$.
The CHAMPS data from the upstream analysis are not publicly available. \cite{Pramanik2026bmjgh} recently released posterior samples of confusion matrices on \href{https://github.com/sandy-pramanik/CCVA-Misclassification-Matrices}{GitHub} for three major CCVA classifiers (EAVA, InSilicoVA, InterVA), two age groups (neonates and children), and eight countries, including Mozambique. The downstream data ($D_2$) consists of VA-only (unlabeled) death counts $\bv$ which can arise from a different population and is sensitive to likelihood specification. Thus, feedback to the upstream model is undesirable. Hence, we implement cut-Bayes to estimate the CSMF vector $\theta$, while only allowing downstream uncertainty propagation of $\Phi$. Here the upstream parameters are $\eta = \Phi$, while the parameters in the downstream are $\theta=(p_1,\ldots,p_C)^\top$.

\paragraph*{Assumption Validation.}
Before using NeVI-Cut, we show how the Assumption \ref{asm:mono} can be verified for a 2-cause version of the problem. Let $\theta=\text{logit}\, p_1 = \log \frac{p_1}{1-p_1}, y=v_1, \eta=\Phi$, then the analogous downstream setup is 
\[
p_1\sim\mathrm{Beta}(a,b),\qquad 
y\mid p_1,\eta\sim\mathrm{Bin} \bigl(n,\ q(p_1,\eta)\bigr),
\qquad q(p_1,\eta)=\eta_{21}+p_1(\eta_{11}-\eta_{21}).
\]
As
\(
p_1 = \sigma(\theta) := \frac{e^\theta}{1+e^\theta},
\)
after Jacobian adjustment,
\[
p(\theta\mid\eta,D)\,\propto\,
q(\sigma(\theta),\eta)^y\,
\bigl(1-q(\sigma(\theta),\eta)\bigr)^{n-y}\,
\sigma(\theta)^a\,
\bigl(1-\sigma(\theta)\bigr)^b.
\]

Assumption \ref{asm:mono}(i) on spikiness
follows immediately from $p(\calD_2 | \theta,\eta)$ and $p(\theta)$ being continuously differentiable in $(\theta,\eta)$. Assumption \ref{asm:mono}(ii) holds because, as $\eta \in$ some compact set $K_\eta$,
$q(\sigma(\theta),\eta)$ stays uniformly in $[\delta,1-\delta]$ for some $0 < \delta < 1$ over $\theta\in\mathbb R$. This uniformly bounds both the likelihood and the log-score in $\eta$, which is sufficient as discussed in Section \ref{sec:asmver}.
For the quantile derivative bound, when $K_\eta$ is a compact set $q(\sigma(\theta),\eta)$ is uniformly bounded away from $0$ and $1$ and so is  $q(\sigma(\theta),\eta)^y\,
\bigl(1-q(\sigma(\theta),\eta)\bigr)^{n-y}$. Moreover,
\[
\sigma(\theta)^a(1-\sigma(\theta))^b \asymp e^{-b\theta}\quad(\theta\to+\infty),
\qquad
\sigma(\theta)^a(1-\sigma(\theta))^b \asymp e^{a\theta}\quad(\theta\to-\infty).
\]
So the prior dominates and has exponential tails and
the quantile growth assumption is met. The tail thinness lower bound also follows immediately from the exponential tails, which are heavier than Gaussian. Thus, the exact univariate (2-cause) version of the problem satisfies all parts of Assumption \ref{asm:mono}, giving justification for the application of NeVI-Cut for the actual multivariate problem with more causes. 

\paragraph*{Methods Comparison.}
For the analysis, we compare NeVI-Cut (using a fully autoregressive RQ-NSF) with the full Bayes VA-calibration posterior and with the uncalibrated point estimate $\widehat \bq$, obtained by setting $\Phi$ to be the identity matrix in the downstream model. For full Bayes, as the upstream data and model are not available, we use the sequential sampling approach of \citet{pramanik2025modeling} that well approximates the full Bayes posterior. This method uses the posterior samples of the misclassification matrices $\eta$ to create an approximate Dirichlet prior for $\eta$ in the downstream analysis. Among the cut-Bayes methods, we use nested MCMC as the primary benchmark for accuracy and runtime. To evaluate whether the nonparametric neural variational family in NeVI-Cut is necessary for this application, we also consider parametric NeVI-Cut variants. Specifically, we consider two parametric variational families: a Dirichlet variational family that imposes a linear structure of $\Phi$ on the concentration parameters, and a logistic-normal variational family whose mean depends on $\Phi$.

We refer to the first method as Parametric Cut (Dirichlet), with variational family
$$
q_{\alpha}(\theta \mid \Phi)
=
\operatorname{Dirichlet}(\theta \mid \Phi^\top \alpha).
$$

We refer to the second method as Parametric Cut (Logistic-normal). For this method, we define the variational family of $\theta$ by 
$$
z \mid \Phi \sim N\bigg(\mu(\Phi;\beta), \Sigma\bigg),
\qquad
\theta = \operatorname{softmax}(z),
$$
where $z=(z_1,\ldots,z_{C-1})^\top$ uses category $C$ as the reference category, and
$$
\mu_j(\Phi;\beta)
=
\log \frac{(\Phi^\top \beta)_j}{(\Phi^\top \beta)_C},
\qquad j=1,\ldots,C-1.
$$
The variational parameters are $\alpha$ for Parametric Cut (Dirichlet), and $\beta$ and $\Sigma$ for Parametric Cut (Logistic-normal).

In our implementation of the methods, for NeVI-Cut, following notations in Section~\ref{sec:norm flow conditional var family}, we consider the class of conditional distributions $q_\vartheta (\theta \given \Phi)$ corresponding to the law
\begin{equation*}
\theta = T_\vartheta(\text{vec}(\Phi_{-C}),Z), \quad Z \sim N(0,I_C).
\end{equation*}

Here $T_\vartheta$ is a composition of neural network-based transformations with parameters $\vartheta$, and is conditioned on the vectorized misclassification matrix $\text{vec}(\Phi_{-C})$. The transformation maps samples from a base Gaussian distribution to the simplex through a sequence of conditional normalizing flows, followed by a final stick-breaking transformation that ensures $\theta$ lies on the probability simplex. When vectorizing $\Phi$, the last column is removed (emphasized by the notation $\Phi_{-C}$) since each row lies on the simplex and one column is therefore redundant. We utilize 1000 upstream samples of $\Phi$. Other model configuration for NeVI-Cut and sensitivity analyses are provided in Section~\ref{app:sensitivity}. For the full Bayes and nested MCMC, a 10,000 burn-in is applied. Across all methods, $10^6$ samples are generated for inference, with nested MCMC producing 1000 samples of $\theta$ for each of the $1000$ samples of $\Phi$.
Following \cite{pramanik2025modeling}, we use the Dirichlet$(1+4C\widehat \bq)$ prior on $\theta$, where $\widehat \bq = v/n$ is the uncalibrated point estimate.

\begin{figure}[!t]
    \centering
    \includegraphics[width=\textwidth]{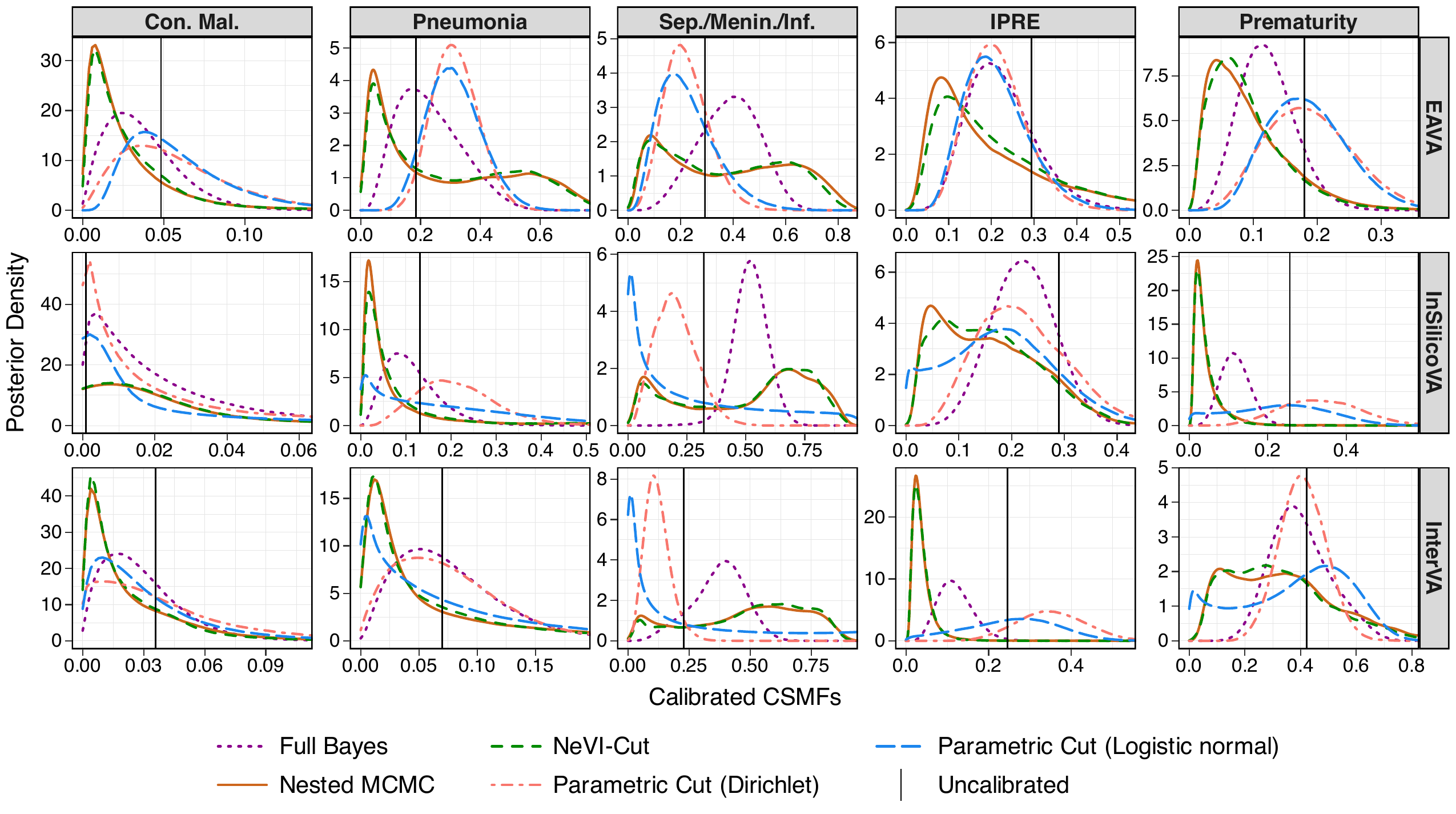}
    \caption{\textbf{Child Mortality Surveillance in Mozambique.} Posterior distributions of calibrated cause-specific mortality fractions (CSMFs) in neonates (0-27 days) based on VA-only data collected by the Countrywide Mortality Surveillance for Action (COMSA) program in Mozambique.
    Con. mal., Sep./Menin./Inf., IPRE denote congenital malformation, sepsis/meningitis/infection, and intrapartum-related events.}\label{fig:neonate_results:supp}
\end{figure}

\begin{figure}[htbp]
    \centering
    \includegraphics[width=.9\textwidth]{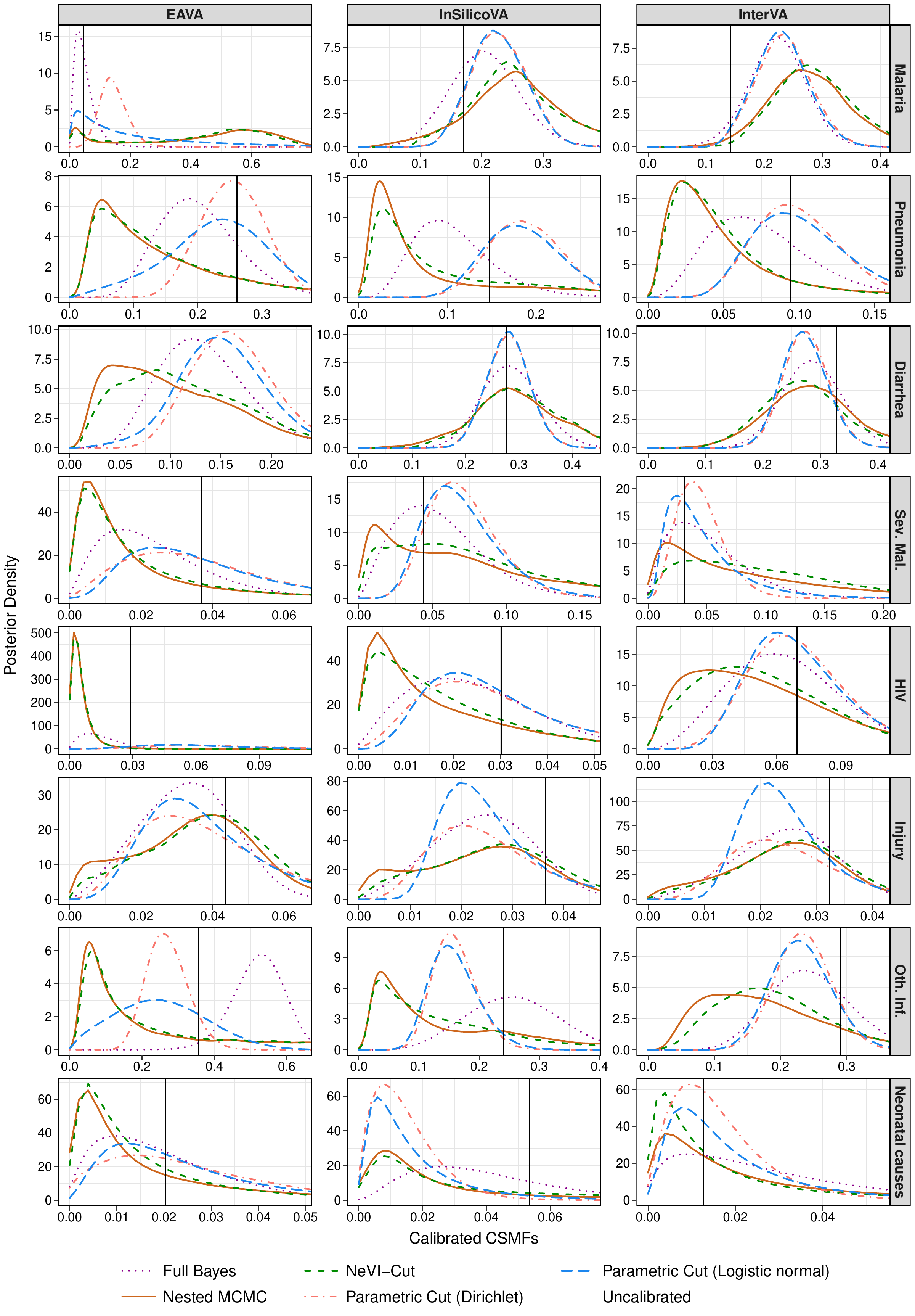}
    \caption{\textbf{Child Mortality Surveillance in Mozambique.} Posterior distributions of calibrated cause-specific mortality fractions (CSMFs) for children aged 1-59 months, based on VA-only data collected by the Countrywide Mortality Surveillance for Action (COMSA) program in Mozambique. NeVI-Cut effectively captures complex distributional characteristics, such as multimodality and skewness, achieving performance comparable to nested MCMC. Sev. mal. and Oth. Inf. denote severe malnutrition and other infections. The horizontal axes have been modified accordingly to improve readability.
    }\label{fig:child_results}
\end{figure}

Figure \ref{fig:neonate_results:supp} presents posterior densities of neonatal CSMFs from different methods. The same for children is reported in Figure \ref{fig:child_results}. We highlight several notable findings. First, the full Bayes posterior differs substantially from the cut-posteriors, suggesting significant sensitivity to feedback. \cite{frazier2026posterior} argued from a risk-assessment point of view that cut-Bayes should be given more weight over full Bayes when the estimates from the two differ considerably. Hence, cutting feedback is preferred here due to potential downstream model misspecification. 

Second, we also observe that the calibrated cut-posteriors display complex structure, including multimodality and skewness. Third, uncalibrated estimates differ noticeably from the calibrated results, highlighting the impact of algorithmic misclassification and the importance of calibration. Fourth, the calibrated estimates from NeVI-Cut closely agree with the gold-standard nested MCMC across causes, CCVA algorithms, and age groups. We quantify this agreement in Table \ref{tab:wasserstein_time} using empirical Wasserstein distances via the \href{https://cran.r-project.org/package=transport}{\texttt{transport}} R package.
Across settings, NeVI-Cut achieves substantially smaller Wasserstein distances than both of the parametric cut methods which perform poorly and depart substantially from nested MCMC and NeVI-Cut, indicating limited expressiveness of the parametric variational family. 

Finally, NeVI-Cut is computationally more efficient than the nested MCMC.
{Note that these runtimes in the tables reflect only the finalized implementations of both methods. In practice, model development requires many trial runs (for example, exploring different hyperparameter settings), and doing this with nested MCMC runs is prohibitively time-consuming. In contrast, simpler NeVI-Cut architectures can run much faster while still providing reasonable approximations to the nested MCMC, enabling rapid exploration before using a more expressive NeVI-Cut for the final fit and yielding substantial additional time savings overall.
}

\begin{table}[!t]
\centering
\caption{\textbf{Child Mortality Surveillance in Mozambique.} Wasserstein distances to Nested MCMC as reference and computation time.}
\label{tab:wasserstein_time}
\resizebox{\textwidth}{!}{
\begin{tabular}{lllrr}
\noalign{\hrule height 1.5pt}
\textbf{Population} & \textbf{Algorithm} & \textbf{Method} & \textbf{WD} & \textbf{Time (Seconds)} \\
\noalign{\hrule height 1.5pt}
& & Nested MCMC & NA & 324.02 \\
& & NeVI-Cut & 0.075 & 241.15 \\
Neonate & EAVA & Full Bayes & 0.232 & 502.59 \\
& & Parametric Cut (Dirichlet) & 0.335 & 5.26 \\
& & Parametric Cut (Logistic normal) & 0.306 & 4.40 \\
\midrule
& & Nested MCMC & NA & 343.70 \\
& & NeVI-Cut & 0.051 & 242.39 \\
Neonate & InSilicoVA & Full Bayes & 0.461 & 480.52 \\
& & Parametric Cut (Dirichlet) & 0.575 & 5.65 \\
& & Parametric Cut (Logistic normal) & 0.425 & 4.23 \\
\midrule
& & Nested MCMC & NA & 314.38 \\
& & NeVI-Cut & 0.075 & 227.99 \\
Neonate & InterVA & Full Bayes & 0.283 & 537.96 \\
& & Parametric Cut (Dirichlet) & 0.575 & 4.99 \\
& & Parametric Cut (Logistic normal) & 0.428 & 3.83 \\
\midrule
& & Nested MCMC & NA & 578.12 \\
& & NeVI-Cut & 0.082 & 372.71 \\
Child & EAVA & Full Bayes & 0.609 & 1111.09 \\
& & Parametric Cut (Dirichlet) & 0.445 & 6.74 \\
& & Parametric Cut (Logistic normal) & 0.335 & 5.36 \\
\midrule
& & Nested MCMC & NA & 634.83 \\
& & NeVI-Cut & 0.072 & 352.93 \\
Child & InSilicoVA & Full Bayes & 0.205 & 1826.48 \\
& & Parametric Cut (Dirichlet) & 0.221 & 5.69 \\
& & Parametric Cut (Logistic normal) & 0.207 & 5.00 \\
\midrule
& & Nested MCMC & NA & 676.42 \\
& & NeVI-Cut & 0.076 & 355.48 \\
Child & InterVA & Full Bayes & 0.146 & 1132.08 \\
& & Parametric Cut (Dirichlet) & 0.171 & 8.61 \\
& & Parametric Cut (Logistic normal) & 0.153 & 3.67 \\
\noalign{\hrule height 1.5pt}
\end{tabular}
}
\vspace{0.5em}
\begin{minipage}{0.99\textwidth}
\footnotesize
\textit{Note:} WD represents the 2-Wasserstein distance computed via the \href{https://cran.r-project.org/package=transport}{\texttt{transport}} R package. Time is reported in seconds. Nested MCMC is used as the reference distribution.
\end{minipage}
\end{table}

\begin{figure}[!t]
    \centering

    \begin{minipage}[c]{0.49\textwidth}
        \centering
        \includegraphics[width=0.48\textwidth]{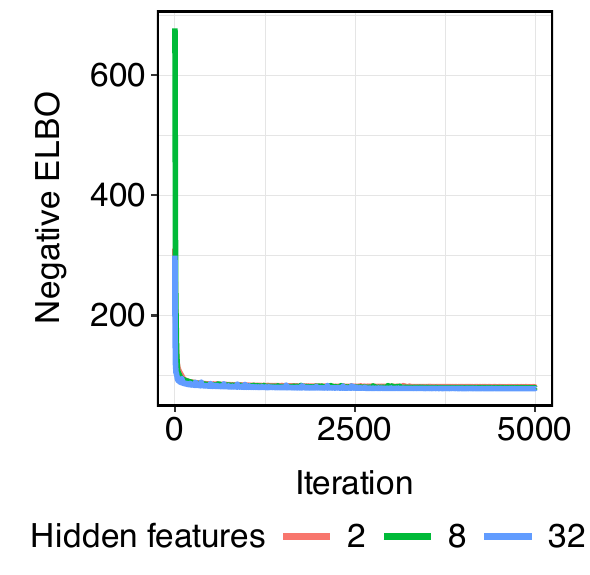}\hfill
        \includegraphics[width=0.48\textwidth]{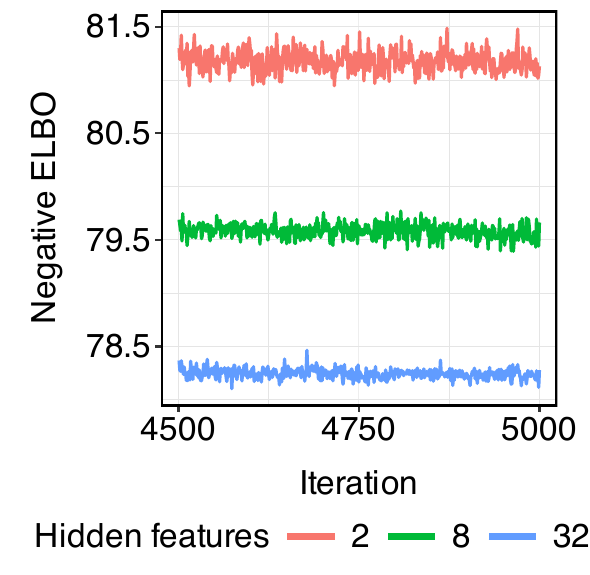}

        \vspace{0.6em}

        \includegraphics[width=0.48\textwidth]{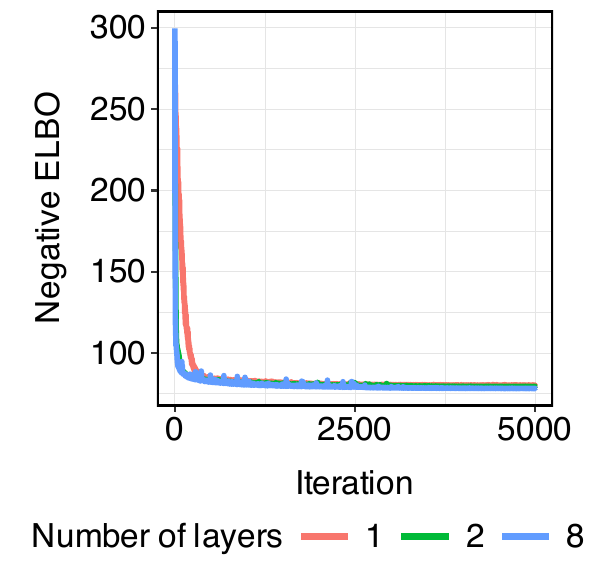}\hfill
        \includegraphics[width=0.48\textwidth]{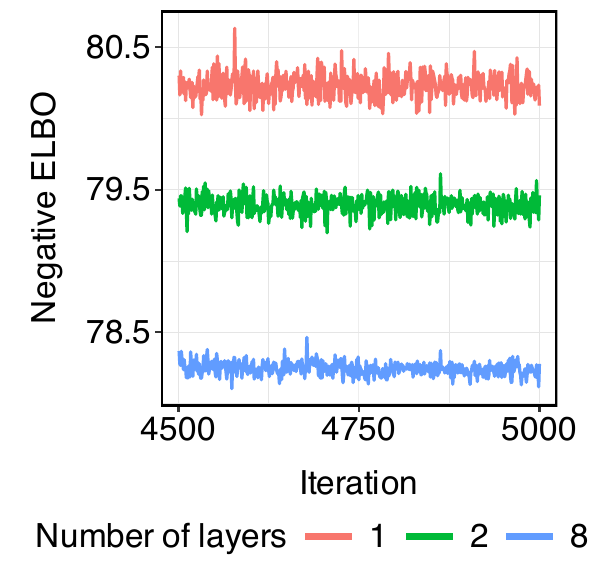}
    \end{minipage}\hfill
    \begin{minipage}[c]{0.49\textwidth}
        \centering
        \includegraphics[width=0.92\textwidth]{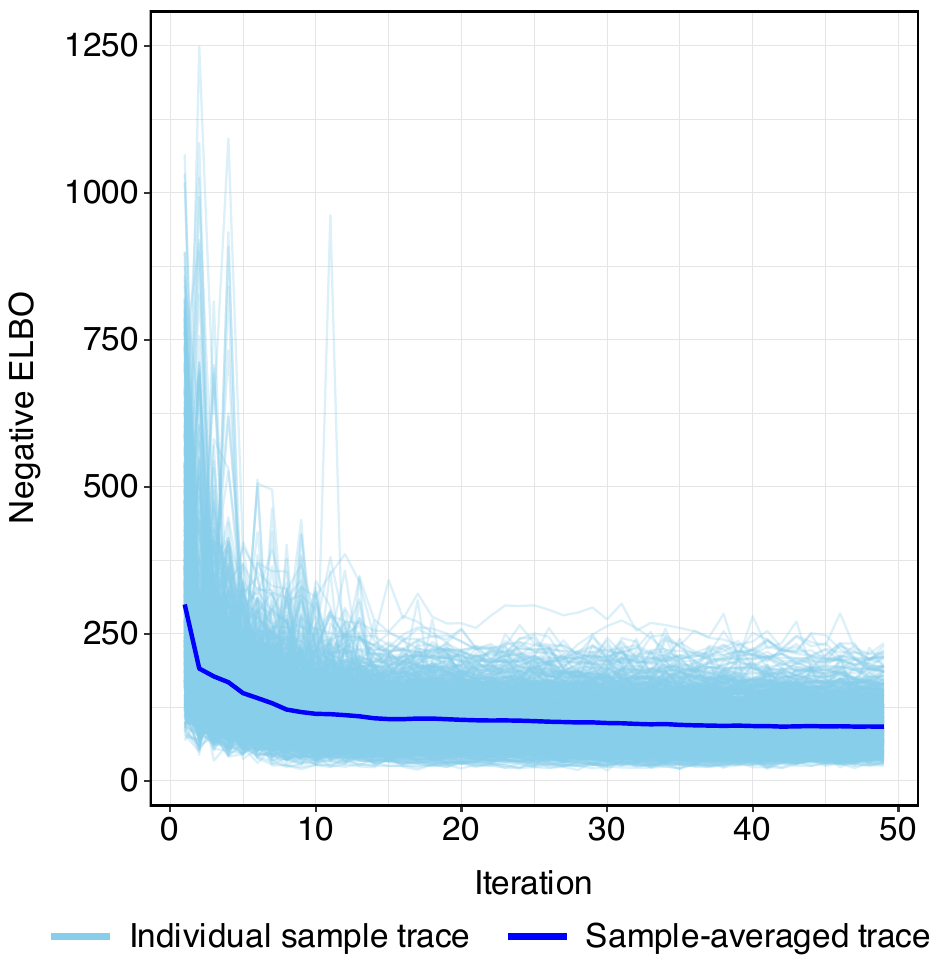}
    \end{minipage}

    \caption{\textbf{Child Mortality Surveillance in Mozambique.} Left panel ($2 \times 2$): negative Evidence Lower Bound (ELBO) for NeVI-Cut as normalizing flow complexity increases. Top row varies the number of hidden features in each layer, shown over the full run (top-left sub-panel) and restricted to the final $500$ iterations (top-right sub-panel). Bottom row varies the number of flow layers, shown over the full run (bottom-left sub-panel) and restricted to the final 500 iterations (bottom-right sub-panel). Right panel: negative ELBO over the first 50 iterations across upstream samples, with individual sample traces in light blue and the sample-averaged trajectory overlaid in dark blue.}
    \label{fig:nelbo_panels:supp}
\end{figure}

\paragraph*{Uniform Convergence for NeVI-Cut.}
Finally, we assess the impact of the choice of neural architecture on NeVI-Cut. The size of a normalizing flow is controlled by the number of transformation layers and the number of hidden features in the neural networks used as conditioners. Figure~\ref{fig:nelbo_panels:supp} presents trace plots of the empirical negative ELBO \eqref{eq:ELBO} for the analysis across different configurations of these parameters. From the left-column plots (entire trace), we see that the ELBO converges across different architectures. However, from the middle-column plots (zoomed in on the final $500$ iterations), we see that the convergence values depend on the complexity of the normalizing flow, with lower values for flows with a higher number of hidden features (top) or a higher number of layers (bottom). This indicates better approximation with more complex architectures, aligning with the theory. The right panel of Figure \ref{fig:nelbo_panels:supp} plots the negative ELBO component coming (as a function of iterations) from each of the upstream parameter ($\eta$) samples. Each light blue line corresponds to one upstream posterior sample of $\eta$, whereas the dark blue line corresponds to the average negative ELBO (over all the upstream samples), which is the loss minimized in NeVI-Cut. These trace plots indicate that the KL divergence is well controlled across all samples, aligning with our uniform convergence result.

\subsection{Association of Air Pollution Risk with Cardiovascular Disease}\label{app:gbd}
\begin{figure}[!h]
\centering
\includegraphics[width=0.5\textwidth]{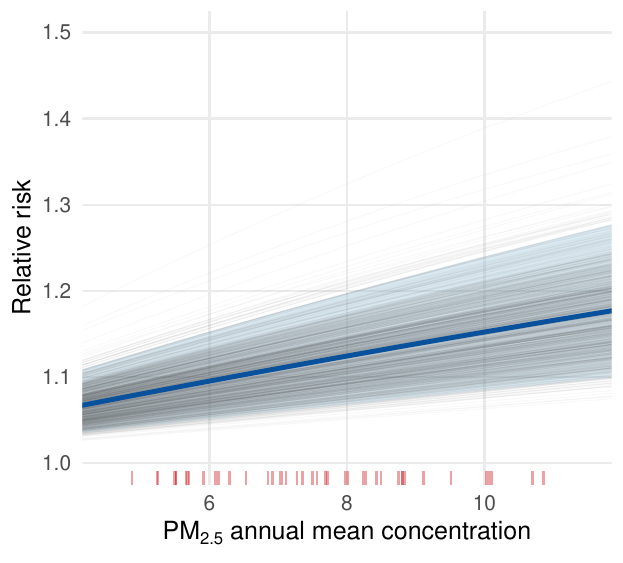}
\caption{\textbf{Global Burden of Disease in California.} Posterior draws of the PM$_{2.5}$ relative-risk curves for ischemic heart disease, ages 65--69. Each gray line represents one posterior draw from the GBD risk curves, the dark blue line is the posterior mean curve, the shaded blue band shows the pointwise 95\% uncertainty interval across draws, and the red rug marks indicate the observed county-level PM$_{2.5}$ values in California.}
\label{fig:gbd_draw}
\end{figure}

\paragraph*{Upstream Results.}
We first describe the Global Burden of Disease (GBD) data on PM$_{2.5}$ risk used in the analysis. The Global Burden of Disease Study 2019 (GBD 2019) estimated the burden of diseases, injuries, and risk factors for 204 countries and selected subnational locations. The study provides uncertainty-quantified long-term PM$_{2.5}$ exposure risk estimates for various conditions, including ischemic heart disease (IHD). These are available as posterior draws of particulate matter risk curves downloadable through the Global Health Data Exchange \citep{gbd2019_pm_risk_curves} at \href{https://ghdx.healthdata.org/record/ihme-data/global-burden-disease-study-2019-gbd-2019-particulate-matter-risk-curves}{GBD 2019}. In this analysis, we use the publicly available GBD 2019 particulate-matter risk-curve draws for IHD among adults aged 65--69 years (best matching the age group of the IHD data described subsequently). As shown in Figure \ref{fig:gbd_draw}, each gray curve represents a single posterior draw of the relative risk across PM$_{2.5}$ concentrations, and the width of the blue band reflects the uncertainty across all posterior draws.

The downstream analysis investigates county-level IHD mortality in California using a Poisson regression model, where the outcome is the county-level number of IHD deaths and the model adjusts for county-level confounders and spatial random effects. This setting naturally motivates a cut-Bayes model. The GBD exposure-response risk-curve draws are publicly available and obtained from a reliable external source, while the downstream analysis assumes a Poisson regression model adjusted for several confounders and may therefore be subject to model misspecification. A full Bayes analysis could allow this downstream misspecification to feed back into the upstream risk-curve model, contaminating the externally derived risk-curve draws. Moreover, the GBD risk curves are estimated at the country and selected subnational levels, whereas the downstream mortality analysis is conducted at the county level. The cut-Bayes model avoids such feedback from the downstream county-level mortality model to the upstream GBD risk-curve model, while still propagating risk-curve uncertainty into downstream inference.

\paragraph*{Downstream Data availability and process.}
County-level mortality data are obtained from the Underlying Cause of Death database for 1999--2020 available on \href{https://wonder.cdc.gov/ucd-icd10.html}{CDC WONDER} \citep{cdc_nchs_wonder_mortality_2026}. This database provides mortality and population counts for all U.S. counties. We extract 2019 death counts for ischemic heart disease, defined using ICD-10 codes I20--I25, among individuals aged 65--70 years as the downstream data. County-level demographic and socioeconomic covariates are obtained from the U.S. Census Bureau's 2015--2019 American Community Survey (\href{https://www.census.gov/acs/www/data/data-tables-and-tools/data-profiles/2019/}{ACS}) 5-year estimates. These covariates included percent male, log median household income, percent with a bachelor's degree or higher, and percent Hispanic. The county-level files are merged by five-digit FIPS codes after harmonizing variable names and removing unmatched or incomplete records.

The key covariate of interest is the relative risk associated with PM$_{2.5}$ concentration. As described in the previous section, the upstream results provide posterior draws of the relative risk across the full range of PM$_{2.5}$ concentrations. To link these upstream results to the downstream analysis, we obtain county-specific PM$_{2.5}$ concentration levels and extract the corresponding cross-sectional posterior draws of the relative risk. The county-level annual PM$_{2.5}$ concentrations for 2019 are obtained by averaging daily county-level PM$_{2.5}$ concentrations. The daily concentration data are obtained from the \href{https://www.datalumos.org/datalumos/project/243609/version/V1/view?path=/datalumos/243609/fcr:versions/V1/Daily_County-Level_PM2.5_Concentrations-_2001-2019_20260112.csv&type=file}{Daily County-Level PM$_{2.5}$ Concentrations} data released by the United States Department of Health and Human Services, Centers for Disease Control and Prevention, and Environmental Health and Toxicology \citep{hhs_cdc_eht_2026_pm25}. The red rug marks in Figure \ref{fig:gbd_draw} indicate the annual county-level PM$_{2.5}$ concentrations in California and are used to extract the corresponding posterior draws of the relative risk for county $j$ from the GBD IHD risk curves.

In this spatial setting, spatial random effects are defined on a 10-km grid over California and linked to county-level outcomes through population-weighted county--tile crosswalks. To obtain non-overlapping county supports, each 10-km grid cell is assigned to a single county. Grid cells intersecting multiple counties are assigned to the county with the largest population contribution. We use \href{https://data.worldpop.org/GIS/Population/Global_2000_2020_1km/2019/USA/}{2019 WorldPop} gridded population counts to distribute county-level age-specific populations over the 10-km spatial grid \citep{worldpop2018population}. For county $j$, let $A_j$ denote the set of 10km grid cells contributing to county $j$, $P_j$ the observed age-specific county population and $p_{ij}$ the WorldPop population in grid cell $i\in A_j$. We define
\[
a_{ij} = P_j \frac{p_{ij}}{\sum_{i \in A_j} p_{ij}},
\]
where $a_{ij}$ denotes the estimated age-specific population at risk in grid cell $i$ for county $j$. By construction $\sum_{i \in A_j} a_{ij} = P_j$. This allows the spatial random effects to be represented at a finer spatial resolution, helping capture intra-county variability, while preserving the observed county-level population totals.

\paragraph*{Downstream model.}
Given upstream posterior draw from GBD study for county $j=1,\dots,J$, we model the county-level ischemic heart disease death count $y_j$ as
\begin{gather*}
y_j \mid \theta, z_j \ind \operatorname{Poisson}\left(\mu_j (\theta, z_j)\right) \quad \forall j, \quad \text{with}\\
\theta = \left( \alpha, \beta_{\mathrm{RR}}, \beta, w = \left\{ w_i \right\}, \sigma_w \right), \quad \mu_j \equiv \mu_j (\theta, z_j) =
\sum_{i \in A_j}
a_{ij}
\exp\left\{
\alpha + z_j \beta_{\mathrm{RR}} + x_j^\top \beta + w_i
\right\} .
\end{gather*}
Here,
$z_{j}=\log(\text{RR}_{j})$ is the draw-specific log relative risk (RR) induced by PM$_{2.5}$ exposure, $x_j$ is the vector of county-level covariates, and $w_i$ is the spatial random effect for grid cell $i$.

We independently assign the priors
\begin{equation*}
\alpha \sim \mathcal{N}(0,\,5^2), \quad
\beta_{\mathrm{RR}} \sim \mathcal{N}(0,\,5^2), \quad \beta \sim \mathcal{N}(0,\,5^2 I_p) .
\end{equation*}

For the spatial random effects $w=(w_1,\dots,w_I)^\top$, we assign an intrinsic conditional autoregressive (ICAR) prior \citep{besag1974spatial,besag1991bayesian}, with weak $L_2$ regularization. Specifically,
\begin{align*}
p(w \mid \sigma_w)
&\propto
\exp\left\{
-\frac{1}{2 \sigma_w^2} \bigg[
\sum_{i \sim {i'}}
(w_i - w_{i'})^2 + \lambda \sum_{i=1}^I w_i^2\bigg]
\right\}, \quad \sum_{i=1}^I w_i = 0,
\end{align*}
where $i\sim {i'}$ indicates that regions $i$ and ${i'}$ are neighbors. We set $\lambda = \frac{1}{5^2}$, with $L_2$ regularization stabilizing the ICAR precision matrix and accommodating disconnected components of the adjacency graph.
For computation, we use the reparameterization
$$w_i=\sigma_w(w^*_i-\bar{w}^*), \quad \bar{w}^* = \frac{1}{I}\sum_{i=1}^I w^*_i,$$
where $w^*$ follows the regularized ICAR prior with unit spatial scale, \begin{align*}
p(w^*)
&\propto
\exp\left\{
-\frac{1}{2} \bigg[
\sum_{i \sim {i'}}
(w_i^* - w_{i'}^*)^2 + \lambda \sum_{i=1}^I (w_i^*)^2\bigg]
\right\}, \quad \sum_{i=1}^I w_i^* = 0.
\end{align*}
Finally, we assign $\sigma_w \sim \text{Half-normal (0,1)}$.

\begin{figure}[!h]
\centering
\begin{subfigure}[b]{0.49\textwidth}
    \centering
    \includegraphics[width=\textwidth]{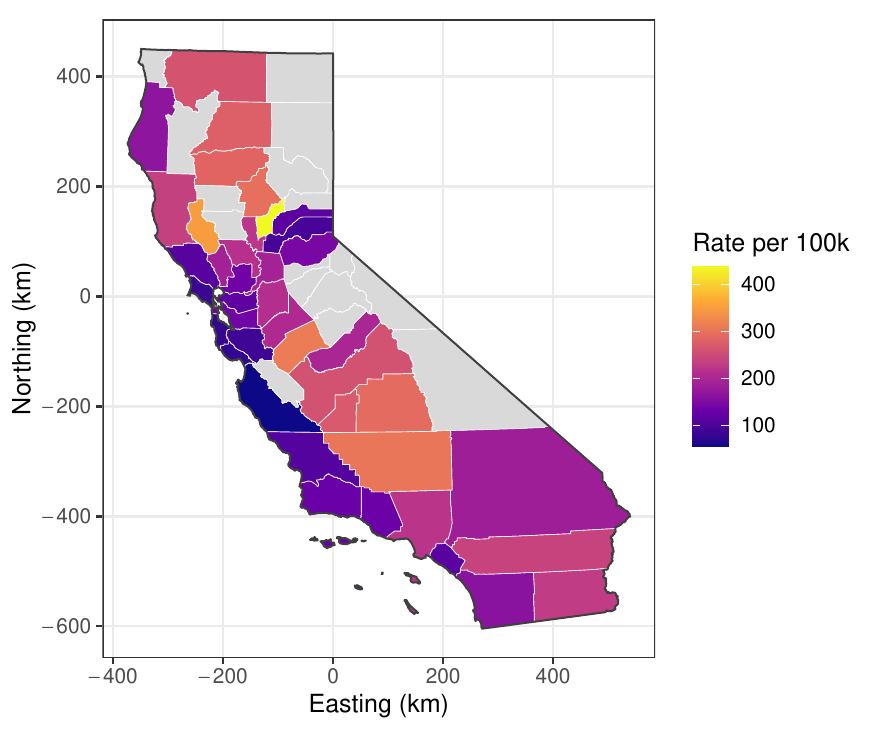}
    \caption{Observed IHD Death Rate by County}
    \label{fig:gbd_crude_rate}
\end{subfigure}
\hfill
\begin{subfigure}[b]{0.5\textwidth}
    \centering
    \includegraphics[width=\textwidth]{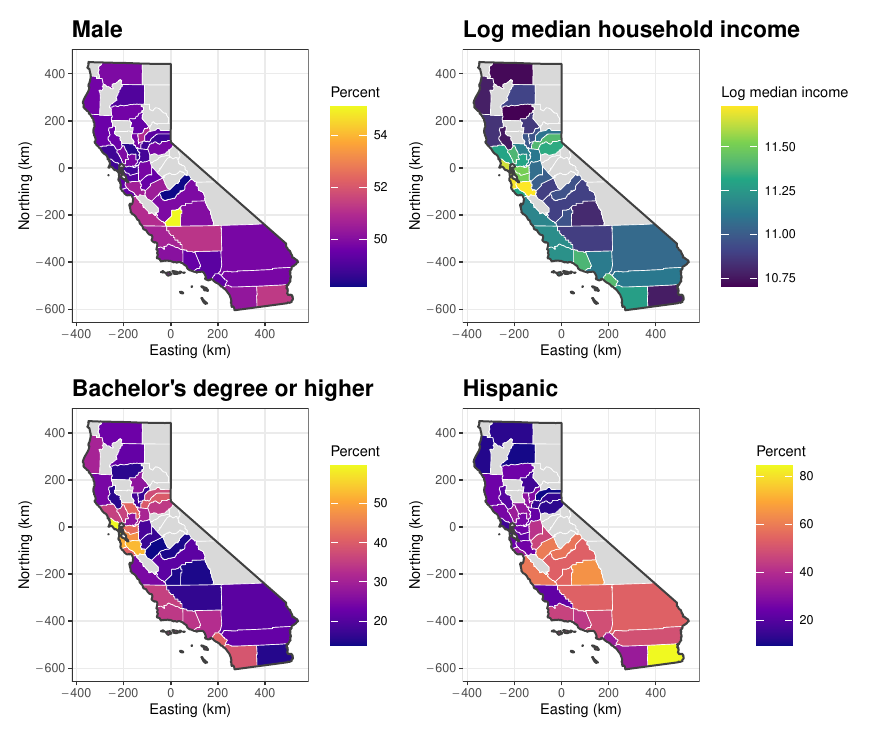}
    \caption{County-Level Confounders}
    \label{fig:gbd_covariate}
\end{subfigure}
\caption{\textbf{Global Burden of Disease in California.} Left panel shows observed county-level ischemic heart disease death rates for the California age 65--69 population. Right panel presents county-level maps of the four confounders included in the model. Grey color indicates data is not available.}
\label{fig:gbd_data}
\end{figure}

Figure~\ref{fig:gbd_data} presents the county-level crude death rates and covariates included in the downstream model. The final dataset consists of $J=42$ California counties with complete data, including 3298 deaths in total, and $I=2730$ grid-level spatial random effects.

For downstream posterior inference, we compare nested MCMC, NeVI-Cut, and Parametric Cut (a parametric variant of NeVI-Cut). NeVI-Cut uses a fully autoregressive RQ-NSF. Other model configurations for NeVI-Cut and sensitivity analyses are provided in Section~\ref{app:sensitivity}.  For Parametric Cut, we use a block mean-field variational family conditional on the upstream samples. Specifically, we use a full-rank Gaussian block for the regression parameters and spatial scale parameter, and a mean-field Gaussian block for the spatial latent effects. Let
$\theta = (\alpha, \beta, \log \sigma_w, w).$ Then the variational family for parametric cut is defined as
$$
q(\theta \mid \eta)
=q(\alpha, \beta, \log \sigma_w \mid \eta)\,
q(w \mid \eta).
$$
We use
$q(\alpha, \beta, \log \sigma_w \mid \eta)=\mathcal{N}\!\bigl(\mu_g(\eta), \Sigma_g\bigr)$, where $\mu_g(\eta)$ is a linear function of $\eta$ and $\Sigma_g$ is a full covariance matrix. For the spatial random effects, we use 
$q_w(w\mid \eta)=\prod_{i=1}^I \mathcal{N}\!\bigl(\mu_{w,i}(\eta), s_{w,i}^2(\eta)\bigr)$, where $\mu_{w,i}(\eta)$ and $\log s_{w,i}(\eta)$ are each modeled as linear functions of $\eta$ with intercept.

\begin{figure}[!t]
\centering
\begin{subfigure}[b]{0.45\textwidth}
    \centering
    \includegraphics[width=\textwidth]{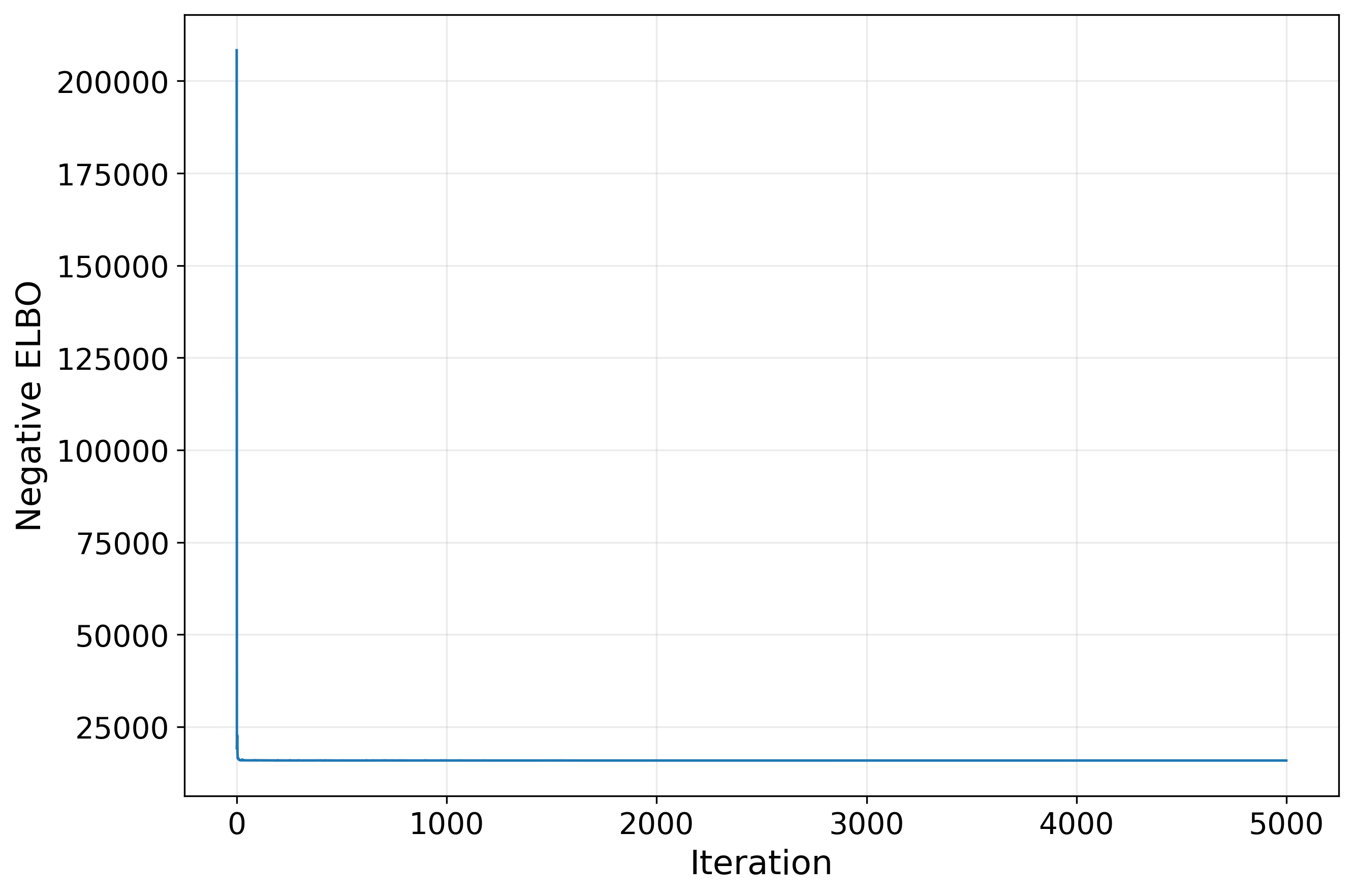}
    \caption{Negative ELBO convergence in NeVI-Cut}
\end{subfigure}
\hfill
\begin{subfigure}[b]{0.45\textwidth}
    \centering
    \includegraphics[width=\textwidth]{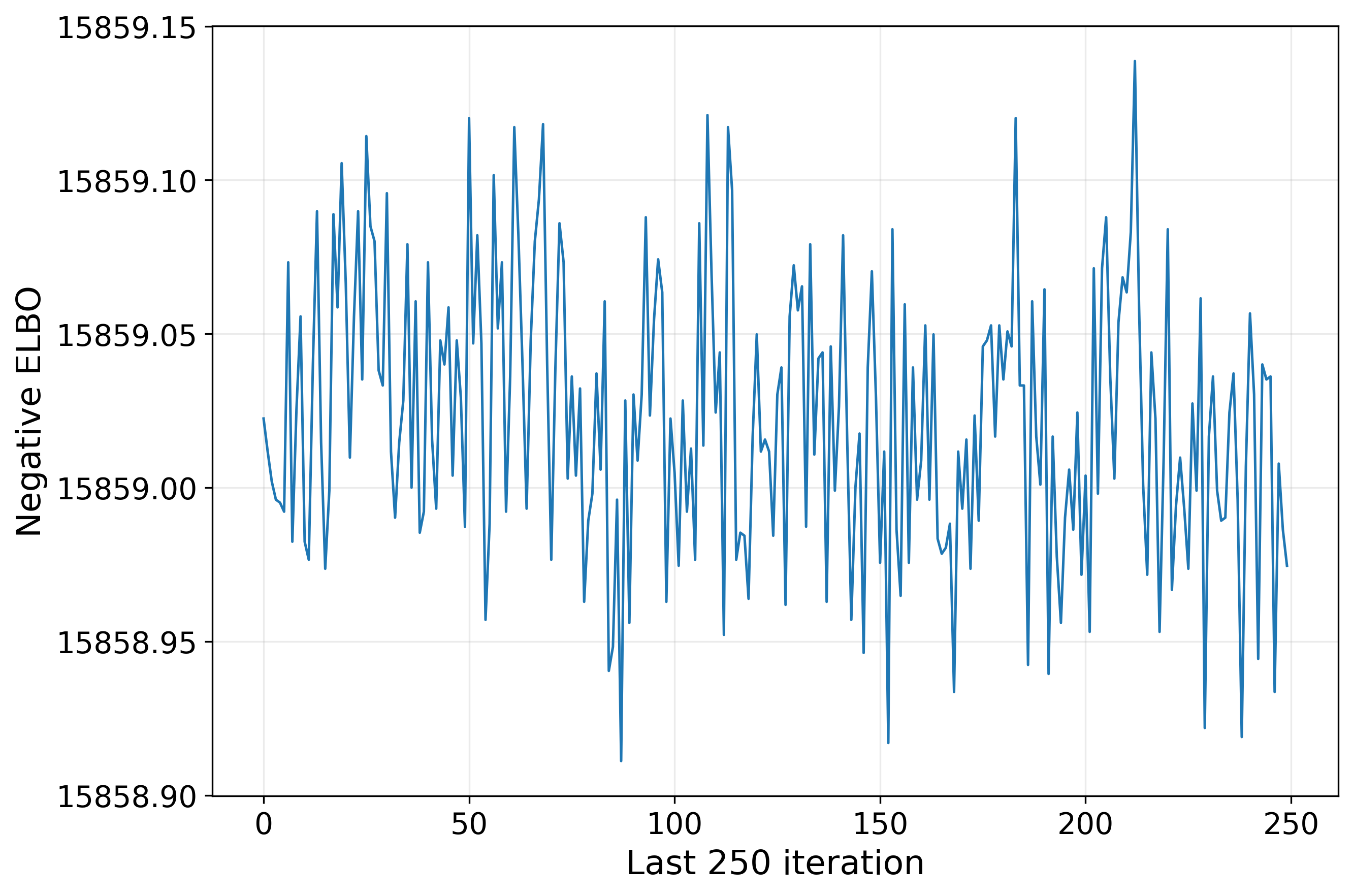}
    \caption{Traceplot of the last 250 iterations}
\end{subfigure}
\caption{\textbf{Global Burden of Disease in California.} Left panel shows the negative ELBO over all 5000 optimization iterations for GBD data analysis, while the right panel zooms in on the final 250 iterations.}
\label{fig:gbd_convergence}
\end{figure}

\begin{table}[!b]
\centering
\caption{\textbf{Global Burden of Disease in California.} Computation time (in hours) and marginal Wasserstein distance for $\beta_{\log RR}$, using Nested MCMC as the reference.}
\label{tab:beta_wasserstein_time}
\begin{tabular}{lccc}
\noalign{\hrule height 1.5pt}
\textbf{Method} & \textbf{Training time} & \textbf{Sampling time} & \textbf{Wasserstein distance} \\
\noalign{\hrule height 1.5pt}
Nested MCMC           & \multicolumn{2}{c}{330.20} & -- \\
NeVI-Cut       & 27.24 & 7.77  & 0.0048 \\
Parametric Cut & 1.80  & 0.016 & 0.0246 \\
\noalign{\hrule height 1.5pt}
\end{tabular}
\end{table}

For the nested MCMC benchmark, the downstream model is fitted for each upstream posterior draw using \textit{rstan}, with 8000 burn-in iterations followed by 2000 retained posterior draws for inference. We run 5000 iterations for NeVI-Cut. The convergence of NeVI-Cut is shown in Figure~\ref{fig:gbd_convergence}.

\begin{figure}[ht]
\centering
\includegraphics[width=0.95\textwidth]{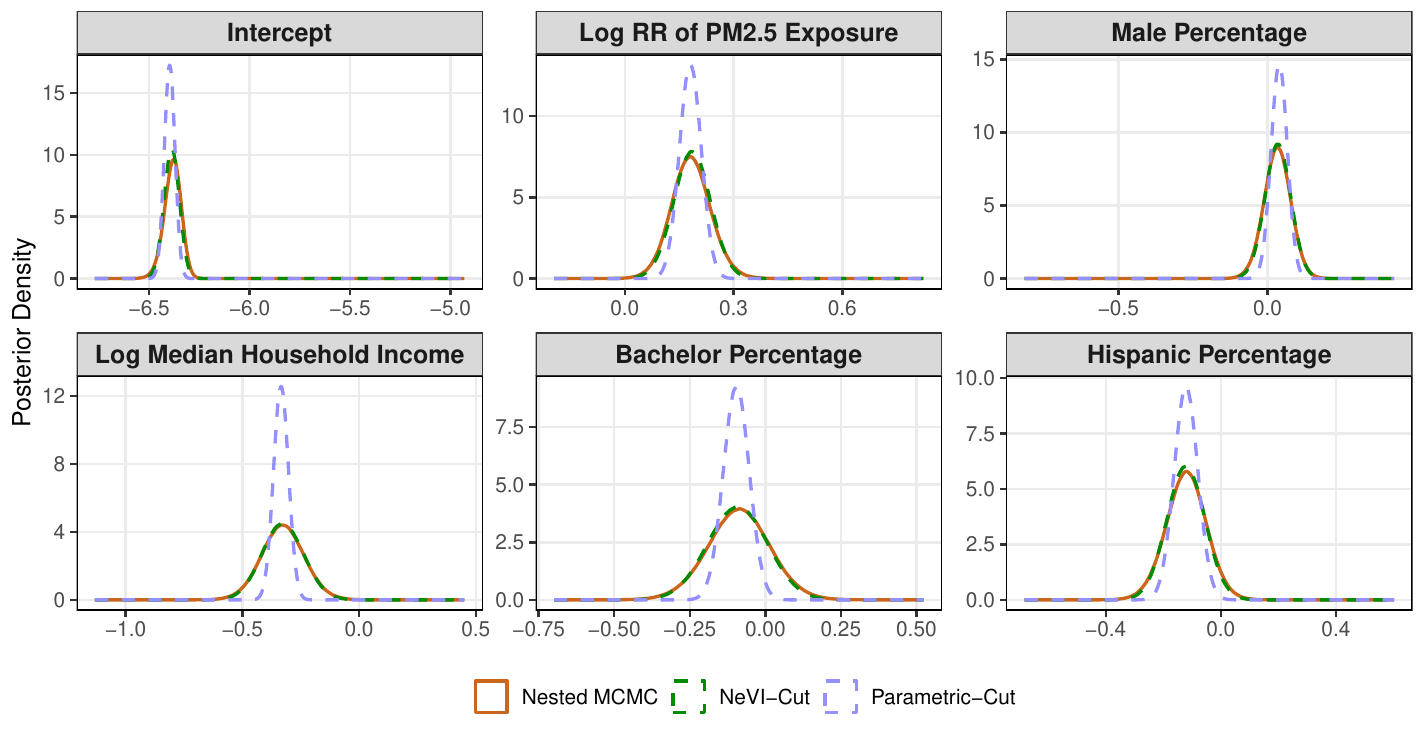}
\caption{\textbf{Global Burden of Disease in California.} Posterior distributions of the regression coefficients estimated by nested MCMC, NeVI-Cut, and parametric NeVI-Cut.}
\label{fig:gbd_coeff}
\end{figure}

\begin{figure}[!t]
\centering
\begin{subfigure}[b]{0.75\textwidth}
    \centering
    \includegraphics[width=\textwidth]{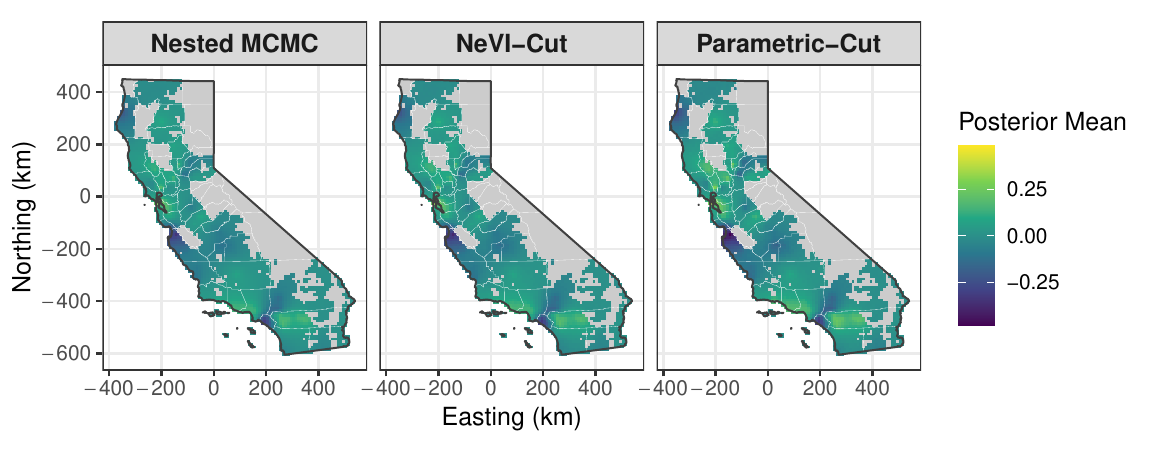}
    \caption{Posterior Mean}
    \label{fig:w_mean_compare}
\end{subfigure}
\vfill
\begin{subfigure}[b]{0.75\textwidth}
    \centering
    \includegraphics[width=\textwidth]{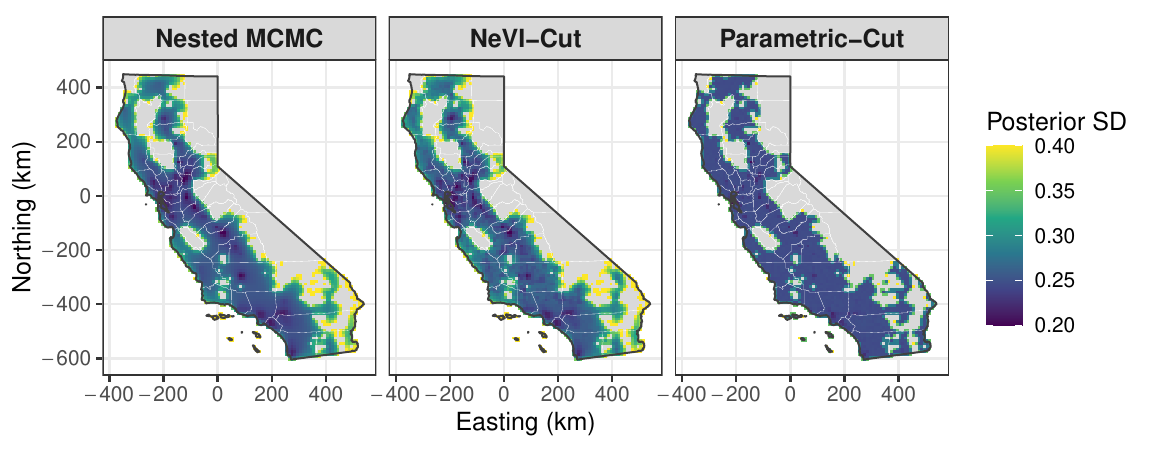}
    \caption{Posterior SD}
    \label{fig:w_sd_compare}
\end{subfigure}
\caption{\textbf{Global Burden of Disease in California.} Posterior distributions of the spatial random effects estimated by nested MCMC, NeVI-Cut, and parametric NeVI-Cut. }
\label{fig:w_compare}
\end{figure}

We compare the posterior densities of the coefficients in Figure \ref{fig:gbd_coeff}. NeVI-Cut aligns well with nested MCMC, whereas parametric NeVI-Cut captures the posterior means reasonably well but underestimates posterior variance. We also compare the marginal Wasserstein distance for $\beta_{\log RR}$ in Table \ref{tab:beta_wasserstein_time}. NeVI-Cut achieves considerably smaller Wasserstein distance to Nested MCMC than the parametric cut. The posterior mean and standard deviation (SD) maps for the spatial random effects $w$ are shown in Figure \ref{fig:w_compare}.
The NeVI-Cut estimates closely match those obtained from nested MCMC for both the posterior mean and variance. In contrast, the parametric cut approximation yields accurate posterior means but underestimates posterior variances, likely due to the mean-field factorization assumption. We further compare computational time. Running nested MCMC over all $1000$ upstream posterior draws required 330 hours in total, making it computationally impractical for routine use. In contrast, NeVI-Cut required only 27.24 hours for training and 7.8 hours for posterior sampling, yielding around nine-fold reduction in computation time.

\section{Architecture of NeVI-Cut}
\label{app:sensitivity}

\begin{figure}[!t]
\centering
\begin{subfigure}[b]{\textwidth}
    \centering
    \includegraphics[width=0.75\textwidth]{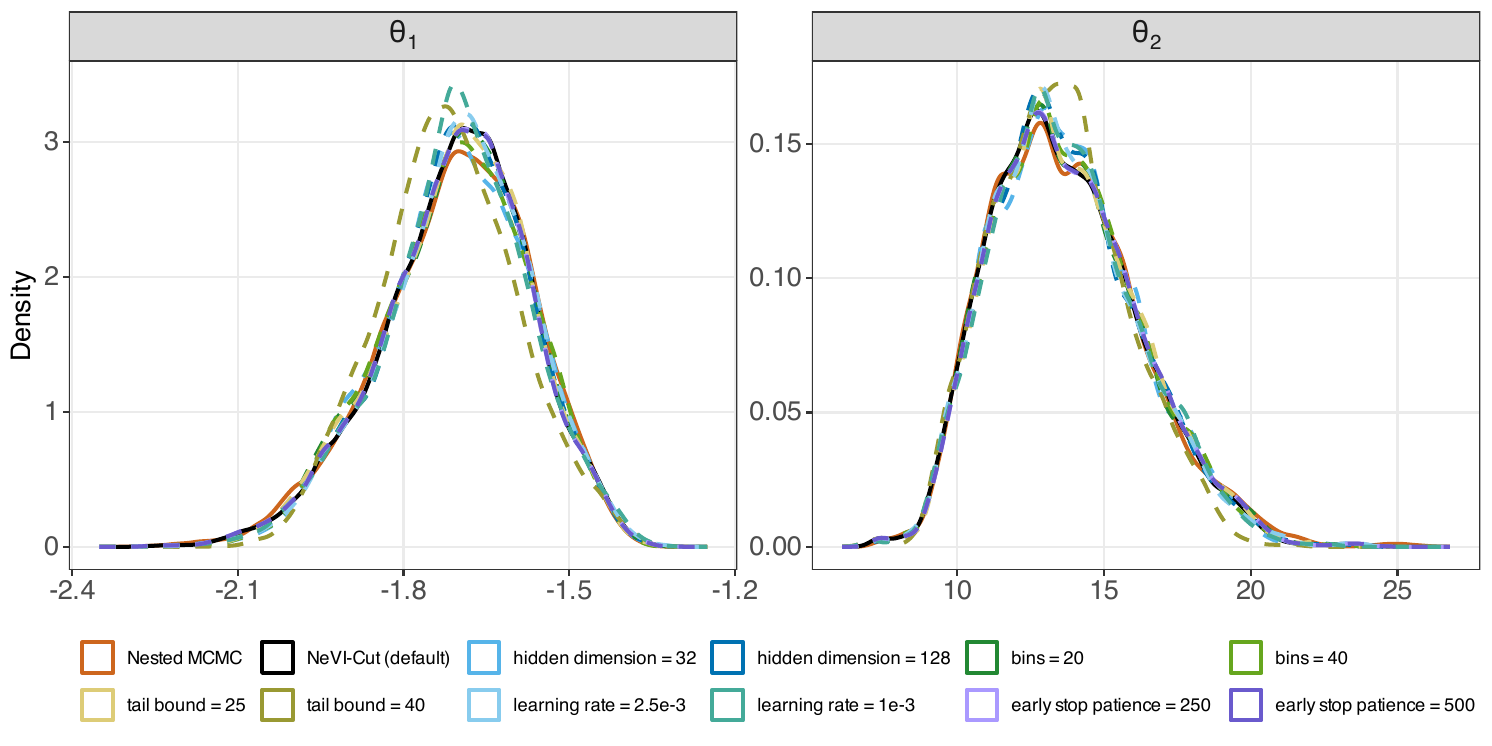}
    \caption{HPV}
\end{subfigure}
\vfill
\begin{subfigure}[b]{\textwidth}
    \centering
    \includegraphics[width=\textwidth]{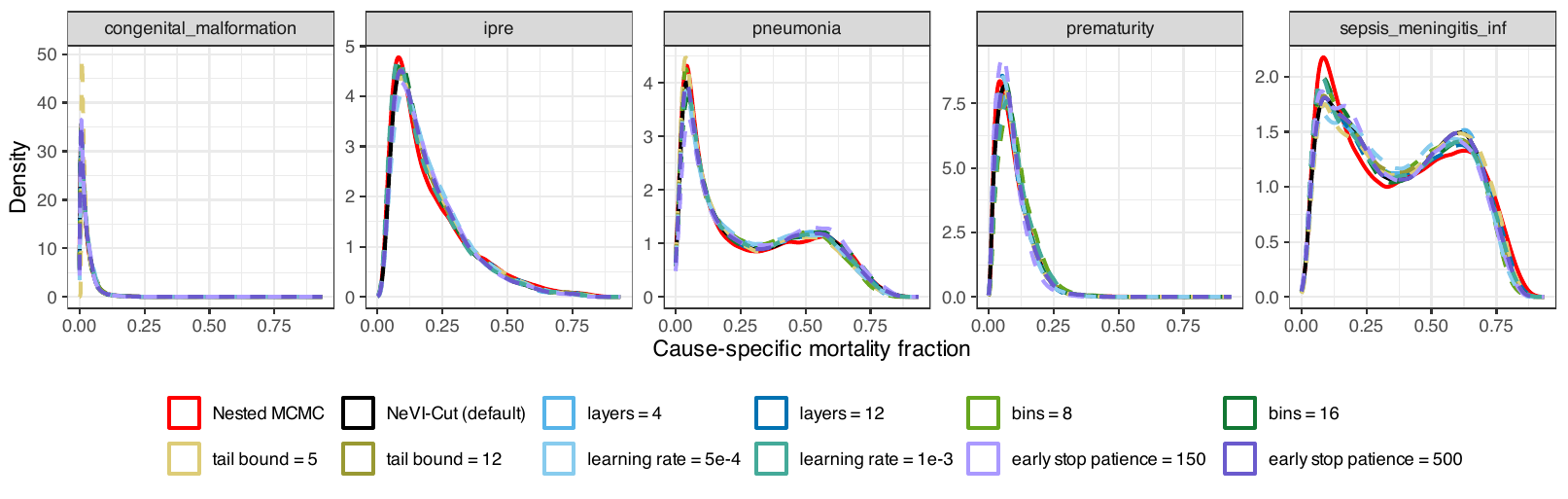}
    \caption{COMSA}
\end{subfigure}
\vfill
\begin{subfigure}[b]{\textwidth}
    \centering
    \includegraphics[width=\textwidth]{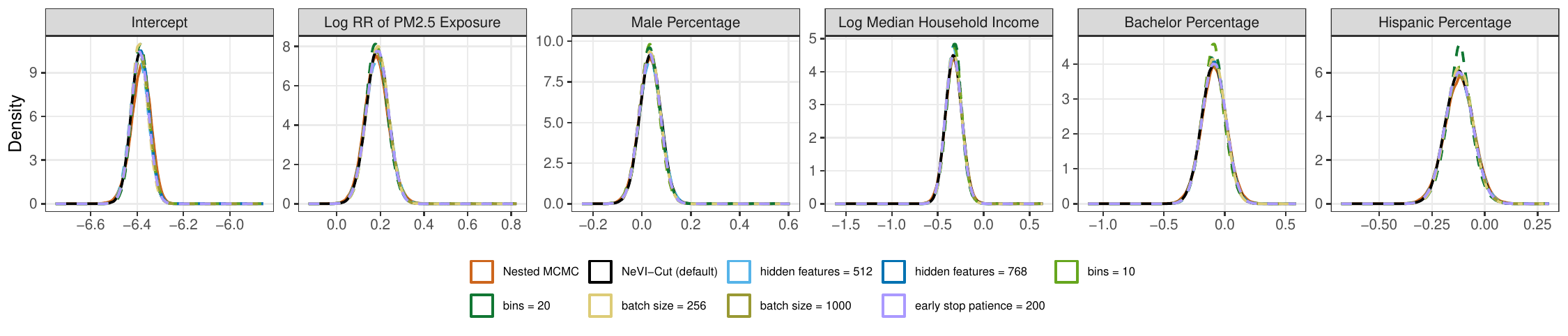}
    \caption{GBD}
\end{subfigure}
\caption{Choices of tuning parameters for NeVI-Cut in the HPV, COMSA, and GBD data analyses. The default NeVI-Cut specification uses the tuning parameters reported in Table \ref{tab:default-tuning}. Each additional line represents a sensitivity analysis in which the indicated tuning parameter is modified while all other settings are held at their default values.}
\label{fig:sensitivity}
\end{figure}

We assess the sensitivity of the NeVI-Cut posterior approximations to the main tuning parameters in both the flow architecture and the training procedure.

In the primary analysis, NeVI-Cut uses an autoregressive neural spline flow. The default configuration for the package and what is used for each analysis is summarized in Table~\ref{tab:default-tuning}. We consider three types of sensitivity analyses. First, we vary the neural spline flow architecture while holding the training data and likelihood specification fixed, including shallower and deeper flows, smaller and larger hidden dimensions, and different numbers of spline bins. Second, we vary the tail-bound parameter, which controls the range over which the spline transformation remains flexible. Third, we assess training sensitivity by varying the learning rate and the early-stopping rule. Figure~\ref{fig:sensitivity} summarizes the sensitivity results for the HPV, COMSA, and GBD examples. Across these settings, the posterior densities from the sensitivity analyses remain closely aligned with those from the primary NeVI-Cut analysis and nested MCMC.

\begin{table}[!h]
\centering
\caption{Tuning parameters for the NeVI-Cut in the default setting, HPV, COMSA and GBD data.}
\label{tab:default-tuning}
\begin{tabular}{lcccc}
\noalign{\hrule height 1.5pt}
\textbf{Parameter} & \textbf{Default} & \textbf{HPV} & \textbf{COMSA-Moz} & \textbf{GBD} \\
\noalign{\hrule height 1.5pt}
Flow type & NSF-AR & NSF-AR & NSF-AR & NSF-AR \\
Number of layers & 8 & 2 & 8& 4 \\
Hidden dimension & 64 & 64 & 32 & 640\\
Number of spline bins & 16 & 30 & 10 & 15\\
Tail bound & 10.0 & 35.0 & 8.0 & 15.0\\
Tail type & Linear & Linear & Linear & Linear \\
Residual blocks & Yes & Yes & Yes  & Yes \\
Number of residual blocks & 2 & 2 & 2 & 2 \\
Map to simplex & No & No & Yes & No\\
Mini-Batch Size & Full Batch & Full Batch & Full Batch & 512\\
Learning rate & $10^{-3}$ & $1.5\times10^{-3}$ & $7.5 \times 10^{-4}$ & $ 10^{-4}$ \\
Training epochs & 3000 & 5000 & 3500 & 5000 \\
Patience & 100 & 500 & 200 & 100\\
Early stopping patience & -- & -- & 300 & 300\\
\noalign{\hrule height 1.5pt}
\end{tabular}
\end{table}

\begin{singlespace}
\bibliography{references}
\end{singlespace}

\end{document}